\documentclass[Afour,sageh,times]{sagej}

\usepackage{moreverb,url}

\usepackage{algorithm}
\usepackage{algcompatible}
\usepackage{amsfonts}
\usepackage{amsmath}
\usepackage{graphicx}
\usepackage{subcaption}
\usepackage{makecell}
\usepackage{color}

\usepackage[colorlinks,bookmarksopen,bookmarksnumbered,citecolor=red,urlcolor=red]{hyperref}

\newcommand\BibTeX{{\rmfamily B\kern-.05em \textsc{i\kern-.025em b}\kern-.08em
T\kern-.1667em\lower.7ex\hbox{E}\kern-.125emX}}

\newcommand{\vect}[1]{{\boldsymbol{#1}}}

\newcommand{\Real}{\mathbb{R} }
\newcommand{\E}{\mathbb{E} }

\newcommand{\target}{\textrm{target}}
\newcommand{\prop}{\textrm{prop}}
\newcommand{\intd}{\textrm{d} }
\newcommand{\KL}{D_{\textrm{KL}}}
\newcommand{\residual}{\textrm{residual}}
\newcommand{\base}{\textrm{base}}

\def\volumeyear{2021}
\begin{document}

\runninghead{Osa}

\title{Motion Planning by Learning the Solution Manifold in Trajectory Optimization}

\author{Takayuki Osa\affilnum{1}\affilnum{2}}

\affiliation{\affilnum{1}Kyushu Institute of Technology, Japan\\
	\affilnum{2}RIKEN Center for Advanced Intelligence Project, Japan}

\corrauth{Takayuki Osa, Kyushu Institute of Technology
	Department of Human Intelligence Systems \&
	Research Center for Neuromorphic AI Hardware
	Behavior Learning Systems Loboratory,
	Hibikino 2-4,
	Wakamatsu, Kitakyushu, Fukuoka,
	808-0135, Japan.}

\email{osa@brain.kyutech.ac.jp}

\begin{abstract}
The objective function used in trajectory optimization is often non-convex and can have an infinite set of local optima. In such cases, there are diverse solutions to perform a given task.
Although there are a few methods to find multiple solutions for motion planning, they are limited to generating a finite set of solutions.
To address this issue, we presents an optimization method that learns an infinite set of solutions in trajectory optimization. 
In our framework, diverse solutions are obtained by learning latent representations of solutions.
Our approach can be interpreted as training a deep generative model of collision-free trajectories for motion planning.
The experimental results indicate that the trained model represents an infinite set of homotopic solutions for motion planning problems.
\end{abstract}


\maketitle

\section{Introduction}
Motion planning has been investigated for decades in the field of robotics because it is an essential component in many robotic systems.
A popular approach for motion planning is optimization-based methods, which finds a solution that minimizes the objective function.
Various optimization methods have been leveraged in motion planning in robotics~\citep{Khatib86,Zucker13,Schulman14}. 
Ideally, an objective function should be designed such that the solution is unique and the optimization problem can be solved stably.
If we can formulate our problem as a convex optimization, we can leverage the techniques established in previous studies~\citep{Boyd04}.
However, the objective function in motion planning is often non-convex and may have many local optima.
To circumvent the difficulty of optimizing the trajectory with respect to the non-convex objective function, prior work has focused on robustly finding a solution.
As previous studies have established various methods for robustly finding a single solution in motion planning, we address an unexplored aspect of motion planning, that is, finding diverse solutions.

\begin{figure}
	\centering
	\begin{subfigure}[t]{0.48\columnwidth}
		\centering
		\includegraphics[width=\textwidth]{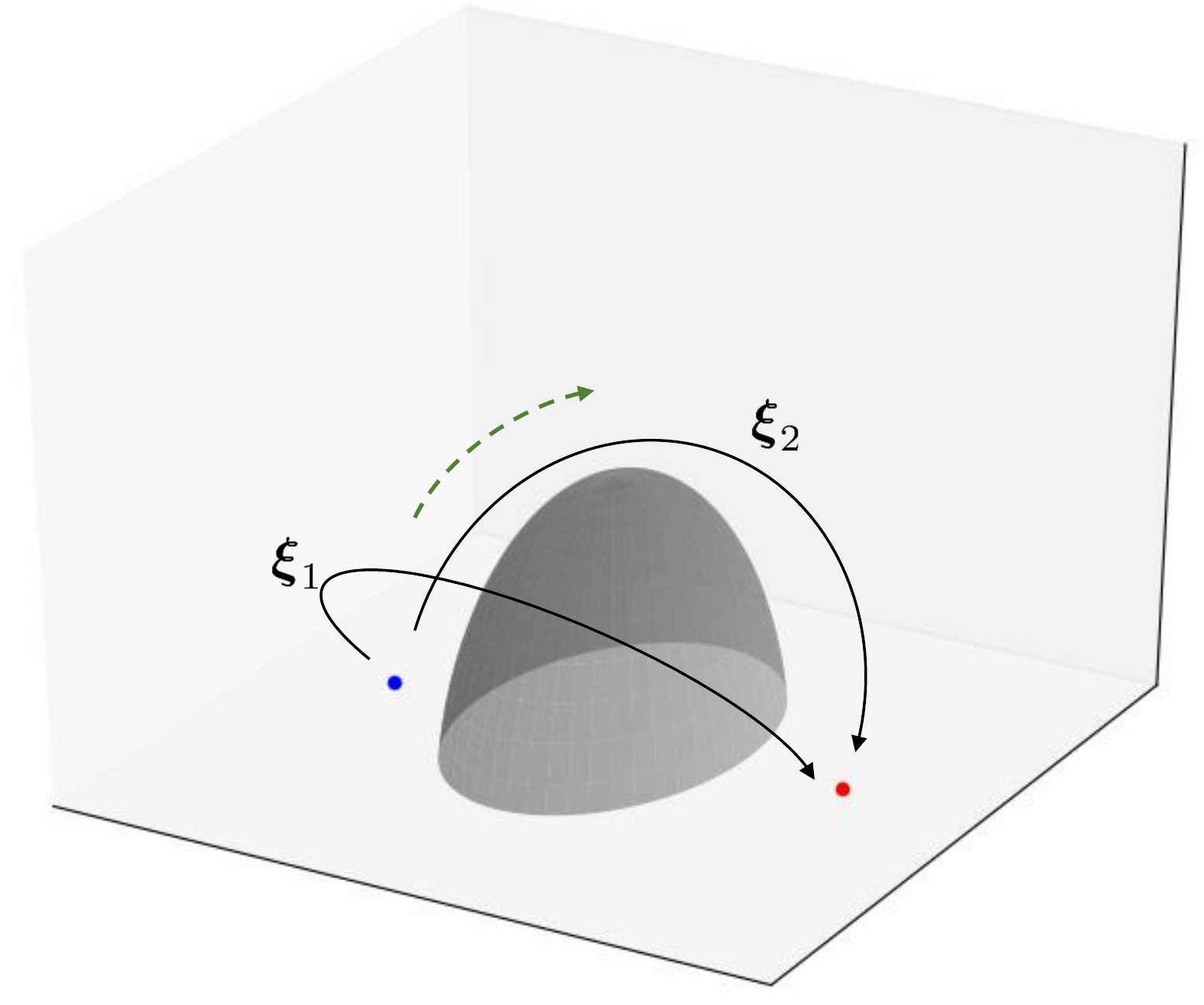}
		\caption{Homotopic collision-free paths in 3D.  Paths $\vect{\xi}_1$ and $\vect{\xi}_2$ are homotopic.}
	\end{subfigure}
	\begin{subfigure}[t]{0.48\columnwidth}
		\centering
		\includegraphics[width=\textwidth]{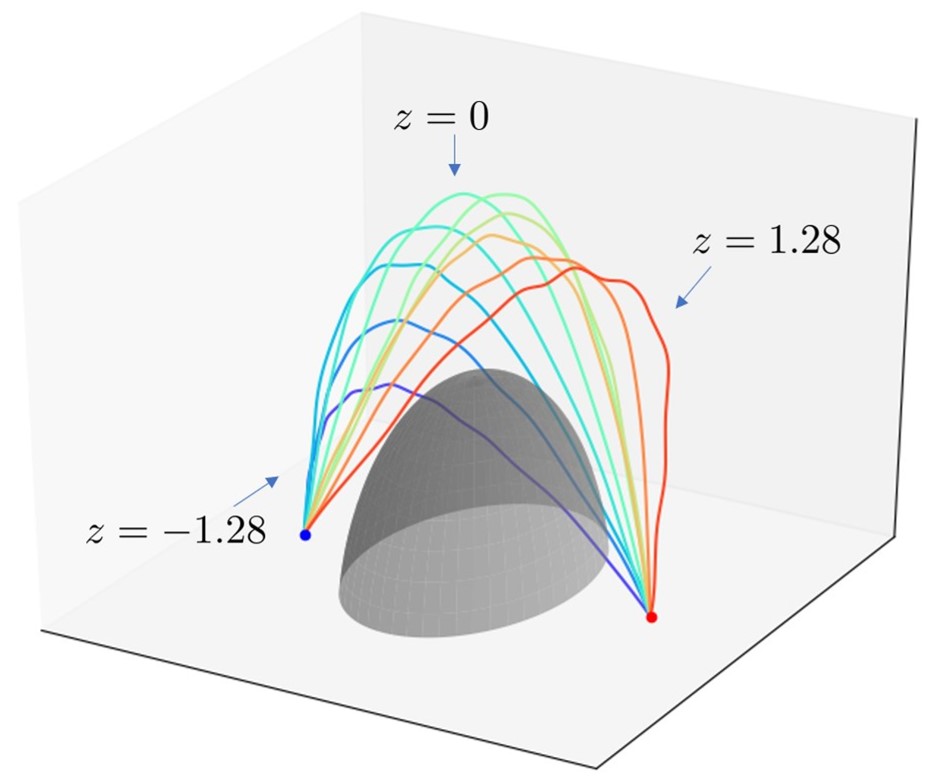}
		\caption{Generation of homotopic solutions with the proposed method.}
	\end{subfigure}
	\caption{Example of homotopic solutions in motion planning.
	} 
	\label{fig:primitive}
\end{figure}

Two trajectories are called \textit{homotopic} when one can be continuously deformed into another~\citep{Jaillet08,Hatcher02}. 
Previous studies by ~\cite{Jaillet08, Orthey20} indicated that there may be an infinite set of homotopic solutions in motion planning.  
Figure~\ref{fig:primitive} shows an example of homotopic collision-free paths in motion planning.
Paths $\vect{\xi}_1$ and $\vect{\xi}_2$ are homotopic, and both are collision-free.
Although only two paths are visualized in Figure~\ref{fig:primitive}, there is an infinite set of collision-free paths between $\vect{\xi}_1$ and $\vect{\xi}_2$.
Existing motion planning methods often use a generic objective function, which is applicable for various tasks but often has many solutions.
It is possible to specify a unique solution by posing additional constraints, but this is impractical for users who are not experts on optimization and/or robotics.
For this reason, it is preferable to suggest various solutions and allow the users to select the preferable one among the candidates.
However, existing methods such as a method proposed by \cite{osa20} are limited to generating a fixed number of solutions, and it is not straightforward to generate a new “middle” solution out of the two obtained solutions.

\begin{figure*}[tb]
	\centering
	\includegraphics[width=2\columnwidth]{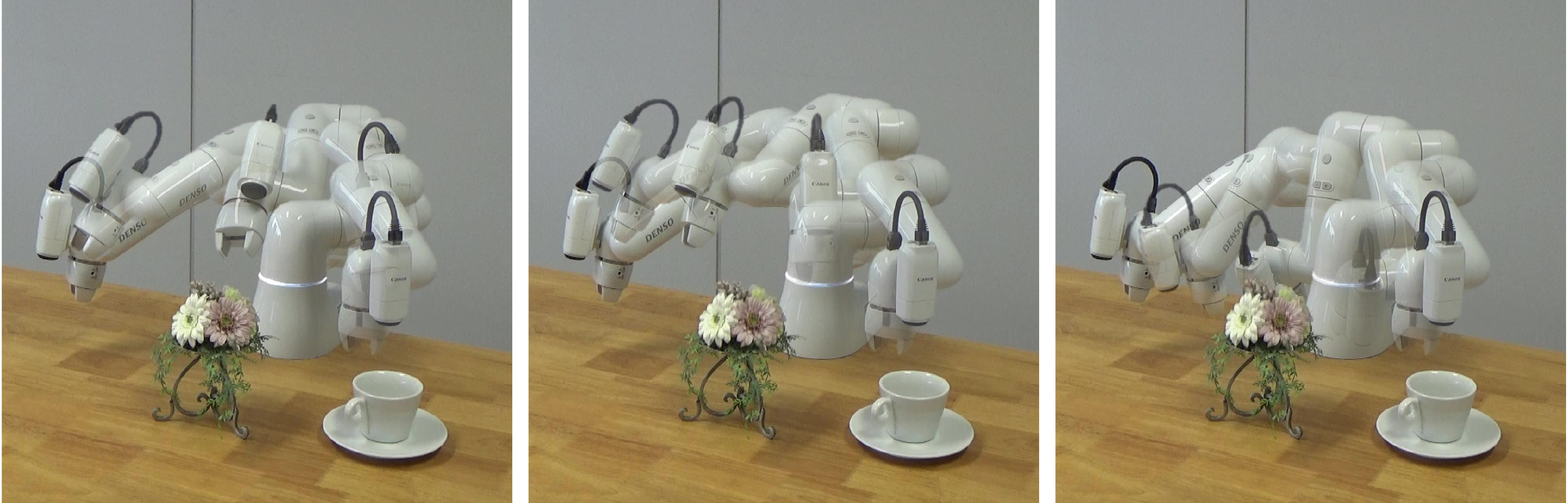}
	\caption{There often exist an infinite set of homotopic solutions in motion planning, although existing methods are often designed to find a single solution.  }
	\label{fig:intro}
\end{figure*}

Our contribution is to propose a framework that can generate diverse collision-free paths in motion planning.
In Figure~\ref{fig:primitive}(b), $z$ represents the latent variable learned by the proposed method, and various solutions can be generated by changing the values of $z$.
In this study, we present a practical algorithm for capturing diverse solutions in motion planning.
We first propose an optimization method that learns an infinite set of solutions for optimization.
Our approach can be interpreted as training a deep generative model of optimal points in optimization.
In the field of information geometry, an $n$-dimensional \textit{manifold} is defined as a set of points such that each point has $n$-dimensional extensions in its neighborhood, 
and a point in a manifold can be specified using its coordinates~\citep{Amari16}.
Our framework encodes the diversity of the solutions into a continuous latent variable, and each solution has extensions in its neighborhood.
In addition, each solution can be specified using the value of latent variable.
In this context, our method learns the manifold of solutions in optimization, and the learned latent variable can be regarded as the coordinates of the obtained manifold.
Therefore, we refer to our optimization method as \textit{Learning the Solution Manifold in Optimization}~(LSMO).

In this paper, we first present the derivation of LSMO.
We then present a motion planning algorithm based on LSMO, which we refer to as Motion Planning by Learning the Solution Manifold~(MPSM), and describe how we can adapt LSMO to solve the motion planning problems efficiently.
In MPSM, a neural network that takes in a latent variable is trained to generate diverse collision-free trajectories.
Using our approach, the variation of collision-free trajectories are encoded in a low-dimensional latent variable, and a user can intuitively obtain various collision-free trajectories in motion planning by changing the value of the latent variable.
In experiments, to analyze the behavior of LSMO, we first applied it to tasks of maximizing test functions that have an infinite set of optimal points. 
We then evaluate MPSM on motion-planning tasks for a robotic manipulator.
We empirically showed that MPSM can learn an infinite set of homotopic trajectories in motion-planning problems.

\section{Related Work}
\subsection{Motion planning methods in robotics} 
A popular class of motion planning methods is optimization-based methods, and Covariant Hamiltonian Optimization for Motion Planning~(CHOMP)~\citep{Zucker13}, STOMP~\citep{Kalakrishnan11}, TrajOpt~\citep{Schulman14}, and Gaussian Process Motion Planner~(GPMP)~\citep{Mukadam18} are included in this class.
These methods determine the trajectory that minimizes the objective function, which quantifies the quality of the trajectory.
Sampling-based methods are also popular in motion planning for robotic systems.
Probabilistic RoadMap (PRM)~\citep{Kavraki96,Kavraki98}, Rapidly-exploring Random
Trees (RRT)~\citep{LaValle01}, RRT*~\citep{Karaman11}, BIT*~\citep{Gammell20} are categorized as sampling-based methods. 
In sampling-based methods, collision-free configurations are sampled in a stochastic manner, and a path is planned by connecting the sampled configurations.
Prior work showed that sampling-based methods work well for complex motion planning problems such as maze tasks. 
In addition, previous studies by~\cite{Koert16,Osa17,Rana17,Rana20,Mukadam20} incorporated the learning-from-demonstration approach~\citep{Argall09,Osa18} with optimization-based and sampling-based approaches.
Recent studies proposed methods that leverage the capability of neural networks~\citep{Srinivas18,Jurgenson19,Chen20}.
These deep-learning-based methods are often built on sampling-based or optimization-based methods, and they exploit the generalization ability of neural networks to achieve robust and/or computationally efficient motion planning. 
Although the focus of this study is the optimization-based method, our work will also contribute to other types of motion planning methods because the various types of motion-planning methods interactively influence the advances in the field of motion planning.

\subsection{Finding Multiple Solutions in Motion planning} 
Works by \cite{Jaillet08, Orthey20} discussed homotopy and the deformability of trajectories, and they implied that there may be an infinite set of collision-free trajectories in motion planning.
Figure~\ref{fig:intro} shows an example of such motion planning tasks.
However, existing methods often ignore the existence of multiple solutions and find only a single solution.
As a consequence, a motion planner may generate a solution that is different from the one that the user expected.
If the user wishes to obtain another type of solution, then she/he needs to modify the objective function of motion planning or re-run the motion planner with different random seeds until a preferable solution is obtained. 
Recent studies by \cite{Toussaint18,Toussaint20} addressed manipulation planning by formulating a problem as a Logic-Geometric Program~(LGP)~\citep{Toussaint15}.
Their method finds diverse plans to achieve a physical manipulation task using a tree-search-based algorithm.
The study by  \citet{Orthey20} employed a tree-based architecture to represent multiple local minima in motion planning.
\cite{osa20} also recently proposed a motion planning algorithm that finds multiple solutions based on multimodal optimization.
Although these methods allow the user to examine diverse solutions and select a preferable one among a given set of solutions, they are limited to generating a finite number of solutions.
If the preferable solution is not included in the given set, then the user will need to re-run the motion planner with different random seeds as in other motion planning methods.
In this work, we propose a framework that models an infinite set of solutions for motion planning.
As our framework learns the continuous latent representation of solutions, the user will be able to examine diverse solutions intuitively and tune the type of solutions efficiently.

\subsection{Multimodal optimization with black-box optimization methods}
Prior studies~\citep{Goldberg87,Deb10,Stoean10,Agrawal14,Karasawa20} addressed multimodal optimization using black-box optimization methods, such as CMA-ES~\citep{Hansen96} and the cross-entropy method~(CEM)~\citep{Boer05}.
For example, a study by \cite{Agrawal14} showed that control policies to achieve diverse behaviors can be learned by maximizing the objective function that encodes the diversity of solutions.
Although these black-box optimization methods are applicable to a wide range of problems, it is difficult to apply them to the optimization of high-dimensional parameters.
Motion planning tasks usually involve high dimensional parameters.
For example, if the manipulator has seven degrees of freedoms~(DoFs) and a trajectory is represented by 50 time steps, then the trajectory of the manipulator is represented by a vector with 350 dimensions.
For this reason, it is not trivial to directly apply black-box optimization methods to motion planning tasks in robotics.  
In addition, the limitation of black-box multimodal optimization methods is that they can learn only a finite set of solutions.

\subsection{Latent Representations in Reinforcement learning and imitation learning}
Recent studies on imitation learning proposed methods for modeling diverse behaviors by learning latent representations~\citep{Li17,Merel19,Sharma19}.
For example, \citet{Merel19} proposed a method for learning diverse behaviors by learning continuous latent variables.
Likewise, in the field of reinforcement learning~(RL), previous studies proposed methods for learning the diverse behaviors~\citep{Eysenbach19}.
Hierarchical RL methods~\citep{Bacon17,Florensa17,Vezhnevets17,Osa19} often learn a hierarchical policy given by  $\pi(\vect{a}|\vect{s})=\sum_{o \in \mathcal{O}} \pi(o|\vect{s})\pi(\vect{a}|\vect{s}, o)$, 
where $\vect{s}$, $\vect{a}$, and $o$ denote the state, action, and option, respectively.
These methods can be interpreted as approaches that models diverse behaviors with a policy conditioned on the latent variable.
Recent studies~\citep{Nachum18,Nachum19,Schaul15} investigated goal-conditioned policies $\pi(\vect{a}|\vect{s}, \vect{g}) $, where $\vect{g}$ denotes the goal. 
A goal-conditioned policy can also be viewed as a way of modeling diverse behaviors conditioned on a latent variable that has semantic meaning. 
However, the problem setting of RL is different from ours because our problem formulation does not involve the Markov decision process.


\section{Problem Formulation}
We denote by $\vect{q}_t \in \Real^{D}$ the configuration of a robot manipulator with $D$ degrees of freedom~(DoFs) at time $t$.
Given the start configuration $\vect{q}_0$ and the goal configuration $\vect{q}_T$, the task is to plan a smooth and collision-free trajectory $\vect{\xi} = [\vect{q}_0, \ldots,\vect{q}_T  ] \in \Real^{D \times T} $, which is given by a sequence of robot configurations.
This problem can be formulated as an optimization problem 
\begin{align}
\vect{\xi}^* = \arg \max_{\vect{\xi}} R(\vect{\xi})
\label{eq:formulation}
\end{align}
where $R(\vect{\xi})$ is the score function that quantifies the quality of a trajectory $\vect{\xi}$.
Although previous studies formulate the motion planning problem as minimization of the cost function $\mathcal{C}(\vect{\xi})$, we employ the formulation in \eqref{eq:formulation} to make the following discussion concise.
In the proposed algorithm, we train the model $p_{\vect{\theta}}(\vect{\xi}|\vect{z})$ to represent a distribution of optimal points.

\begin{figure*}
	\centering
	\begin{subfigure}[t]{0.6\columnwidth}
		\centering
		\includegraphics[width=\textwidth]{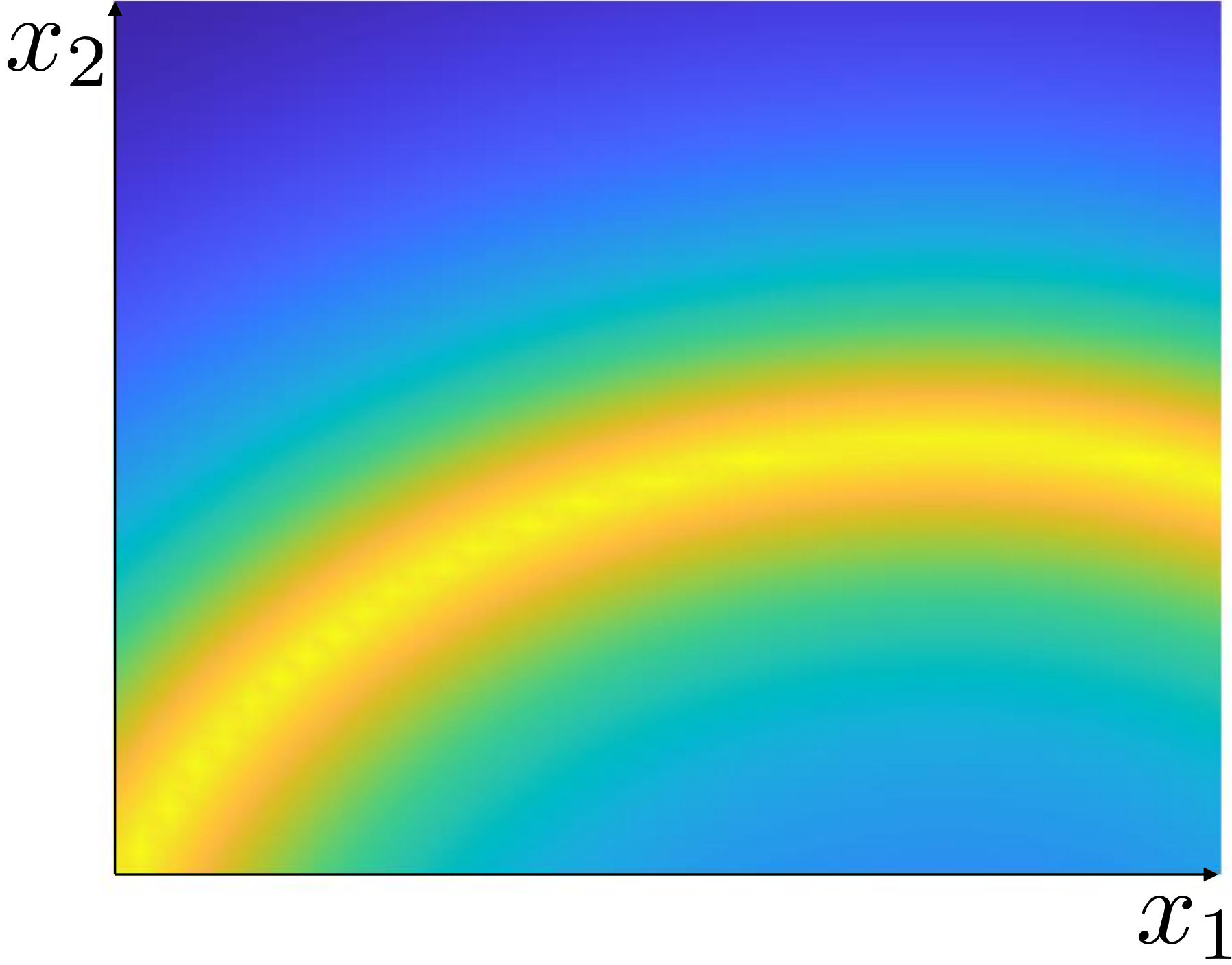}
		\caption{Visualization of the objective function that has an infinite set of optimal points. }
	\end{subfigure}
	\begin{subfigure}[t]{0.6\columnwidth}
		\centering
		\includegraphics[width=\textwidth]{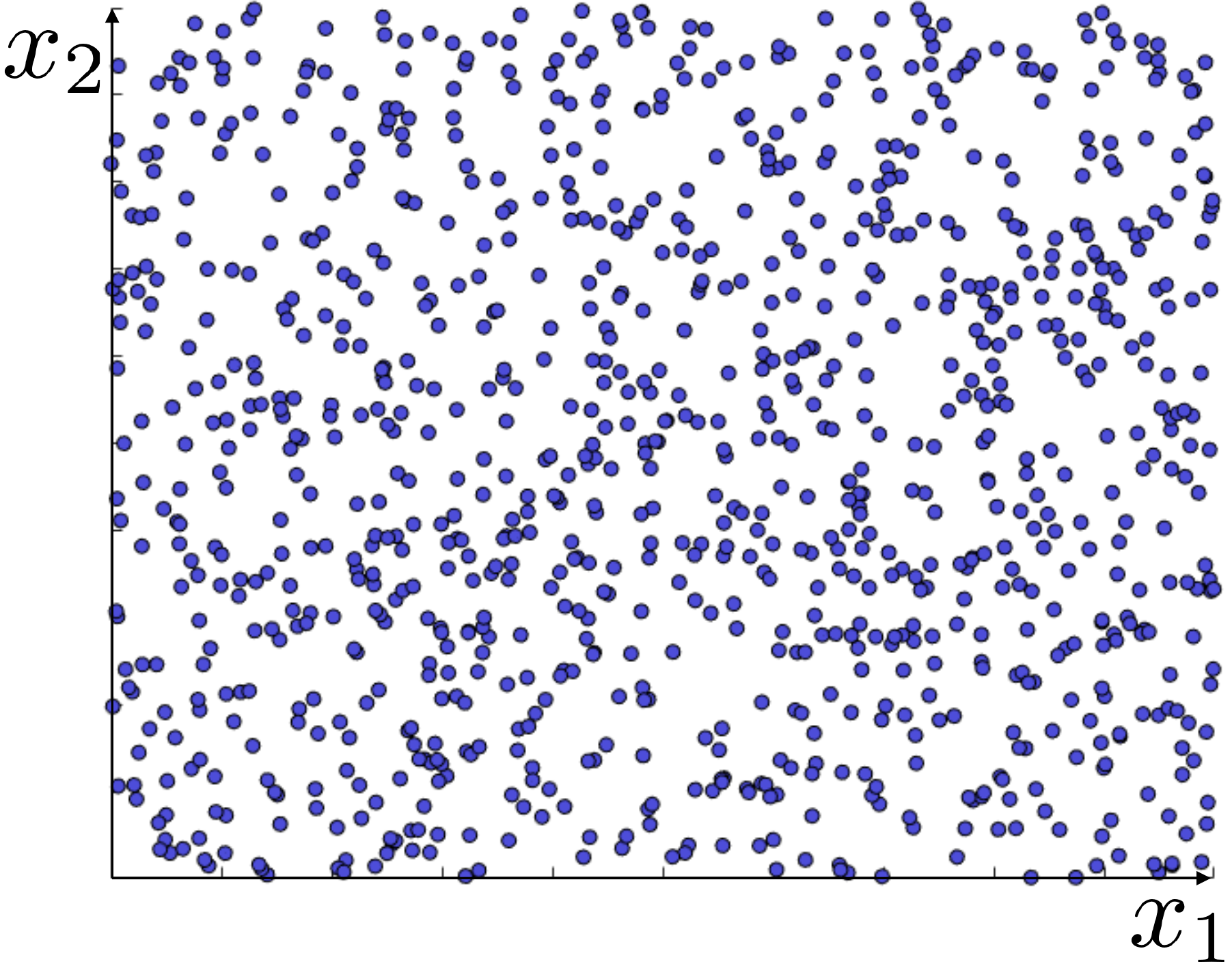}
		\caption{Samples drawn from the uniform distribution.}
	\end{subfigure}
	\begin{subfigure}[t]{0.6\columnwidth}
		\centering
		\includegraphics[width=\textwidth]{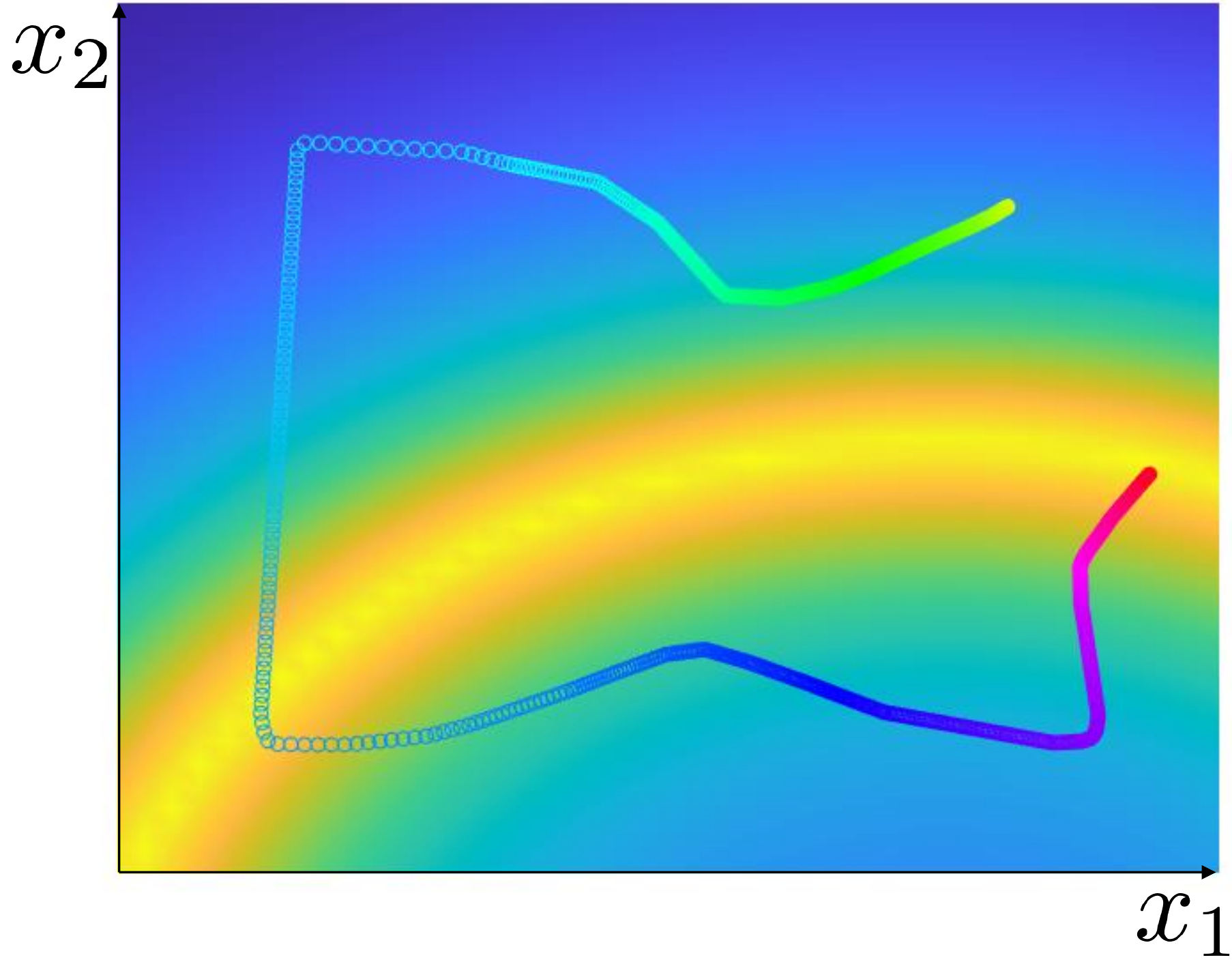}
		\caption{Visualization of the outputs of the model trained with the VAE objective function.}
	\end{subfigure}
	\begin{subfigure}[t]{0.6\columnwidth}
		\centering
		\includegraphics[width=\textwidth]{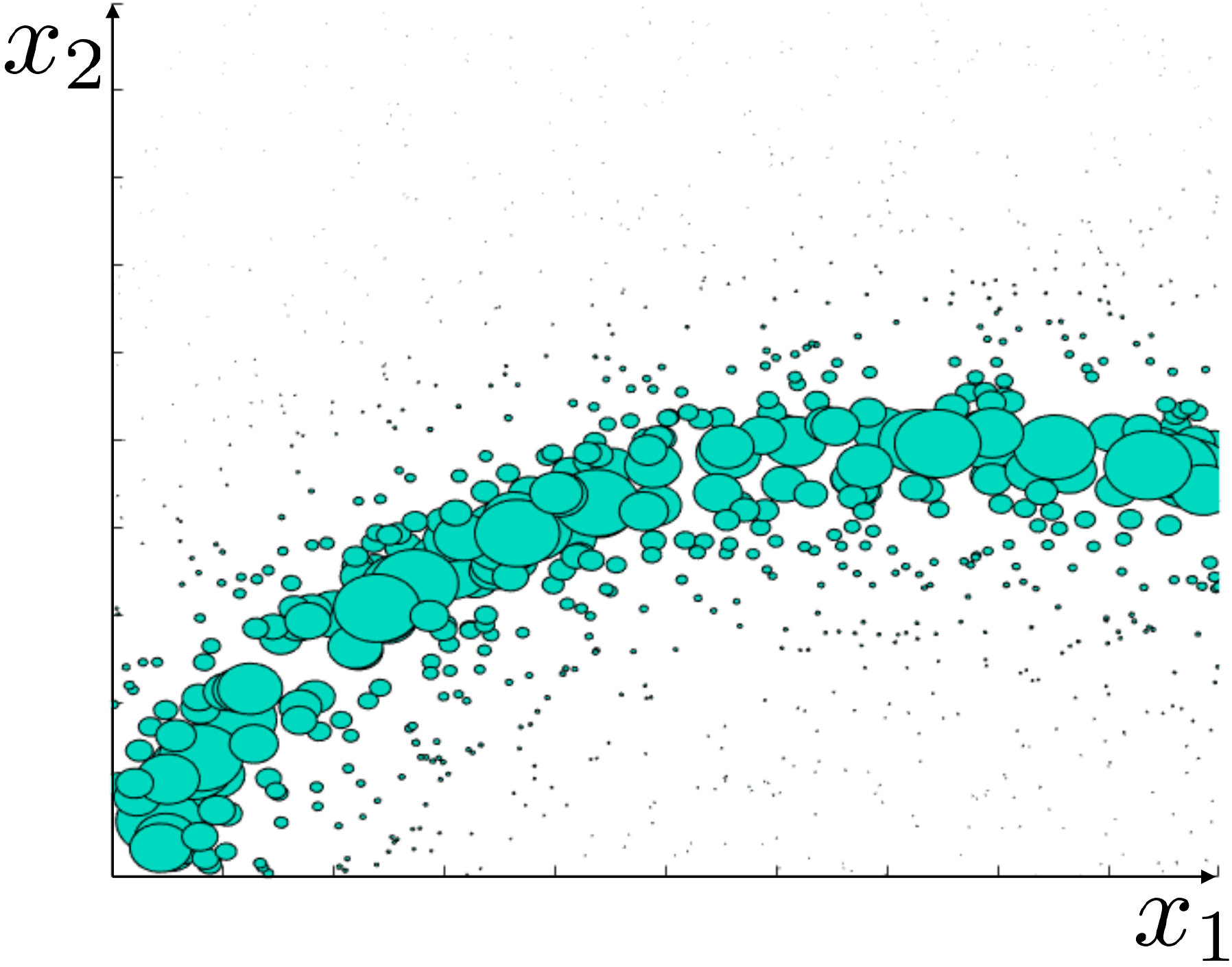}
		\caption{The same samples in (b) with the scaling based on $f\big(R(\vect{x})\big)$.}
	\end{subfigure}
	\begin{subfigure}[t]{0.6\columnwidth}
		\centering
		\includegraphics[width=\textwidth]{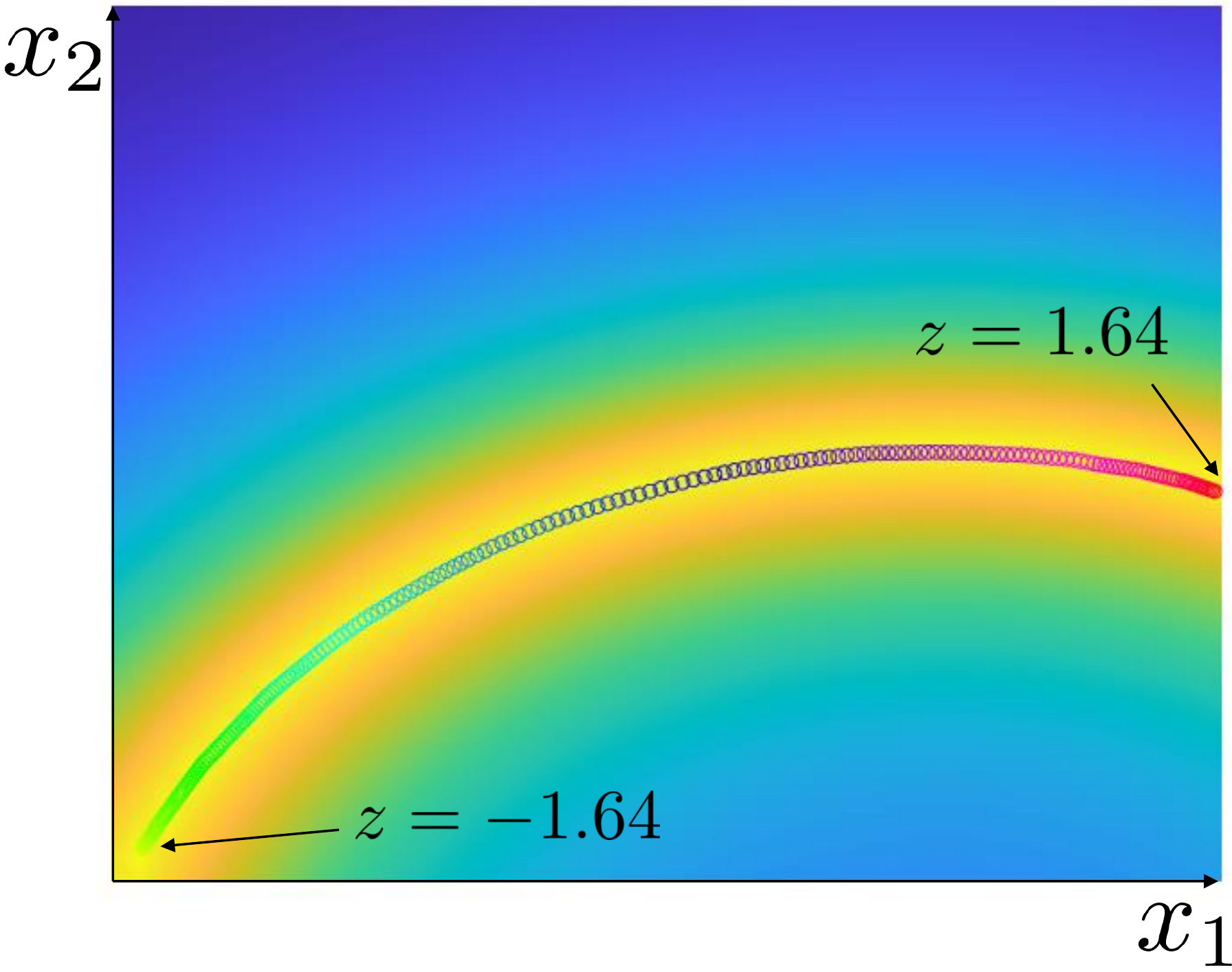}
		\caption{Visualization of the outputs of the model trained with LSMO.}
	\end{subfigure}
	\caption{Example of learning the solution manifold.  (a) shows an objective function that has an infinite set of solutions. The warmer color represents the higher score in (a).  
	(b) shows samples drawn from a uniform distribution. 
		In (c), circles represent the output of the model trained with the VAE objective, and the color of the circle indicates the value of the latent variable.
		In (d), the same samples as those in (b) are shown, but samples with higher importance are drawn as larger circles. We train $p_{\vect{\theta}}(\vect{x}|\vect{z})$ with this importance weight. 
		In (e), circles represent the output of the model  $p_{\vect{\theta}}(\vect{x}|\vect{z})$ trained with LSMO, and the color of the circle indicates the value of the latent variable.
		The output of the trained model continuously changes by changing the value of the latent variable continuously.
	} 
	\label{fig:approach}
\end{figure*}

In this study, we are particularly interested in problems where there exist multiple, and possibly infinite solutions.
For example, the objective function shown in Figure~\ref{fig:approach}(a) has an infinite number of solutions.
The goal of our study is to learn the solution manifold when optimizing such objective functions.
Instead of finding a single solution, we aim to train a model that represents the distribution of optimal points $p_{\vect{\theta}}(\vect{\xi})$ parameterized with a vector $\vect{\theta}$ given by  
\begin{align}
p_{\vect{\theta}}(\vect{\xi})= \int p_{\vect{\theta}}(\vect{\xi} | \vect{z} )p(\vect{z}) \textrm{d}\vect{z},
\label{eq:model}
\end{align}
where $\vect{z}$ is the latent variable.
We train the model $p_{\vect{\theta}}(\vect{x} | \vect{z} )$ by maximizing the surrogate objective function
\begin{align}
J(\vect{\theta}) = \E_{\vect{\xi} \sim p_{\vect{\theta}}(\vect{\xi})} [ f\big(R(\vect{\xi})\big) ],
\label{eq:surrogate_obj}
\end{align}
where $f(\cdot)$ is monotonically increasing and $f(x) \geq 0$ for any $x \in \Real$.
Because $f(\cdot)$ is monotonically increasing, maximizing $R(\vect{\xi})$ is equivalent to maximizing $f(R(\vect{\xi}))$. Therefore, $f(\cdot)$ can be interpreted as a shaping function.
This shaping function $f(\cdot)$ is used to derive the proposed algorithm in Section~\ref{sec:derivation}.

\section{Learning Solution Manifold in Optimization}
If a dataset of diverse solutions is available, we can employ standard machine-learning techniques, such as generative adversarial networks~(GANs)~\citep{Goodfellow14} and variational autoencoders~(VAEs)~\citep{Kingma14} to train a generative model of optimal solutions. 
However, obtaining such a dataset of diverse solutions is challenging in practice.
Hence, we present an algorithm for training the generative model of optimal solutions using non-optimal samples obtained stochastically from a proposal distribution. 

\subsection{Overview of Proposed Optimization Algorithm}
We first present an overview of our optimization algorithm, which can be applied to general optimization problems that are not limited to the motion planning problems.
In this section, we consider the problem of maximizing the objective function
\begin{align}
J(\vect{\theta}) = \E_{\vect{x} \sim p_{\vect{\theta}}(\vect{x})} [ f\big(R(\vect{x})\big) ],
\label{eq:surrogate_obj_x}
\end{align}
where $p_{\vect{\theta}}(\vect{x})$ represent a model parameterized by a vector $\vect{\theta}$.
\begin{align}
p_{\vect{\theta}}(\vect{x})= \int p_{\vect{\theta}}(\vect{x} | \vect{z} )p(\vect{z}) \textrm{d}\vect{z}.
\label{eq:model_x}
\end{align}
Here, $\vect{x}$ represents a data point in general and is not necessarily a trajectory.

The main concept of the proposed algorithm is illustrated in Figure~\ref{fig:approach}.
We focused on optimization problems that involve an infinite set of optimal points, as shown in ~Figure~\ref{fig:approach}(a).
The aim of our algorithm is to train a generative model of the optimal points. 

To this end, we used non-optimal samples to train the model, e.g., samples obtained from a uniform distribution, as shown in Figure~\ref{fig:approach}(b). 
If we naively use the original VAE objective function for the training, then the resulting model will generate points that do not correspond to the optimal solutions, as shown in Figure~\ref{fig:approach}(c). 
To address this issue, we introduce an importance weight based on $f(R(\vect{x}))$.
If we scale samples based on $f(R(\vect{x}))$, then we can obtain a density whose modes correspond to the optimal solutions, as shown in Figure~\ref{fig:approach}(d).
In the proposed method, the generative model $p_{\vect{\theta}}(\vect{x}|\vect{z})$ is trained as a part of VAE by maximizing the weighted log-likelihood with respect to samples obtained from a proposal distribution.

In Fig.~\ref{fig:approach}(e), circles represent the output of the model trained with LSMO, and the color of the circle indicates the value of the latent variable.
As shown in Fig.~\ref{fig:approach}(e), the output of the trained model continuously changes by continuously changing the value of the latent variable.
Because the value of the latent variable indicates the similarity of the solution, the user can intuitively go through various solutions using the model trained with LSMO.


When generating a solution from the trained model, the user specifies the value of the latent value $\vect{z}$ and generates a sample from the trained neural network. 
As the output of the neural network model is the result of inference, it may not correspond to the exact optimal point in the objective function.
Therefore, the output of the neural network model is fine-tuned to obtain the exact solution if necessary.
The proposed algorithm is summarized in Algorithm~\ref{alg:LSMO}.
In the next section, we present the derivation of the proposed method and show the relation between the objective function in \eqref{eq:surrogate_obj_x} and the weighted log-likelihood in the next section.

\begin{algorithm}[t]
	\caption{Abstract of Learning the Solution Manifold in Optimization~(LSMO) }
	\begin{algorithmic}[1]
		\STATEx{\textbf{Input:} Objective function $R(\vect{x})$, proposal distribution  $p^{\prop}(\vect{x})$, shaping function $f$}
		\STATEx{\textbf{Training phase:}}
		\STATE{Generate $N$ synthetic samples $\{\vect{x}_i\}^N_{i=1}$ from the proposal distribution $p^{\prop}(\vect{x})$  }
		\STATE{Evaluate the objective function $R(\vect{x}_i)$ and compute the weight $f\big(R(\vect{x}_i)\big)$ for $i=1,\ldots, N$  }
		\STATE{Train $p_{\vect{\theta}}(\vect{x} |\vect{z})$ by maximizing  $\mathcal{L}(\vect{\theta}, \vect{\psi})$ in \eqref{eq:loss} }
		\STATEx{\textbf{Generation phase:}}
		\STATE{Generate  $\vect{x}^*$ with $p_{\vect{\theta}} (\vect{x}|\vect{z})$ by specifying the value of $\vect{z}$}
		\STATE{(Optional) Fine-tune $\vect{x}^*$ with a gradient-based method }
		\STATEx{\textbf{Return:} $\vect{x}^*$ }
	\end{algorithmic}
	\label{alg:LSMO}
\end{algorithm}

\subsection{Learning Latent Representations in Optimization}
\label{sec:derivation}
To derive our algorithm, we  first consider the lower bound of the surrogate objective function $J(\vect{\theta} )$ in \eqref{eq:surrogate_obj_x}  in the same manner as in previous studies by ~\cite{Dayan97,Kober11,osa20}.
Although evaluating $J(\vect{\theta} )$ requires computing the expectation with respect to samples drawn from $p_{\vect{\theta}}( \vect{x})$, iteration of sampling from $p_{\vect{\theta}}( \vect{x})$ and updating $\vect{\theta}$ would be time-consuming, especially when $p_{\vect{\theta}}( \vect{x})$ is modeled with a neural network.
Instead, we consider a proposal distribution $p_{\prop}(\vect{x})$ for generating a set of samples $X=\{\vect{x}_i\}^N_{i=1}$. 
To derive the relation between $p_{\prop}(\vect{x})$ and $ J( \vect{\theta}  )$,  we apply Jensen's inequality as follows:
\begin{align}
\log J( \vect{\theta}  ) 
& = \log \int p_{\vect{\theta}}( \vect{x}) f\big(R(\vect{x})\big) \textrm{d}\vect{x}
\nonumber \\
&= \log \int p_{\textrm{prop}}( \vect{x})  f\big(R(\vect{x})\big) \frac{p_{\vect{\theta}}( \vect{x})}{ p_{\textrm{prop}}( \vect{x}) } 
 \textrm{d}\vect{x} \nonumber \\
&\geq  \int p_{\textrm{prop}}( \vect{x})  f\big(R(\vect{x})\big) \log \frac{p_{\vect{\theta}}( \vect{x})}{ p_{\textrm{prop}}( \vect{x}) } 
\textrm{d}\vect{x} \nonumber \\
&=   \int p_{\textrm{prop}}( \vect{x}) f\big(R(\vect{x})\big) \log p_{\vect{\theta}}( \vect{x}) \textrm{d}\vect{x} + \textrm{const.} 
\label{eq:bound}
\end{align}
In the derivation of the lower bound in \eqref{eq:bound},  we used $f(R(\vect{x})) > 0$ for any $x$ from line 2 to line 3. 
Based on the lower bound in \eqref{eq:bound}, given a set of data points $X=\{\vect{x}_i\}^N_{i=1}$ drawn from the proposal distribution $p_{\prop}(\vect{x})$,  maximizing the  weighted log-likelihood
\begin{align}
L( \vect{\theta}; X )  = \int  f\big(R(\vect{x})\big) p_{\textrm{prop}}(\vect{x}) \log p_{\vect{\theta}}(\vect{x} ) \textrm{d}\vect{x} \\
\approx \frac{1}{N} \sum_{ i=1 }^N f\big(R(\vect{x})\big) \log p_{\vect{\theta}}(\vect{x}_{i} ),
\label{eq:ml}
\end{align}
is equivalent to maximizing the lower bound of the objective function $J(\vect{\theta} )$ in \eqref{eq:surrogate_obj}.

To train the model $p_{\vect{\theta}}(\vect{x}|\vect{z})$, we also leverage the variational lower bound as in VAE~\citep{Kingma14}.
The variational lower bound on the marginal likelihood of data point $i$ is given by
\begin{align}
\log p_{\vect{\theta}}(\vect{x}_i) & \geq \mathcal{L}(\vect{\psi}, \vect{\theta};\vect{x}_i) \nonumber \\
& = - \KL\left( q_{\vect{\psi}}(\vect{z}|\vect{x}_i)|| p(\vect{z}) \right)  \nonumber \\
& \ \ \ + \E_{\vect{z} \sim q(\vect{z}|\vect{x}_i)} \left[ \log p_{\vect{\theta}}(\vect{x}_i | \vect{z}) \right]
\label{eq:elbo}
\end{align}
Using the variational lower bound in \eqref{eq:elbo}, the lower bound of the objective function in~\eqref{eq:ml} is given by
\begin{align}
L( \vect{\theta}; X ) \geq & \mathcal{L}(\vect{\psi}, \vect{\theta}; X) \\
 = & \sum^{N}_{i=1} f\big(R(\vect{x}_i)\big)\big( - \KL\left( q_{\vect{\psi}}(\vect{z}|\vect{x}_i)|| p(\vect{z}) \right)  \nonumber \\
&  + \log p_{\vect{\theta}}(\vect{x}_i | \vect{z}_i) \big).
\label{eq:vae_obj}
\end{align}
Therefore,   we can train $p_{\vect{\theta}}(\vect{x} | \vect{z})$ using the training procedure of VAE with importance weights $f\big(R(\vect{x})\big)$.
The above discussion indicates that we can leverage various techniques for training VAE in our framework.
In our implementation, we adapted the objective function proposed by \cite{Dupont18} using the importance weight, as follows:
\begin{align}
\tilde{\mathcal{L}}(\vect{\psi}, \vect{\theta}; X) 
 = \sum^{N}_{i=1} f\big(R(\vect{x}_i)\big) \ell(\vect{\theta}, \vect{\psi}),
\label{eq:loss}
\end{align}
where $\ell(\vect{\theta}, \vect{\psi})$ is given by
\begin{align}
\ell(\vect{\theta}, \vect{\psi})  =  \log p_{\vect{\theta}} (\vect{x}|\vect{z} )  
- \gamma \left| \KL\big( q_{\vect{\psi}}(\vect{z} | \vect{x}) || p(\vect{z}) \big) - C_{\vect{z}} \right|,
\label{eq:joint_loss}
\end{align}
where $C_{\vect{z}}$ is the information capacity of latent variable $\vect{z}$, and $\gamma$ is a coefficient.
The study by \cite{Dupont18} showed that this objective function encourages learning disentangled latent representations.

Although we showed how to maximize the lower bound of the surrogate objective function $J(\vect{\theta})$, one needs to be aware that our approach is based on \textit{amortized variational inference}~\citep{Cremer18} because training of the neural network is based on that of VAE.
In other words, our training procedure is amortized over the dataset instead of optimizing the output for each data point.
Therefore, the output of the trained model $p_{\vect{\theta}}(\vect{x}|\vect{z})$ may not be the exact solution to the optimization problem.
Thus, it may be necessary to fine-tune the output of $p_{\vect{\theta}}(\vect{x}|\vect{z})$ in practice, e.g., using a gradient-based method.
We use the shaping function $f(R(\vect{x})) = \exp(\alpha R(\vect{x}))$ in our implementation, and we deal with a scaling parameter $\alpha$ as a hyperparameter. 
We investigate the effect of the value of $\alpha$ in the experiments described in Section~\ref{sec:exp}.

\subsubsection{Connection to Density Estimation}
To describe the connection to the density estimation, we consider a distribution
\begin{equation}
p^{\target}( \vect{x} ) = \frac{ f \left( R(\vect{x}) \right)}{ Z },
\label{eq:target}
\end{equation}
which we refer to as the \textit{target distribution} and $Z$ is a partition function given by $Z=\int f \left( R(\vect{x}) \right) \intd \vect{x}$.
When $p^{\target}( \vect{x} )$ is followed, a sample with a higher score is drawn with a higher probability. 
The problem of estimating the density induced by the target distribution $p^{\target}( \vect{x} )$ can be formulated as
\begin{align}
\min_{\vect{\theta}} D_{\textrm{KL}}( p^{\target}(\vect{x}) || p_{\vect{\theta}}(\vect{x} ) ),
\label{eq:prob_KL}
\end{align}
where $ D_{\textrm{KL}}( p^{\target}(\vect{x}) || p_{\vect{\theta}}(\vect{x} ) )$ is the KL divergence. 
Given a set of samples $X=\{ \vect{x}_i \}^N_{i=1}$ drawn from the proposal distribution, the minimizer of $ D_{\textrm{KL}}( p^{\target}(\vect{x}) || p_{\vect{\theta}}(\vect{s}, \vect{\tau} ) )$ is given by the maximizer of the weighted log likelihood:
\begin{align}
L'( \vect{\theta} )  = \int  W(\vect{x}) p_{\textrm{prop}}(\vect{x}) \log p_{\vect{\theta}}(\vect{x} ) \textrm{d}\vect{x} \\
\approx \frac{1}{N} \sum_{ i=1 }^N W(\vect{x}_{i}) \log p_{\vect{\theta}}(\vect{x}_{i} ),
\label{eq:ml2}
\end{align}
where $W(\vect{x})$ is the importance weight given by
\begin{align}
W(\vect{x}) = \frac{p^{\target}( \vect{x} )}{ p_{\prop}(\vect{x}) }.
\label{eq:weight}
\end{align}
$L( \vect{\theta} )$ in \eqref{eq:ml} and $L'( \vect{\theta} )$ in \eqref{eq:ml2} are equivalent if $ p_{\textrm{prop}}(\vect{x})$ is the uniform distribution.
Therefore, if $ p_{\textrm{prop}}(\vect{x})$ is a uniform distribution, then our approach is equivalent to estimating the density induced by $p^{\target}( \vect{x} )$.
Furthermore, if the shaping function is given by the exponential function as $f(\cdot)=\exp(\cdot)$, then the target distribution can be regarded as the Boltzmann distribution~\citep{LeCun06}.

\section{Motion Planning by Learning the Solution Manifold in Trajectory Optimization}
When applying LSMO to motion planning, we can employ various trajectory representations, including a waypoint representation as in previous studies~\citep{Zucker13}.
However, for solving motion planning problems efficiently, it is effective to employ a structured trajectory representation that incorporates the desired property of a trajectory.
In this section, we present a trajectory representation and an exploration strategy that makes LSMO sample-efficient in motion planning.
 
\subsection{Trajectory Representation}
\label{sec:ntp}
Incorporating the desired property of the trajectory into the trajectory parameterization is essential to reducing the computational cost of training a neural network.
For this purpose, we parameterize a trajectory using the following form:
\begin{align}
\vect{\xi}_{\vect{\theta}}(\vect{q}_0, \vect{q}_T; \vect{w}) = \vect{\xi}^{\base}(\vect{q}_0, \vect{q}_T) +  \vect{F} \vect{\xi}^{\residual}_\vect{w},
\label{eq:ntp_model}
\end{align}
where $\vect{\xi}^{\base}(\vect{q}_0, \vect{q}_T)$ is the baseline trajectory, $\vect{\xi}^{\residual}_\vect{w}$ is the term parameterized with a vector $\vect{w}$, and $\vect{F}$ is a time-dependent scaling matrix given by a diagonal matrix defined as
\begin{align}
\vect{F} = 
\left[
\begin{array}{cccc}
s(t_0) & 0 &\ldots & 0 \\
0 & \ddots & \ddots & \vdots\\
\vdots &\ddots  & \ddots & 0 \\
0 & \ldots & 0 & s(t_T) 
\end{array}
\right],
\end{align}
where $s(t)$ is a time-dependent scaling function designed to satisfy $s(0)=s(1)=0$. In our implementation, $s(t)$ is given by
\begin{align}
s(t) = 
\left\{
\begin{array}{ll}
at & \textrm{if} \ 0 \leq  t < \epsilon \\
1 &  \textrm{if} \ \epsilon \leq  t < 1-\epsilon\\
1 - at &  \textrm{if} \ 1 -\epsilon \leq  t \leq  1
\end{array}
\right. ,
\end{align}
where $a$ and $\epsilon$ are constants.
Using this parameterization, we can guarantee that the trajectories obtained from $p(\vect{\xi}|\vect{z})$ satisfy the constraints of the start and goal configurations.
The residual term $\vect{\xi}^{\residual}$ is represented by a linear combination of the basis function.
\begin{align}
\vect{\xi}^{\residual}_\vect{w} = \vect{\Phi} \vect{w},
\label{eq:ntp_residual}
\end{align}
where $\vect{\Phi} \in \Real^{T \times B}$ is a feature matrix, $\vect{w} \in \Real^{B \times D}$ is the weight matrix, and $B$ is the number of basis functions.
In our framework, the neural network is trained to output weight matrix $\vect{w}$.
The $i$ th column of the feature matrix $\Phi(t)$ is given by the basis function
\begin{align}
b^i_{\textrm{log}}(t)  = \frac{1}{1 + \exp\big( \alpha(t -c_i)  \big)},
\label{eq:logistic_basis}
\end{align}
where $\alpha$ defines the slope of the function, and $c_i$ defines the center of the $i$th basis function.

\begin{figure}[]
	\centering
	\begin{subfigure}[t]{0.48\columnwidth}
		\includegraphics[width=\textwidth]{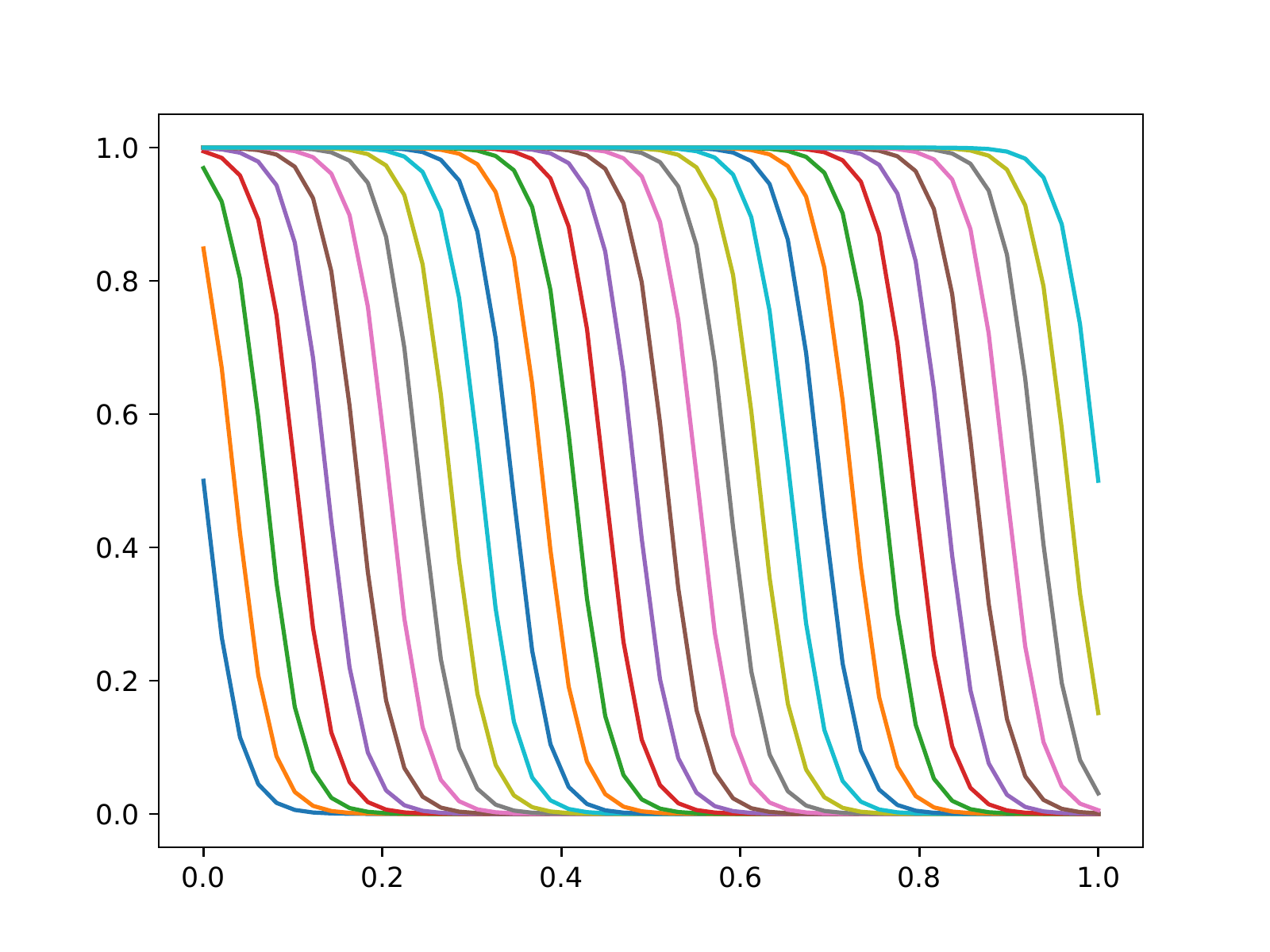}
		\caption{Basis functions $b_{\textrm{log}}(t)$ with $\alpha=50$.  }
	\end{subfigure}
	\begin{subfigure}[t]{0.48\columnwidth}
		\includegraphics[width=\textwidth]{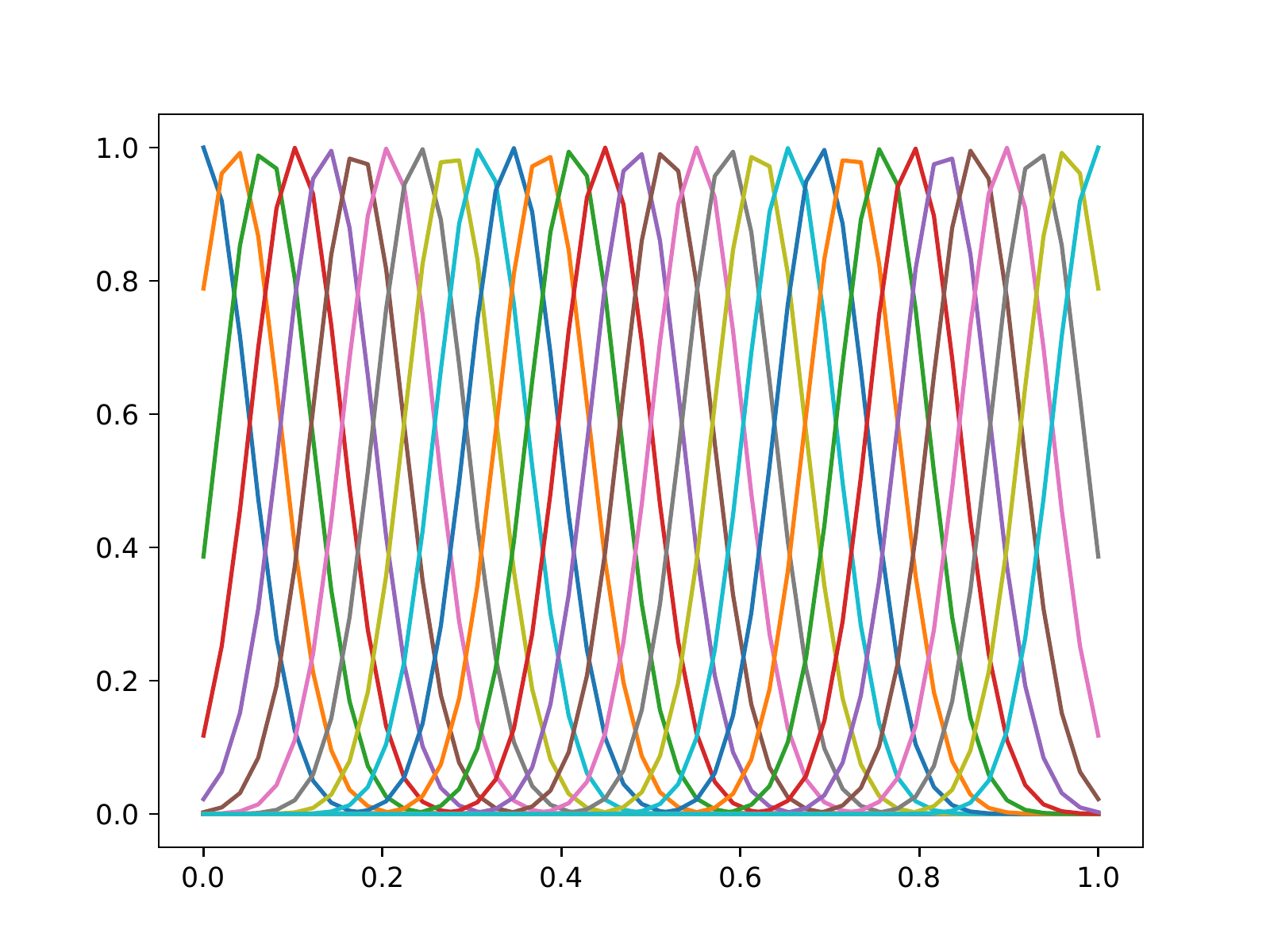}
		\caption{Basis functions $b_{\textrm{exp}}(t)$ with $h = 0.01$ }
	\end{subfigure}
	\caption{Visualization of basis functions. The number of basis functions is 30.}
	\label{fig:basis_func}
\end{figure}

A popular choice of the basis function for movement primitives is an exponential function given by
\begin{align}
b^i_{\textrm{exp}}(t) = \exp\left( \frac{- (t -c_i )^2}{h}  \right),
\label{eq:exp_basis}
\end{align}
where $h$ defines the band width, and $c_i$ defines the center of the basis function.
The basis functions in \eqref{eq:logistic_basis} and \eqref{eq:exp_basis} are shown in Fig.~\ref{fig:basis_func}.
Although the form in \eqref{eq:exp_basis} is popular, we use the basis function in \eqref{eq:logistic_basis} in our implementation because it leads to a smaller condition number of the feature matrix $\vect{\Phi}$.
Although our framework is not limited to a specific form of the basis function, it is important to select appropriate parameters of the basis functions to obtain a feature matrix with a low condition number.
For a discussion of the condition number of feature matrix $\vect{\Phi}$, please refer to the Appendix.

For convenience, we refer to the trajectory representation based on \eqref{eq:ntp_model}-\eqref{eq:ntp_residual} as Residual Trajectory Primitives~(RTPs) hereinafter.
If we employ the waypoint representation, the resulting path can be jerky, particularly when a path involves a longer detour. 
By contrast, parameterization based on RTPs can ensure the smoothness of the trajectory as well as the constraints of the start and goal configurations.
As described in the experimental section, the parameterization based on RTP can reduce the time for training the neural network for motion planning tasks.

\subsection{Proposal Distribution}
In our framework, it is essential to employ an appropriate proposal distribution for solving motion planning problems efficiently.
In this work, we use the proposal distribution proposed by \cite{Kalakrishnan11}:
\begin{align}
\beta_{\textrm{traj}}(\vect{\xi}) = \mathcal{N}(\vect{\xi}_{\textrm{base}}, a\Sigma),
\label{eq:exploration_traj}
\end{align}	
where $a$ is a constant and $\Sigma$ is the covariance matrix given by the inverse of the matrix $\vect{A}^{\top}\vect{A}$ and $\vect{A}$ is defined as
\begin{align}
\vect{A} & = 
\left[
\begin{array}{ccccccc}
   2 & -1 & 0 & \dots & &0\\ 
 -1 & 2 & -1 & &  &\\
 0 & -1 & & \ddots & & \vdots\\
 \vdots & & \ddots & \ddots & -1 & 0\\
  & & & -1 & 2  &- 1 \\
 0 & \dots & & 0 & -1 & 2
\end{array}
\right].
\label{eq:metric}
\end{align}
The mean of the proposal distribution $\vect{\xi}_{\textrm{base}}$ is the baseline trajectory, and we define the baseline trajectory as the trajectory obtained by linearly interpolating the given start and goal configurations.
Previous studies showed that the exploration based on this sampling strategy works well in the context of trajectory optimization~\citep{Kalakrishnan11} and inverse reinforcement learning~\citep{Kalakrishnan13}.

\begin{figure}[tb]
	\centering
	\includegraphics[width=\columnwidth]{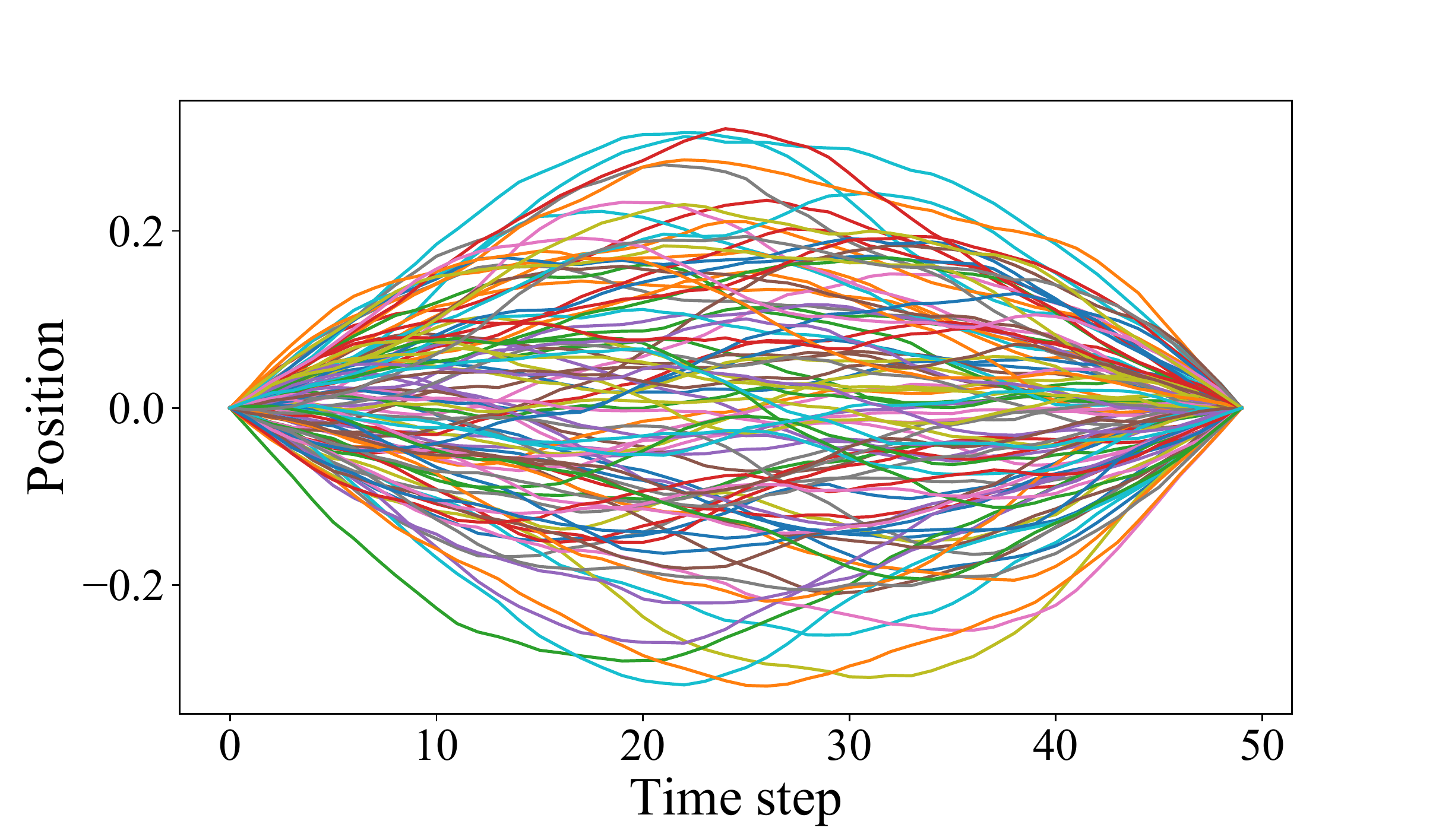}
	\caption{Noise sampled from $\beta_{\textrm{traj}}(\vect{\xi})$.  }
	\label{fig:expl_traj}
\end{figure}

\subsection{Projection onto Constraint Solution Space}
Although LSMO can learn the distribution of solutions, there is no guarantee that the output of $p_{\vect{\theta}}(\vect{\xi}|\vect{z})$ satisfies the desired constraint such as joint limits.
For this reason, we fine-tune the output of $p_{\vect{\theta}}(\vect{\xi}|\vect{z})$ by using CHOMP~\citep{Zucker13}.
The update rule of CHOMP is given by:
\begin{align}
\vect{\xi}^{*} = \arg \min_{\vect{\xi}} \left\{  \mathcal{C}(\vect{\xi}^c) + \vect{g}^{\top} (\vect{\xi} - \vect{\xi}^c) + \frac{\eta}{2} \left\| \vect{\xi} - \vect{\xi}^c \right\|^2_{M}  \right\},
\label{eq:chomp}
\end{align}
where $\vect{g} = \nabla \mathcal{C}(\vect{\xi})$, $\vect{\xi}^c$ is the current plan of the trajectory, 
$\eta$ is a regularization constant, 
and  $\left\| \vect{\xi} \right\|^2_{M}$ is the norm defined by a matrix $M$ as $\left\| \vect{\xi} \right\|^2_{M} = \vect{\xi}^{\top}M\vect{\xi}$.

The third term on the right-hand side of \eqref{eq:chomp} can be interpreted as the trust region;
it penalizes the change in the velocity profile of the motion when $\vect{M}=\vect{A}$ and $\vect{A}$ is given by \eqref{eq:metric}.
By minimizing the cost function regularized with this penalty, we can obtain a trajectory that minimizes the collision cost with a minimal change in the velocity profile from the initial trajectory.
In our framework, the generative model outputs trajectories with different velocity profiles for different values of the latent variable $\vect{z}$.
To maintain the diversity of the solutions, the change in the velocity profile before and after the fine-tuning of the trajectory must be minimized.
Hence, the properties of CHOMP are suitable for our framework.
Although the output of the deep generative model is fine-tuned using CHOMP in our implementation, we can also employ other existing methods such as GPMP~\citep{Mukadam18}.

\begin{algorithm}[t]
	\caption{Motion Planning by Learning the Solution Manifold in Trajectory Optimization (MPSM) }
	\begin{algorithmic}[1]
		\STATEx{\textbf{Input:} start configuration $\vect{q}_0$, goal configuration $\vect{q}_T$  
		}
		\STATEx{\textbf{Training phase:}}
		\STATE{Initialize the trajectory, e.g., linear interpolation between $\vect{q}_0$ and $\vect{q}_T$}
		\STATE{Generate $N$ synthetic samples $\{\vect{\xi}_i\}^N_{i=1}$ from $\beta_{\textrm{traj}}(\vect{\xi})$ in \eqref{eq:exploration_traj}  }
		\STATE{Evaluate the objective function $R(\vect{\xi}_i)$ and compute the weight $W(\vect{\xi}_i)$ for $i=1,\ldots, N$  }
		\STATE{Convert the trajectory $\vect{\xi}$ into the trajectory parameter $\vect{w}$ using RTP}
		\STATE{Train $p_{\vect{\theta}}( \vect{w} |\vect{z})$ by maximizing  $\mathcal{L}(\vect{\theta}, \vect{\psi})$ in \eqref{eq:loss} }
		\STATEx{\textbf{Generation phase:}}
		\STATE{Generate  $\vect{w}$ with $p_{\vect{\theta}} (w|\vect{z})$ by specifying the value of $\vect{z}$}
		\STATE{Reconstruct a trajectory $\vect{\xi}$ from $\vect{w}$ using the trajectory parameterization in Section~\ref{sec:ntp}}
		\IF{the trajectory $\vect{\xi}$ is not collision-free }
		\STATE{Fine-tune $\vect{\xi}$, e.g., using CHOMP \eqref{eq:chomp}  }
		\ENDIF
		\STATEx{\textbf{Return:}  planned trajectory $\vect{\xi}$ }
	\end{algorithmic}
	\label{alg:MPLSM}
\end{algorithm}

\subsection{Summary of the Proposed Motion Planning Algorithm}

We summarize the motion planning algorithm based on LSMO in Algorithm~\ref{alg:MPLSM}.
In the training phase,  we sample trajectories by following the proposal distribution~$\beta_{\textrm{traj}}(\vect{\xi})$ in \eqref{eq:exploration_traj}.
Subsequently, the costs of the sampled trajectories are evaluated, and the trajectories are parameterized based on the RTP. 
The generative model $p_{\vect{\theta}} (\vect{w}|\vect{z})$ is trained to maximize  $\mathcal{L}(\vect{\theta}, \vect{\psi})$ in \eqref{eq:loss}.
In the generation phase, the trained model  $p_{\vect{\theta}} (\vect{w}|\vect{z})$ outputs the trajectory parameter $\vect{w}$ for a given value of the latent variable $\vect{z}$.
A trajectory in configuration space $\vect{\xi}$ is then recovered from $\vect{w}$.
If the recovered trajectory is not collision-free, then the trajectory $\vect{\xi}$ is fine-tuned with CHOMP~\citep{Zucker13}.
We refer to this algorithm for motion planning as \textit{Motion Planning by Learning the Solution Manifold in Trajectory Optimization ~(MPSM)}.


\begin{figure*}
	\centering
	\begin{subfigure}[t]{0.5\columnwidth}
		\centering
		\includegraphics[width=\textwidth]{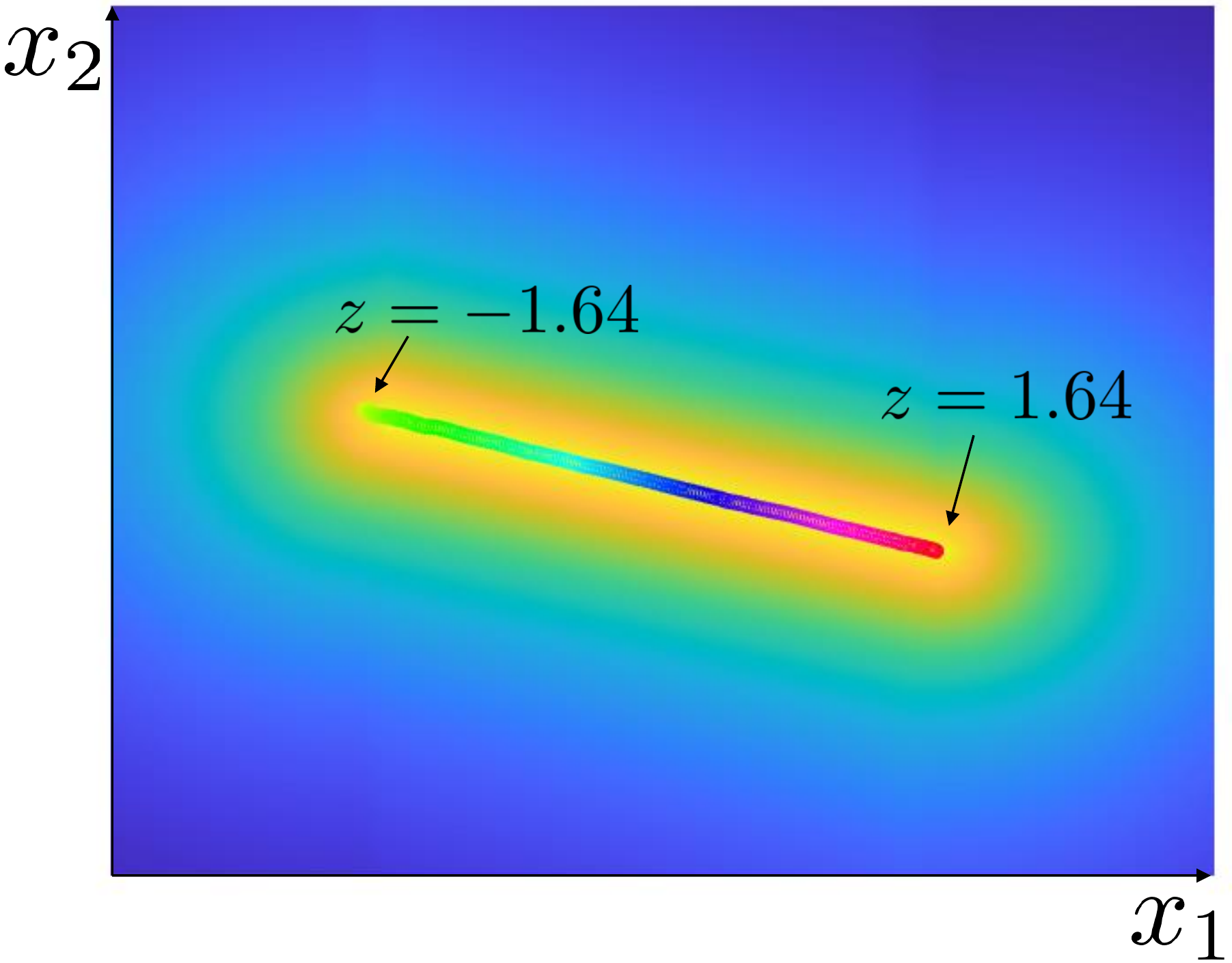}
		\caption{LSMO without fine-tuning on Func.~1. }
	\end{subfigure}
	\begin{subfigure}[t]{0.5\columnwidth}
		\centering
		\includegraphics[width=\textwidth]{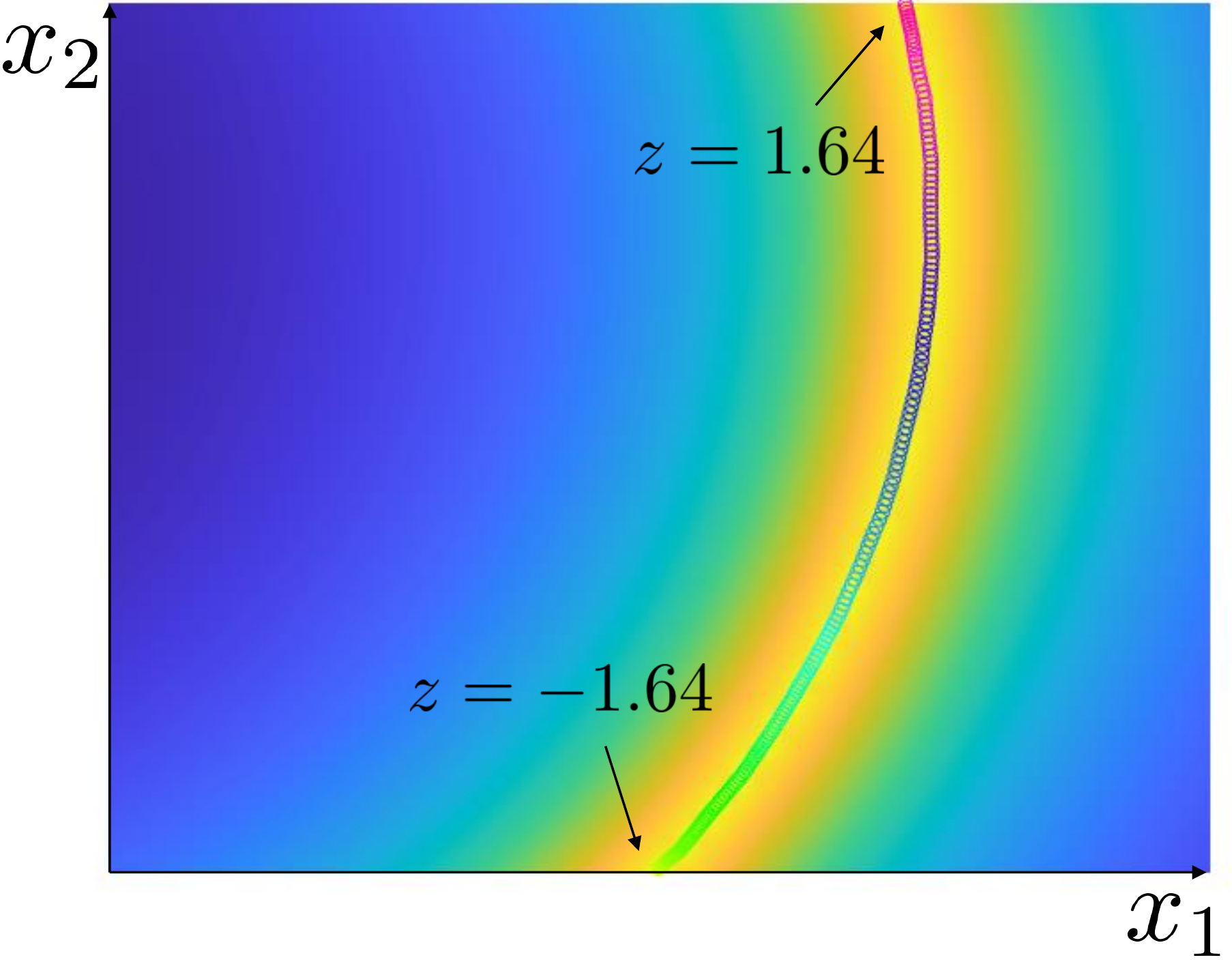}
		\caption{LSMO without fine-tuning on Func.~2. }
	\end{subfigure}
	\begin{subfigure}[t]{0.5\columnwidth}
		\centering
		\includegraphics[width=\textwidth]{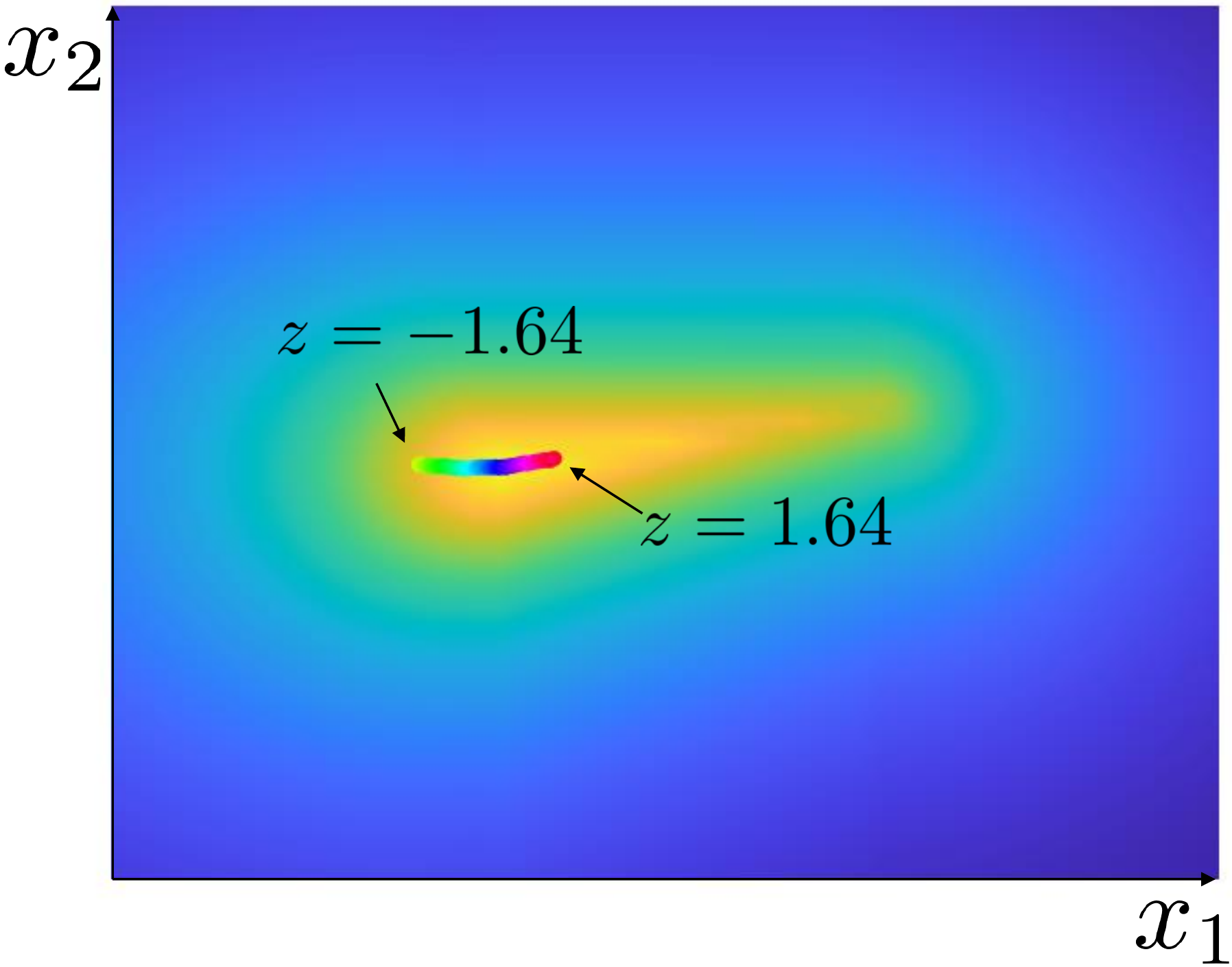}
		\caption{LSMO without fine-tuning on Func.~3. }
	\end{subfigure}
	\begin{subfigure}[t]{0.5\columnwidth}
		\centering
		\includegraphics[width=\textwidth]{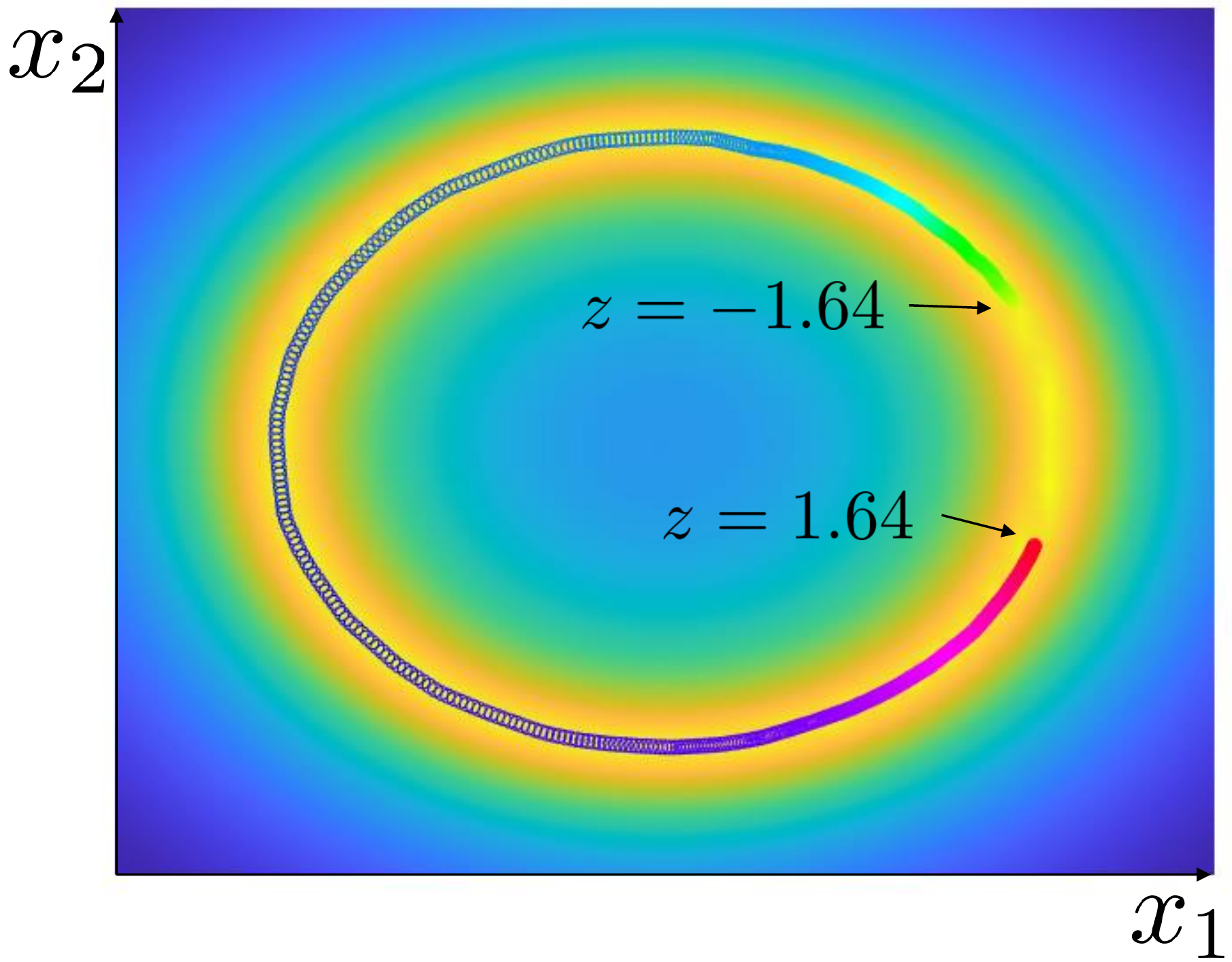}
		\caption{LSMOwithout fine-tuning  on Func.~4. }
	\end{subfigure}
	\begin{subfigure}[t]{0.5\columnwidth}
		\centering
		\includegraphics[width=\textwidth]{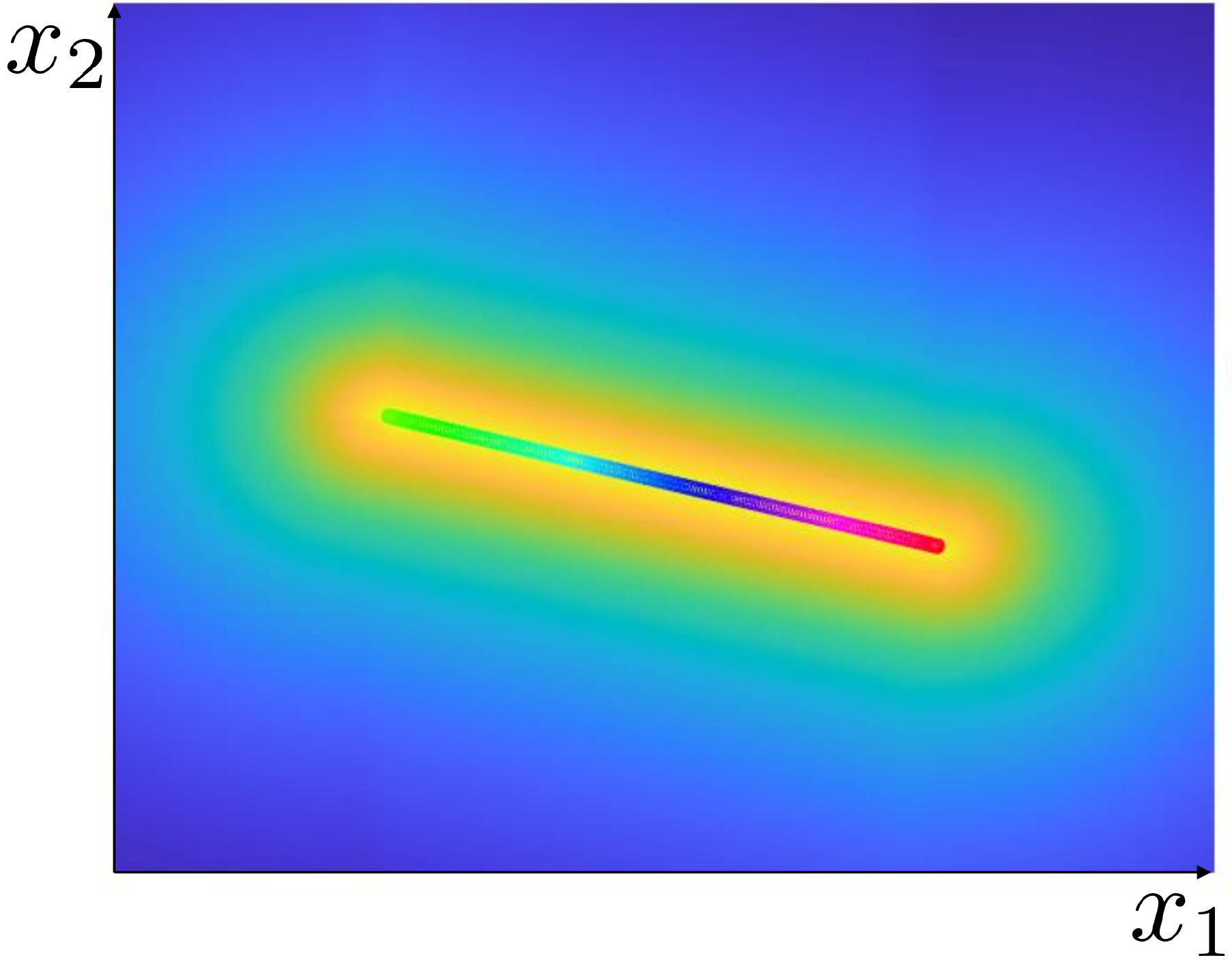}
		\caption{LSMO with fine-tuning on Func.~1. }
	\end{subfigure}
	\begin{subfigure}[t]{0.5\columnwidth}
		\centering
		\includegraphics[width=\textwidth]{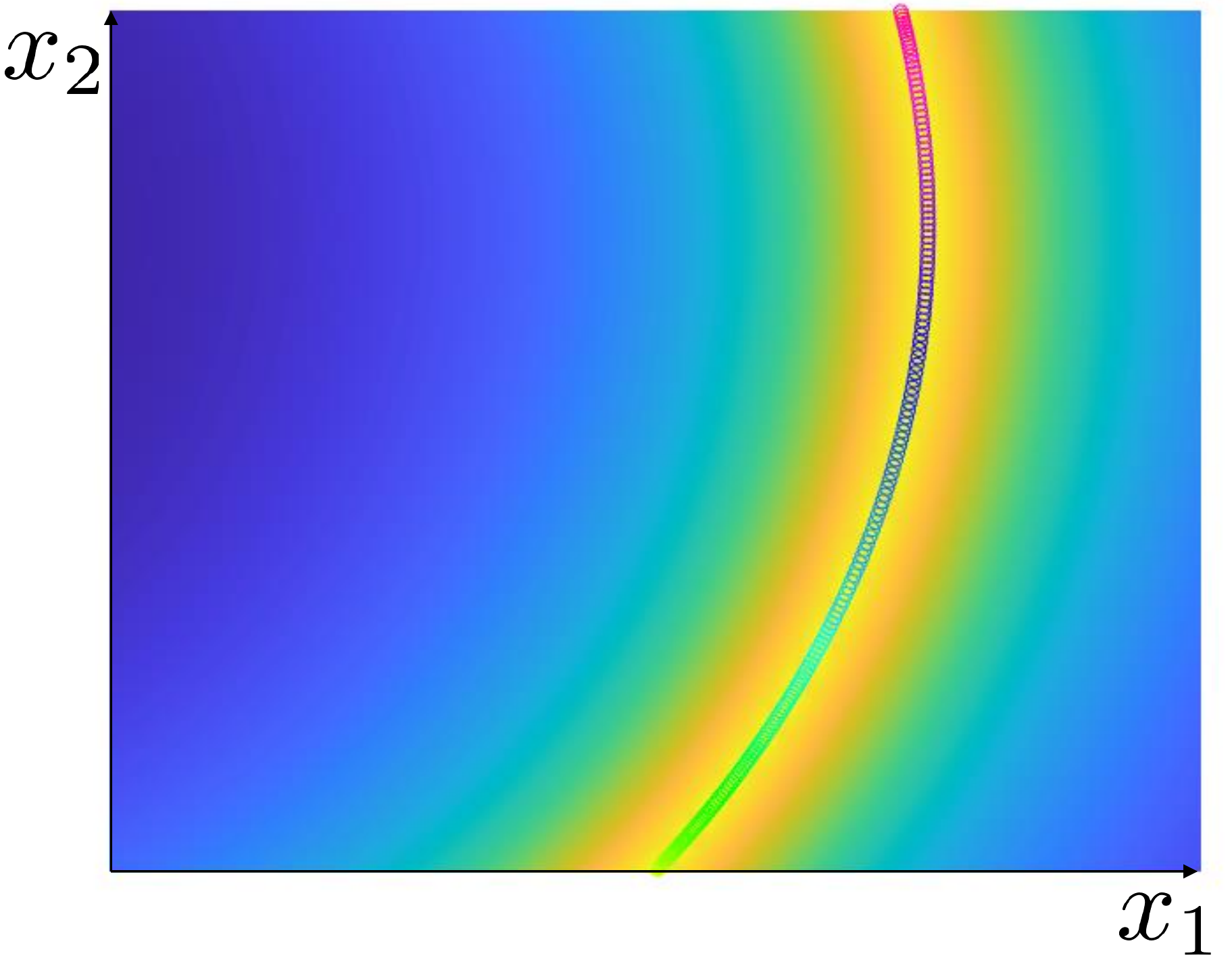}
		\caption{LSMO with fine-tuning on Func.~2. }
	\end{subfigure}
	\begin{subfigure}[t]{0.5\columnwidth}
		\centering
		\includegraphics[width=\textwidth]{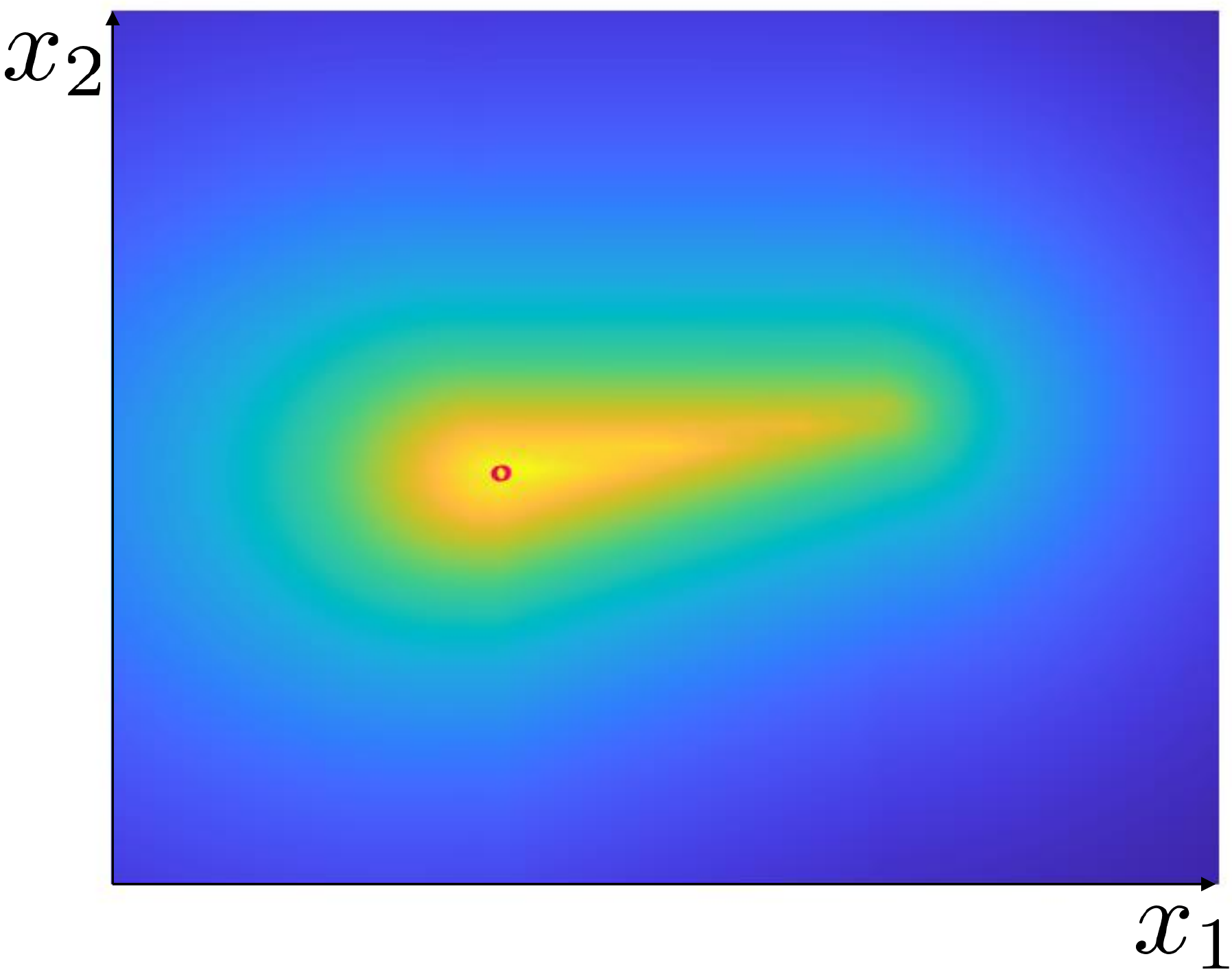}
		\caption{LSMO with fine-tuning on Func.~3. }
	\end{subfigure}
	\begin{subfigure}[t]{0.5\columnwidth}
		\centering
		\includegraphics[width=\textwidth]{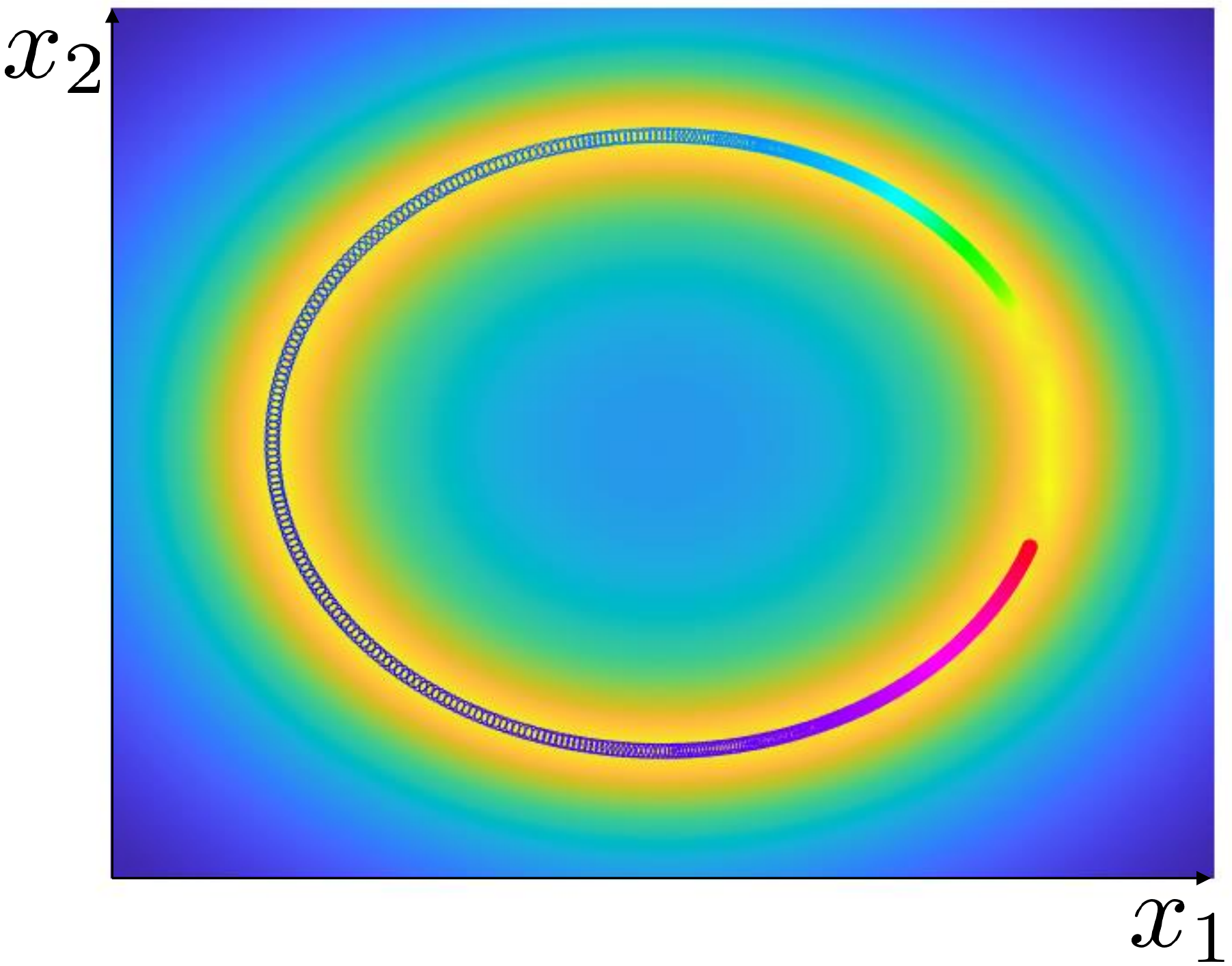}
		\caption{LSMOwith fine-tuning  on Func.~4. }
	\end{subfigure}
	\begin{subfigure}[t]{0.5\columnwidth}
		\centering
		\includegraphics[width=\textwidth]{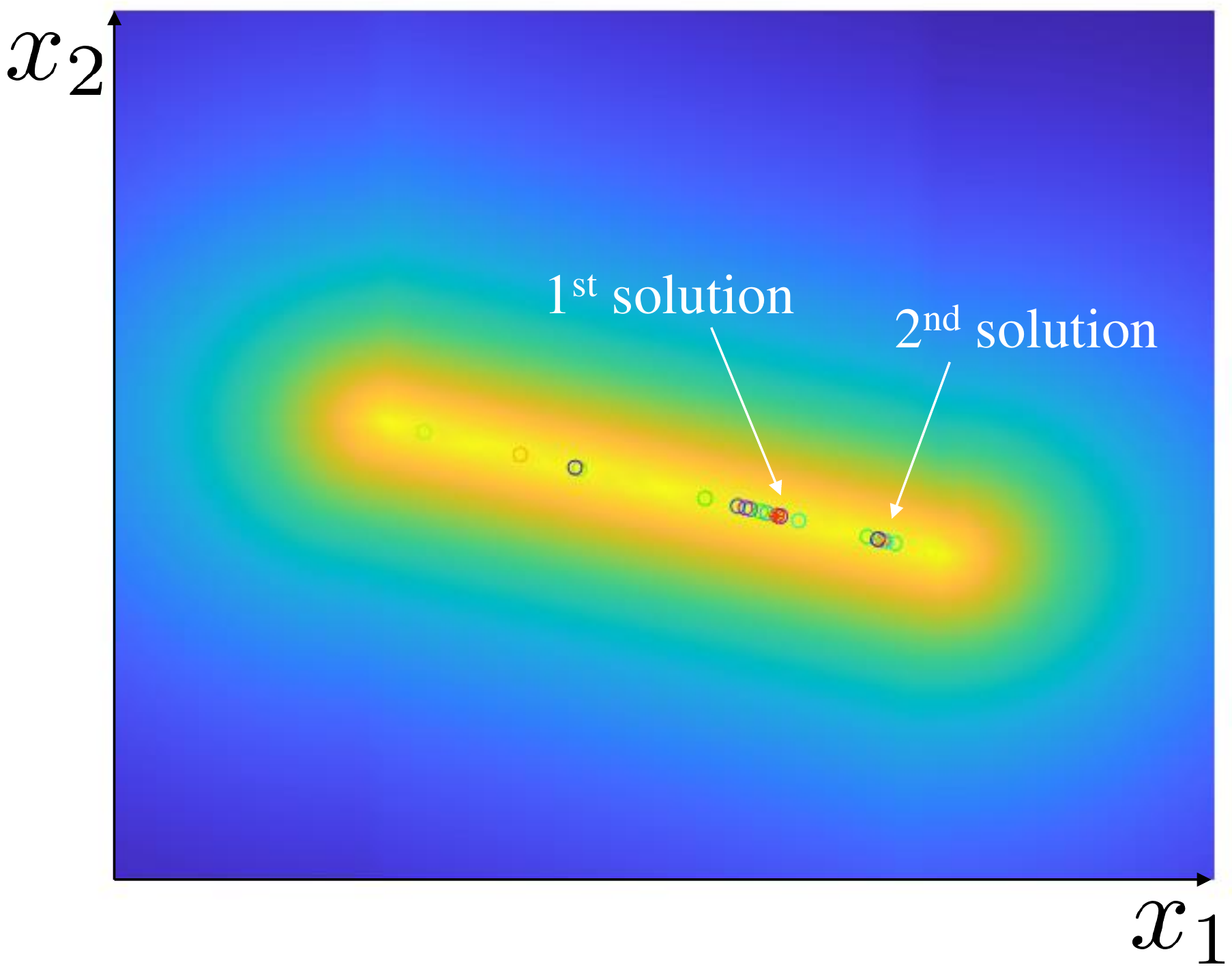}
		\caption{CEM on Func.~1. }
	\end{subfigure}
	\begin{subfigure}[t]{0.5\columnwidth}
		\centering
		\includegraphics[width=\textwidth]{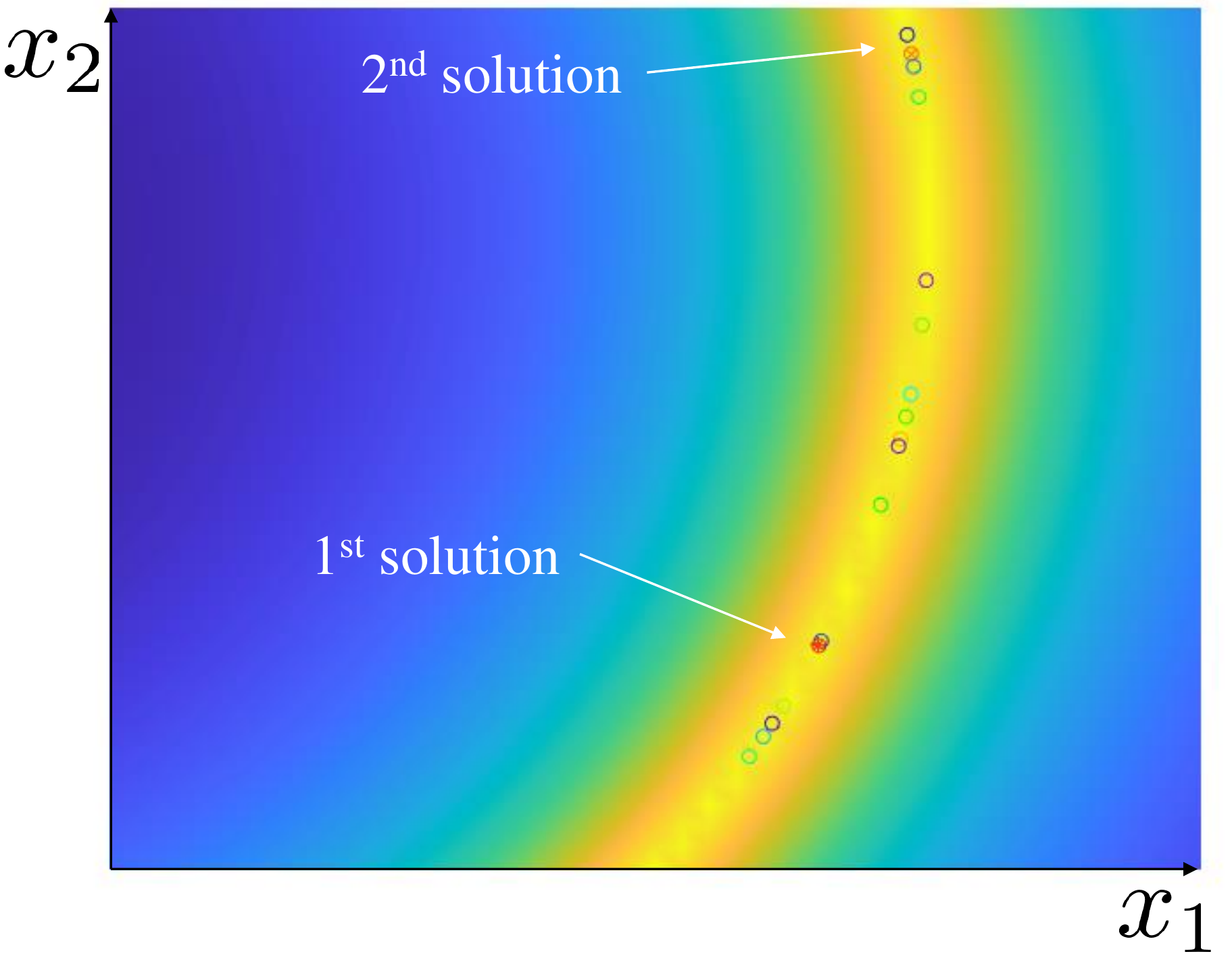}
		\caption{CEM on Func.~3. }
	\end{subfigure}
	\begin{subfigure}[t]{0.5\columnwidth}
		\centering
		\includegraphics[width=\textwidth]{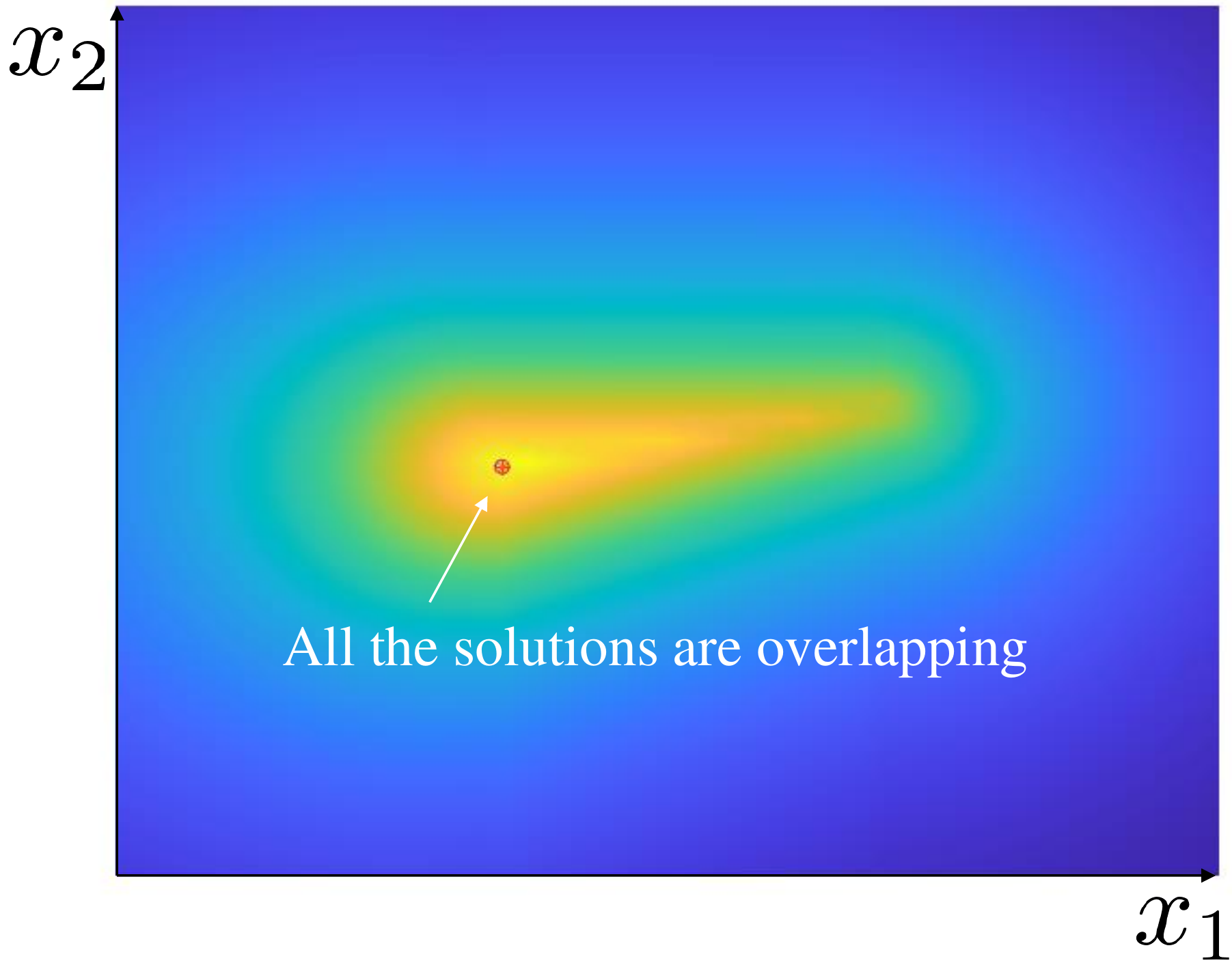}
		\caption{CEM on Func.~3. }
	\end{subfigure}
	\begin{subfigure}[t]{0.5\columnwidth}
		\centering
		\includegraphics[width=\textwidth]{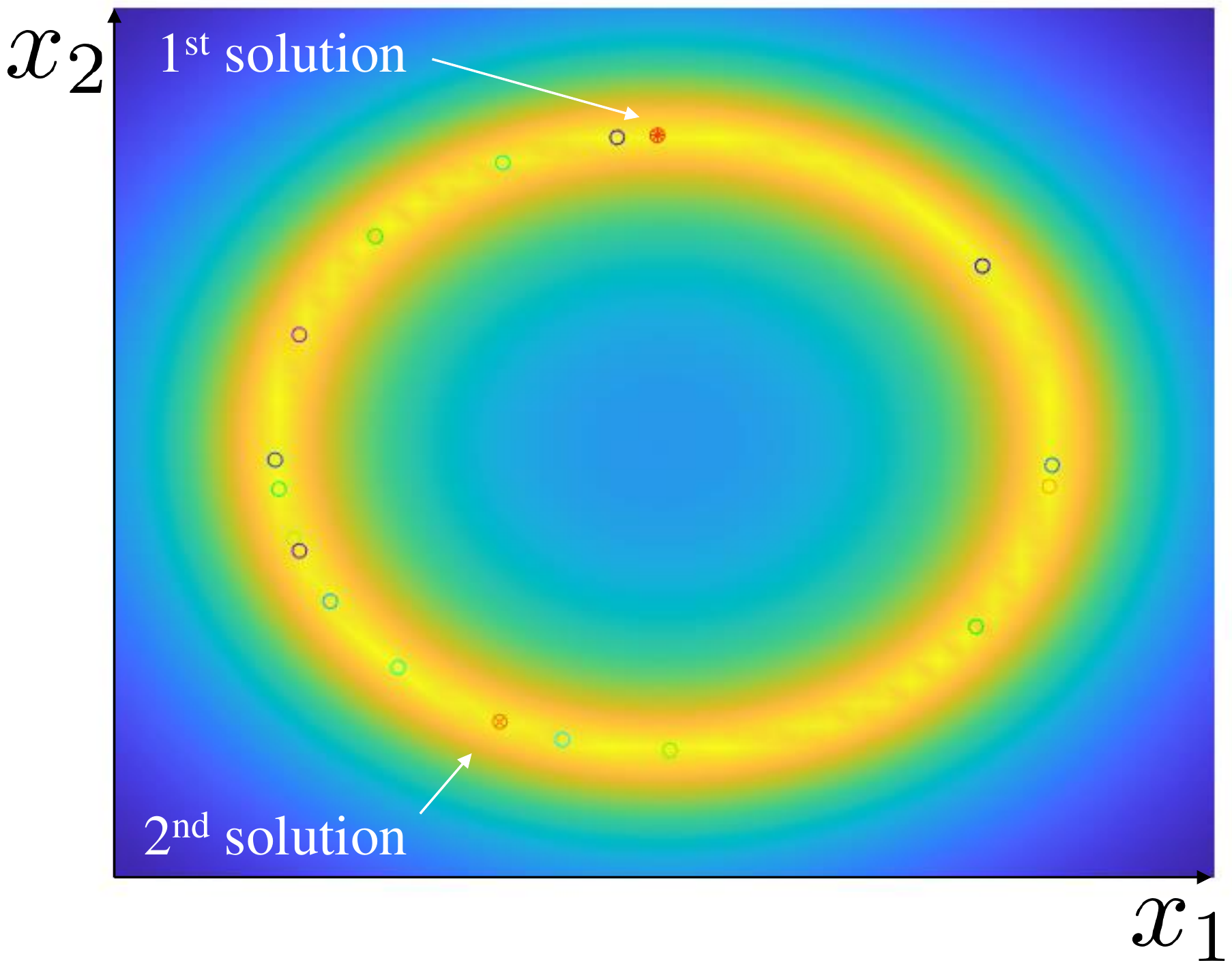}
		\caption{CEM on Func.~4. }
	\end{subfigure}
	\caption{Behavior of LSMO when optimizing the test objective function. The warmer color represents a higher value of the objective function. In (a)-(h), circles represent the outputs of a model trained with LSMO, and color of circle indicates value of the latent variable. Outputs of the trained model continuously change by continuously changing the value of the latent variable.
	Outputs of the model are generated by linearly changing value of $z$ in $[-1.64, 1.64]$.  In (e)-(h), centers of Gaussian distributions are drawn as circles.
	} 
	\label{fig:test}
\end{figure*}

\begin{table*}
	\caption{Values of the objective function for the obtained solutions (mean $\pm$ standard deviation)}
	\centering
	\begin{tabular}{lllll}
		\toprule
		& Func.~1     & Func.~2 & Func.~3 & Func.~4 \\
		\midrule
		LSMO w/o fine-tuning &  $0.990 \pm 0.021 $  & $0.994 \pm 0.0059$ & $0.889\pm0.0897$ &  $0.973 \pm  0.0346$   \\
		LSMO w/ fine-tuning &  $ 1.000 \pm 1.2\times10^{-5}$  & $1.000 \pm 4.1\times10^{-6}$ & $0.9991\pm6.1\times 10^{-4}$ &  $1.0000 \pm 3.7\times 10^{-5}$   \\
		CEM     &$ 1.000 \pm 7.1\times10^{-6}$       &$ 1.000 \pm 2.5\times10^{-5}$  & $ 0.9999 \pm 5.0\times 10^{-5}$ & $ 0.9998 \pm 3.1\times 10^{-3}$ \\
		\bottomrule
	\end{tabular}
	\label{tbl:test}
\end{table*}

\section{Evaluation with synthetic test functions}
To evaluate the capability of LSMO to capture the solution manifold in optimization, we applied LSMO to optimization problems for synthetic test functions.
In this evaluation, we trained a neural network using LSMO, and the output of the trained model was fine-tuned.
To fine-tune the solutions, we employed the cross-entropy method~(CEM) with a Gaussian distribution. 
To achieve a trust-region-based update such as CHOMP, the objective function for CEM at the $k$th iteration is given by 
\begin{align}
R'(\vect{x}) = R(\vect{x}) - \eta_1 \sqrt{||\vect{x} - \vect{\mu}_k||^2_2}
\end{align}
where $R(\vect{x})$ is the test function, $\eta_1$ is a coefficient, and $\vect{\mu}_k$ is the center of the Gaussian distribution used for sampling at the $k$th iteration of CEM.
The second term penalizes the deviation from the initial estimation, which is necessary to maintain the diversity of solutions captured by LSMO.
The conditions for training the neural network in LSMO are provided in the Appendix.

For the shaping function, we used the following form in the implementation:
\begin{align}
f(R(\vect{x})) = 
\left\{
\begin{array}{cll}
\exp \left( \frac{ \alpha \big( R(\vect{x}) - R_{\textrm{max}} \big) }{R_{\textrm{max}} - R_{\textrm{med}} } \right)  & \textrm{if} & R(\vect{x}) \geq R_{\textrm{med}} \\
0 & \textrm{if} & R(\vect{x}) < R_{\textrm{med}} 
\end{array}
\right. 
\label{eq:shape}
\end{align}
where $R_{\textrm{max}}$ and $R_{\textrm{med}}$ are the maximum and median values, respectively, among the samples drawn from the proposal distribution.
We set $\alpha=10$ in this experiment.

For visualization, we used the test functions that take in two-dimensional inputs and outputs the one-dimensional value.
As a baseline method, we applied CEM with a multimodal sampling distribution.
In this baseline method, a mixture of 20 Gaussian distributions is used as the sampling distribution.
The conditions for training the neural network are summarized in Table~\ref{tbl:param_test}.
These test functions are designed such that they have an infinite set of solutions, and the maximum and minimum values are approximately 1 and 0, respectively.
Detailed definitions of the test functions are provided in the Appendix~\ref{app:test_func}.

The outputs of the model $p_{\vect{\theta}}(\vect{x}|\vect{z})$ trained with LSMO and the solutions obtained by CEM are illustrated in Fig.~\ref{fig:test}. 
The outputs of $p_{\vect{\theta}}(\vect{x}|z)$ are generated by linearly changing the value of $z$ in $[-1.64, 1.64]$.
This value of $z$ is chosen because $P(z< -1.64)= 0.05$ and $P(z< 1.64)= 0.95$ when $z \sim \mathcal{N}(0, 1)$. 
In Fig.~\ref{fig:test}(a)-(h), circles represent the output of the model trained with LSMO, and the color of the circle indicates the value of the latent variable. 
It is evident that samples drawn from the model $p_{\vect{\theta}}(\vect{x}|\vect{z})$ correspond to the region of optimal solutions of the objective function. 
The output of the trained model continuously changes by continuously changing the value of the latent variable.

While CEM finds multiple solutions for objective functions, the number of solutions needs to be manually specified. 
Moreover, the similarity of the obtained solutions is not indicated by CEM.
As shown in Figure~\ref{fig:test}(i)-(l), the solutions found by CEM that correspond to the first and second components of GMMs are separated from each other.
Although there are many black-box methods for multimodal optimization, they have a common property of CEM: the similarity of solutions is not indicated, and the user would need to examine all the solutions to find the most preferable one.
It is possible to find 100 solutions using CEM for test functions in these experiments, but it is be tedious for the user to check all of them.
By contrast, an infinite set of solutions are modeled with a neural network, and the similarity of solutions is indicated by the value of the latent variable in LSMO.
Using LSMO, the user can intuitively examine the various solutions by changing the value of the latent variable.

A limitation of LSMO indicated by this experiment is that the manifold learned by LSMO may not capture all variations of solutions, although it is clear that LSMO can capture more diverse solutions than CEM.
For example, although the model trained by LSMO captures the various solutions in Fig.~\ref{fig:test}(d), there is a region of optimal points which is not covered by the outputs of the model.

The scores of the output of the model $p_{\vect{\theta}}(\vect{x}|\vect{z})$ are summarized in Table~\ref{tbl:test}. 
The result indicates that the output of the model trained with LSMO does not necessarily correspond to an exact solution. 
For example, although the test function shown in Fig.~\ref{fig:test}(c) actually has a unique solution, the model trained with LSMO learns the manifold corresponding to the direction in which the gradient of the test function is gradual.
As a result, the solutions before fine-tuning are not as accurate as those found by CEM. 
This result is natural because the output of the trained model is the result of amortized variational inference and not the optimization for each point~\citep{Kim18,Cremer18}. 
Therefore, fine-tuning the output of $p_{\vect{\theta}}(\vect{x}|\vect{z})$ is necessary to obtain exact solutions.
Figure~\ref{fig:test} and Table~\ref{tbl:test} show that we can obtain accurate and diverse solutions by fine-tuning the output of $p_{\vect{\theta}}(\vect{x}|\vect{z})$.
The process of fine-tuning is approximately 0.1 s for each solution, and we think that it is negligible in practice.

\section{Evaluation of motion planning tasks}
\label{sec:exp}
\subsection{Implementation Details}

In motion planning tasks, we used the objective function given by $R(\vect{\xi}) = - \mathcal{C}(\vect{\xi})$, where $\mathcal{C}(\vect{\xi})$ is the cost function used in previous studies on trajectory optimization~\citep{Zucker13} given by  
\begin{align}
\mathcal{C}(\vect{\xi}) = c_{\textrm{obs} }(\vect{\xi}) + \alpha c_{\textrm{smoothenss} }(\vect{\xi}).
\label{eq:mp_obj}
\end{align}
The first term in \eqref{eq:mp_obj}, $c_{\textrm{obs} }(\vect{\xi})$, is the penalty for collision with obstacles.
Given a configuration $\vect{q}$, we denote by $\vect{x}_{u}(\vect{q}) \in \Real^3$ the position of the bodypoint $u$ in the task space. $c_{\textrm{obs} }(\vect{\xi})$ is then given by
\begin{equation}
c_{\textrm{obs} }(\vect{\xi}) = \frac{1}{2} \sum_{t} \sum_{u \in \mathcal{B} } c \left( \vect{x}_{u}(\vect{q}_t) \right)  \left\| \frac{d}{dt}\vect{x}_{u}(\vect{q}_t) \right\|,
\end{equation}
and $\mathcal{B}$ is a set of body points that comprise the robot body. The local collision cost function $c(\vect{x}_{u})$ is defined as
\begin{equation}
c ( \vect{x}_{u} ) =
\left\{
\begin{array}{cll}
0 , & \textrm{if} & d(\vect{x}_{u}) >  \epsilon, \\
\frac{1}{2  \epsilon} (d (\vect{x}_{u}) -   \epsilon)^{2} , & \textrm{if} & 0 < d(\vect{x}_{u}) < \epsilon, \\
- d(\vect{x}_{u})+\frac{1}{2}\epsilon,& \textrm{if} & d(\vect{x}_{u})< 0,
\end{array}
\right. 
\end{equation}
where $\epsilon$ is the constant that defines the margin from the obstacle, and $d (\vect{x}_{u})$ is the shortest distance in task space between the bodypoint $u$ and obstacles.
The second term in \eqref{eq:mp_obj},  $c_{\textrm{smoothenss} }(\vect{\xi})$, is the penalty on the acceleration defined as $c_{\textrm{smoothenss} }(\vect{\xi}) = \sum^T_{t=1} \left\| \ddot{\vect{q}}_t \right\|^2.$
To make the computation efficient, the body of the robot manipulator and obstacles are approximated by a set of spheres. 
We used the shaping function in \eqref{eq:shape} as in the previous experiment.
The effect of the value of $\alpha$ was investigated in the following experiments.

\begin{figure}[tb]
	\centering
	\begin{subfigure}[t]{0.48\columnwidth}
		\centering
		\includegraphics[width=\textwidth]{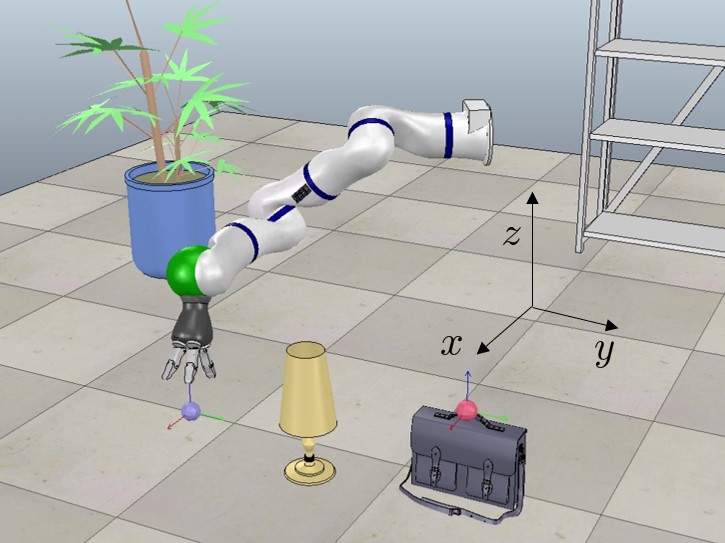}
		\caption{Start configuration for Task~1. }
	\end{subfigure}
	\begin{subfigure}[t]{0.48\columnwidth}
		\centering
		\includegraphics[width=\textwidth]{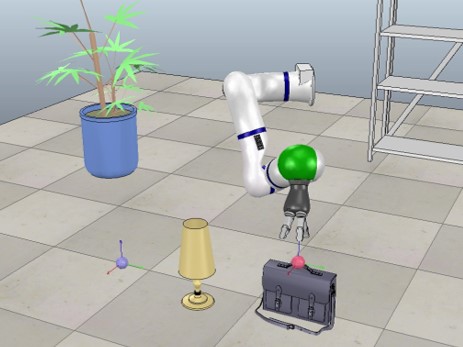}
		\caption{Goal configuration for Task~1. }
	\end{subfigure}
	\caption{Setting of Task~1.  }
	\label{fig:task1_setting}
\end{figure}

\begin{figure}[tb]
	\centering
	\begin{subfigure}[t]{0.48\columnwidth}
		\centering
		\includegraphics[width=\textwidth]{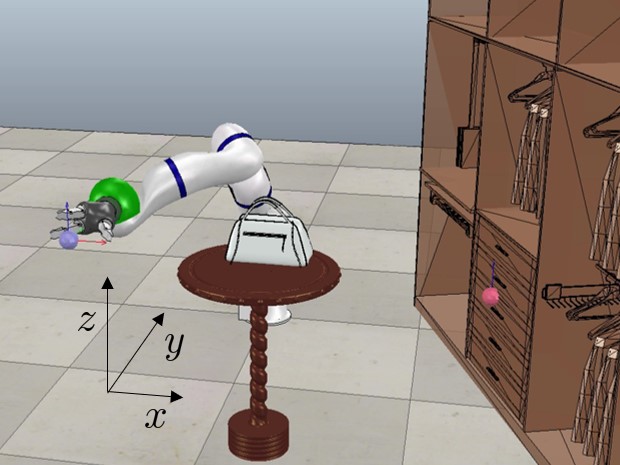}
		\caption{Start configuration for Task~2. }
	\end{subfigure}
	\begin{subfigure}[t]{0.48\columnwidth}
		\centering
		\includegraphics[width=\textwidth]{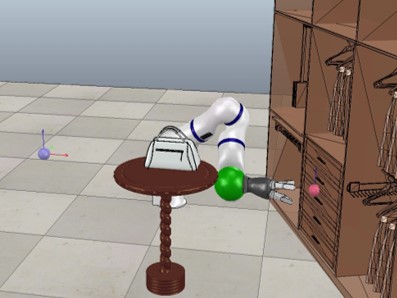}
		\caption{Goal configuration for Task~2. }
	\end{subfigure}
	\caption{Setting of Task~2.  }
	\label{fig:task2_setting}
\end{figure}

\begin{figure}[tb]
	\centering
	\begin{subfigure}[t]{0.48\columnwidth}
		\centering
		\includegraphics[width=\textwidth]{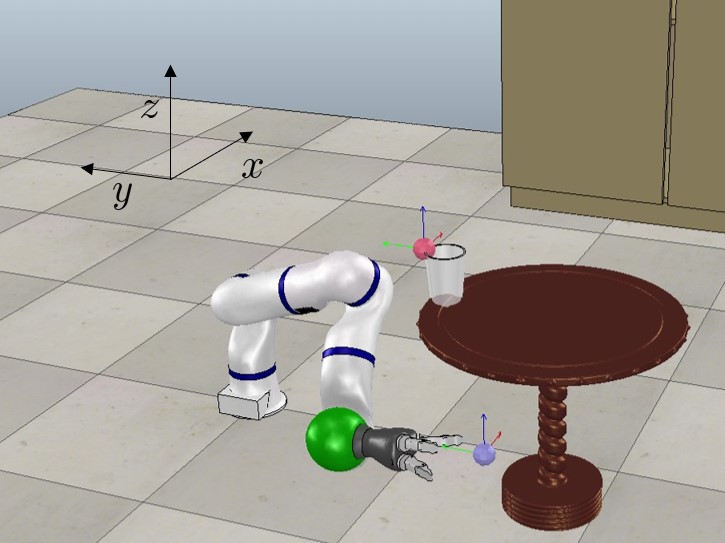}
		\caption{Start configuration for Task~1. }
	\end{subfigure}
	\begin{subfigure}[t]{0.48\columnwidth}
		\centering
		\includegraphics[width=\textwidth]{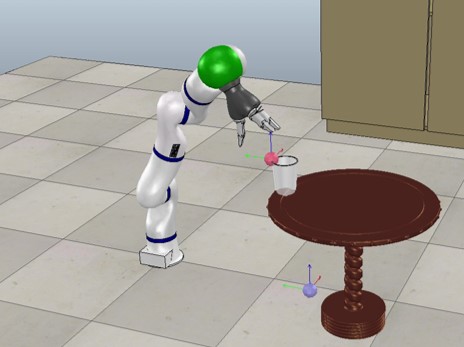}
		\caption{Goal configuration for Task~3. }
	\end{subfigure}
	\caption{Setting of Task~3.  }
	\label{fig:task3_setting}
\end{figure}

\begin{figure*}[t]
	\centering
	\begin{subfigure}[t]{2\columnwidth}
		\centering
		\includegraphics[width=\textwidth]{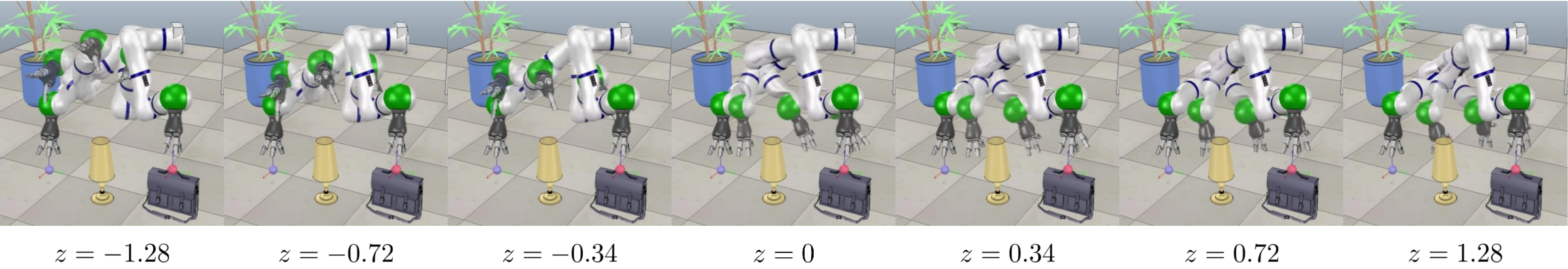}
		\caption{Result for Task~1. }
	\end{subfigure}
	\begin{subfigure}[t]{2\columnwidth}
		\centering
		\includegraphics[width=\textwidth]{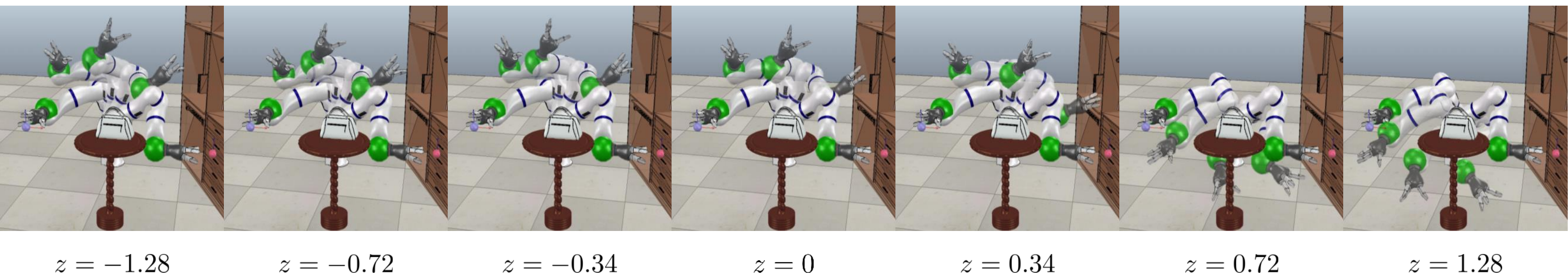}
		\caption{Result for Task~2. }
	\end{subfigure}
		\begin{subfigure}[t]{2\columnwidth}
		\centering
		\includegraphics[width=\textwidth]{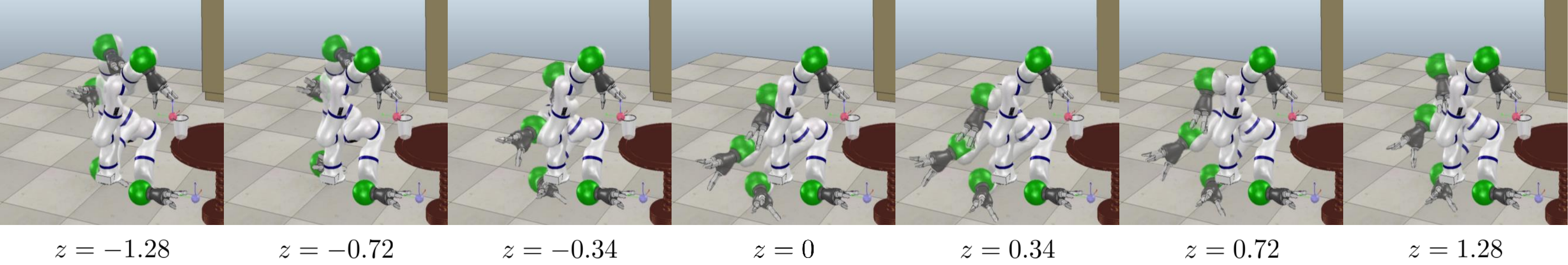}
		\caption{Result for Task~3. }
	\end{subfigure}
	\caption{Solutions generated from $p_{\vect{\theta}}(\vect{\xi}|\vect{z})$ with different values of latent variable $\vect{z}$. Result with one-dimensional latent variable.}  
	\label{fig:LSMO_sim_1d}
\end{figure*}

\subsection{Evaluation in Simulation }
We first evaluated the proposed method on three tasks in a simulation with a KUKA Light Weight Robot~(LWR) with 7 DoFs.
The task settings used are shown in Figures~\ref{fig:task1_setting}-\ref{fig:task3_setting}.
The conditions for training the neural network in MPSM is provided in the Appendix.

To analyze the effect of the scaling parameter $\alpha$ in \eqref{eq:shape}, we performed motion planning with $\alpha=$10, 20 and 50.
In addition, to see the effect of the parameterization based on RTP, we compared the results between the parametrization based on RTP and the waypoint parameterization.

As baseline methods, we also evaluated SMTO~\citep{osa20}, CHOMP~\citep{Zucker13}, and STOMP~\citep{Kalakrishnan11}.
SMTO was recently proposed by \cite{osa20}, and it finds multiple solutions for motion planning.
CHOMP and STOMP find a single solution for motion planning, and we choose them as baseline methods because our motion planning adapted CHOMP and STOMP in our framework. CHOMP is used for fine-tuning the trajectory, and the exploration strategy used in STOMP is adapted for a proposal distribution in MPSM. 
To evaluate the computation time of MPSM, we trained the model five times with different random seeds
and generated 30 samples by drawing the value of the latent variable from the uniform distribution $U(-1.28, 1.28)$.
The computation time for CHOMP and STOMP was evaluated by performing the motion planning tasks with different initializations 30 times. 
The initial trajectories for CHOMP and STOMP were drawn from the proposal distribution $\beta_{\textrm{traj}}(\vect{\xi})$ in \eqref{eq:exploration_traj}.


\begin{table}
	\caption{Number of solutions found by SMTO. }
	\centering
	\begin{tabular}{lccc}
		\toprule
		& Task~1     & Task~2   & Task~3 	 \\
		\midrule
		SMTO & $ 3.0 \pm 0.0$ &  $2.0\pm0.0$ & $3.0\pm0.0$ \\
		\bottomrule
	\end{tabular}
	\vspace{-0.3cm}
	\label{tbl:motion_num}
\end{table}

\begin{table*}[t]
	\caption{Effect of hyperparameters on computation time in motion planning tasks. }
	\centering
	\begin{tabular}{lcccccc}
		\toprule
		& \multicolumn{2}{c}{Task~1}     & \multicolumn{2}{c}{Task~2}   & \multicolumn{2}{c}{Task~3} 	 \\
		& \makecell{Training \\ $[ \textrm{min} ]$ }  & \makecell{Generation \\ $[ \textrm{s} ]$ }  & \makecell{Training \\ $[ \textrm{min} ]$ }   & \makecell{Generation \\ $[ \textrm{s} ]$ } & \makecell{Training \\ $[ \textrm{min} ]$ }   & \makecell{Generation \\ $[ \textrm{s} ]$ } \\
		\midrule
		MPSM w/ RTP, 1d, $\alpha=20$ & $\boldsymbol{5.18\pm0.06}$ & $\boldsymbol{0.21 \pm 0.31} $  & $5.26\pm 0.01$ &$ 0.23 \pm 0.31$ & $5.22\pm0.06$ &$0.11 \pm 0.17$ \\
		MPSM w/ RTP, 2d, $\alpha=20$ & $5.21 \pm0.01$ & $0.41 \pm 0.55 $  & $5.25\pm 0.04$ &$ 0.22 \pm 0.30$ & $5.22\pm0.03$ &$0.12 \pm 0.17$  \\
		MPSM w/o RTP, 1d, $\alpha=20$ & $24.4\pm0.16$ & $0.68 \pm 0.73 $  & $24.6\pm0.12$ &$ 0.24 \pm 0.04$ & $24.5\pm 0.03$ &$0.24 \pm 0.04$  \\
		MPSM w/ RTP, 1d, $\alpha=10$ & $5.24\pm0.07$ & $0.42\pm 0.43 $  & $5.21\pm0.04 $ & $\boldsymbol{0.19\pm0.26} $ & $5.10\pm0.03$ &$ 0.50\pm 0.31$ \\
		MPSM w/ RTP, 1d, $\alpha=50$ & $5.21\pm0.06 $ & $  0.33\pm0.53 $  & $\boldsymbol{5.17\pm0.07} $ &$  0.36\pm0.37 $ & $\boldsymbol{5.08\pm0.04}$ &$\boldsymbol{ 0.06\pm0.01 }$ \\
		\midrule
		SMTO    & - &$ 51.3\pm 0.69$  &- &$50.5\pm 1.0$&  - & $53.4\pm 1.7$ \\
		CHOMP    & -  &$1.8\pm 1.9$  & - &$ 0.52 \pm 0.22$&  - &$7.3 \pm 11.4$ \\
		STOMP    & -  &$7.91\pm 24.3$  & - &$ 1.24 \pm 2.23$&  - &$44.1 \pm 53.0$ \\
		\bottomrule
	\end{tabular}
	\vspace{-0.3cm}
	\label{tbl:motion_time}
\end{table*}

\begin{table*}[t]
	\caption{Effect of hyperparameters on scores of the trajectories before fine-tuning. Higher is better.}
	\centering
	\begin{tabular}{lccc}
		\toprule
		& Task~1     & Task~2   & Task~3 	 \\
		\midrule
		MPSM w/ RTP, 1d, $\alpha=20$ & $- 2.80 \pm 2.16$  & $- 2.61\pm2.78$   & $- 2.22\pm0.18$ \\
		MPSM w/ RTP, 2d, $\alpha=20$ & $- 3.83 \pm 3.23$ &  $- 2.60\pm2.83$ & $- 2.28\pm0.19$ \\
		MPSM w/o RTP, 1d, $\alpha=20$ & $\boldsymbol{-2.69 \pm 1.78}$ & $\boldsymbol{- 2.04\pm2.52}$ & $- 2.27\pm0.14$ \\
		MPSM w/ RTP, 1d, $\alpha=10$ & $- 3.74 \pm 2.53$  & $-2.40\pm2.52 $   & $-2.70 \pm 0.49$ \\
		MPSM w/ RTP, 1d, $\alpha=50$ & $- 3.04\pm2.94 $  & $- 3.86\pm3.37$   & $\boldsymbol{- 1.79\pm 0.16}$ \\
		\bottomrule
	\end{tabular}
	\vspace{-0.3cm}
	\label{tbl:motion_score_ablation}
\end{table*}

\subsubsection{Diversity of Solutions}
The results obtained using the model with the one-dimensional latent variable are presented in Figure~\ref{fig:LSMO_sim_1d}.
As shown, the model $p_{\vect{\theta}}(\vect{\xi}|\vect{z})$ trained with MPSM can generate various collision-free trajectories for the specified start and goal configurations.
SMTO found 2 to 3 solutions as summarized in Table~\ref{tbl:motion_num}, and it is evident that MPSM found more diverse solutions.
For example, the results of MPSM for Task~1 are shown in Figure~\ref{fig:LSMO_sim_1d}(a).
When the latent variable is one-dimensional on Task~1, the end-effector moves over the obstacle if $z=-1.28$, whereas the end-effector moves behind the obstacle if $z=1.28$.
As shown, the trajectory generated with the trained model continuously changes when the value of the latent variable $\vect{z}$ is changed continuously between $z=-1.28$ and $z=-0.34$, and between $z=0$ and $z=1.28$.
At the same time, Figure~\ref{fig:LSMO_sim_1d}(a) indicates a discontinuous change in the trajectory between $z=0$ and $z=-0.34$. This discontinuous change is resulted from fine-tuning with CHOMP.

\begin{figure}[tb]
	\centering
	\begin{subfigure}[t]{0.48\columnwidth}
		\centering
		\includegraphics[width=\textwidth]{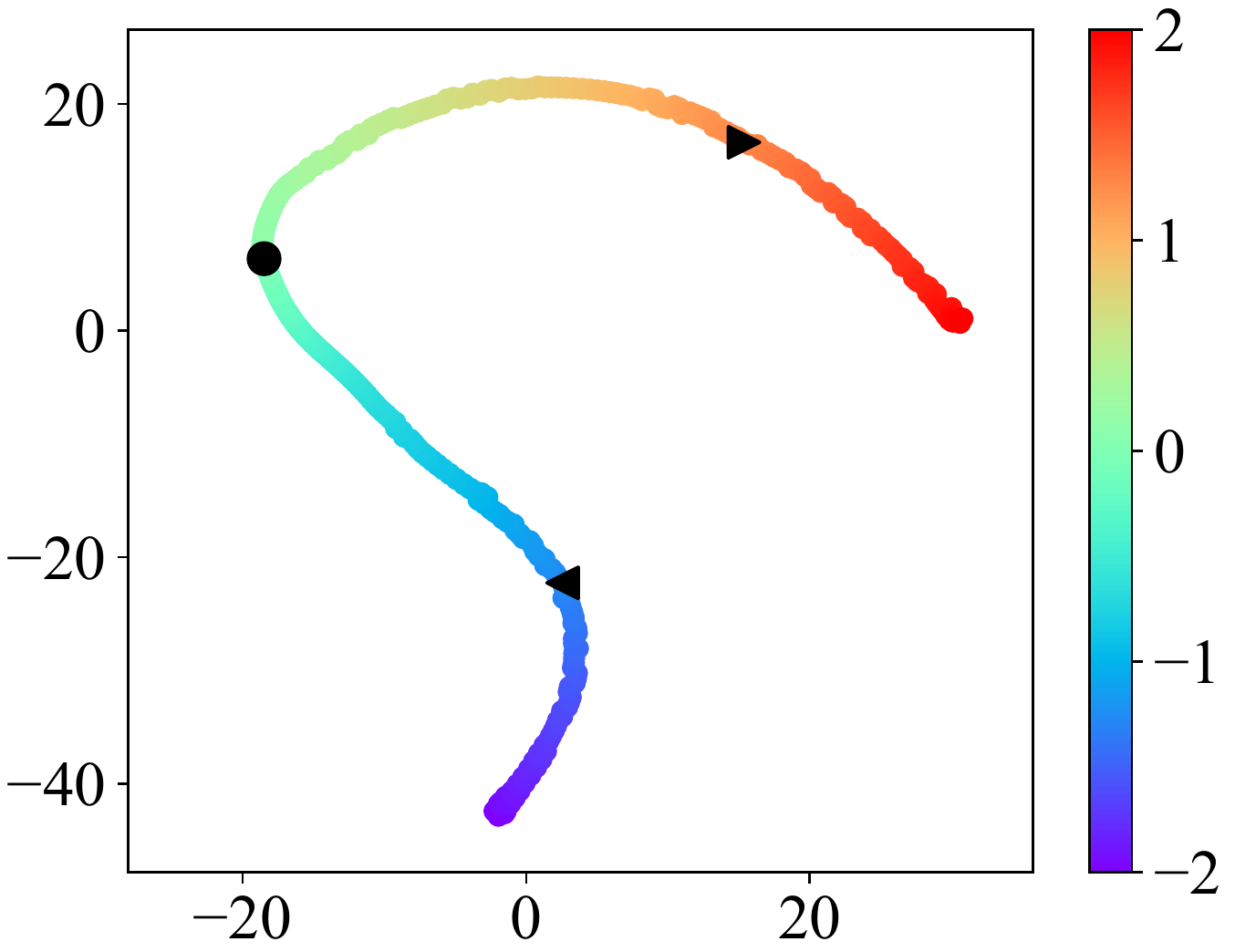}
		\caption{Distribution of solutions before fine-tuning. }
	\end{subfigure}
	\begin{subfigure}[t]{0.48\columnwidth}
		\centering
		\includegraphics[width=\textwidth]{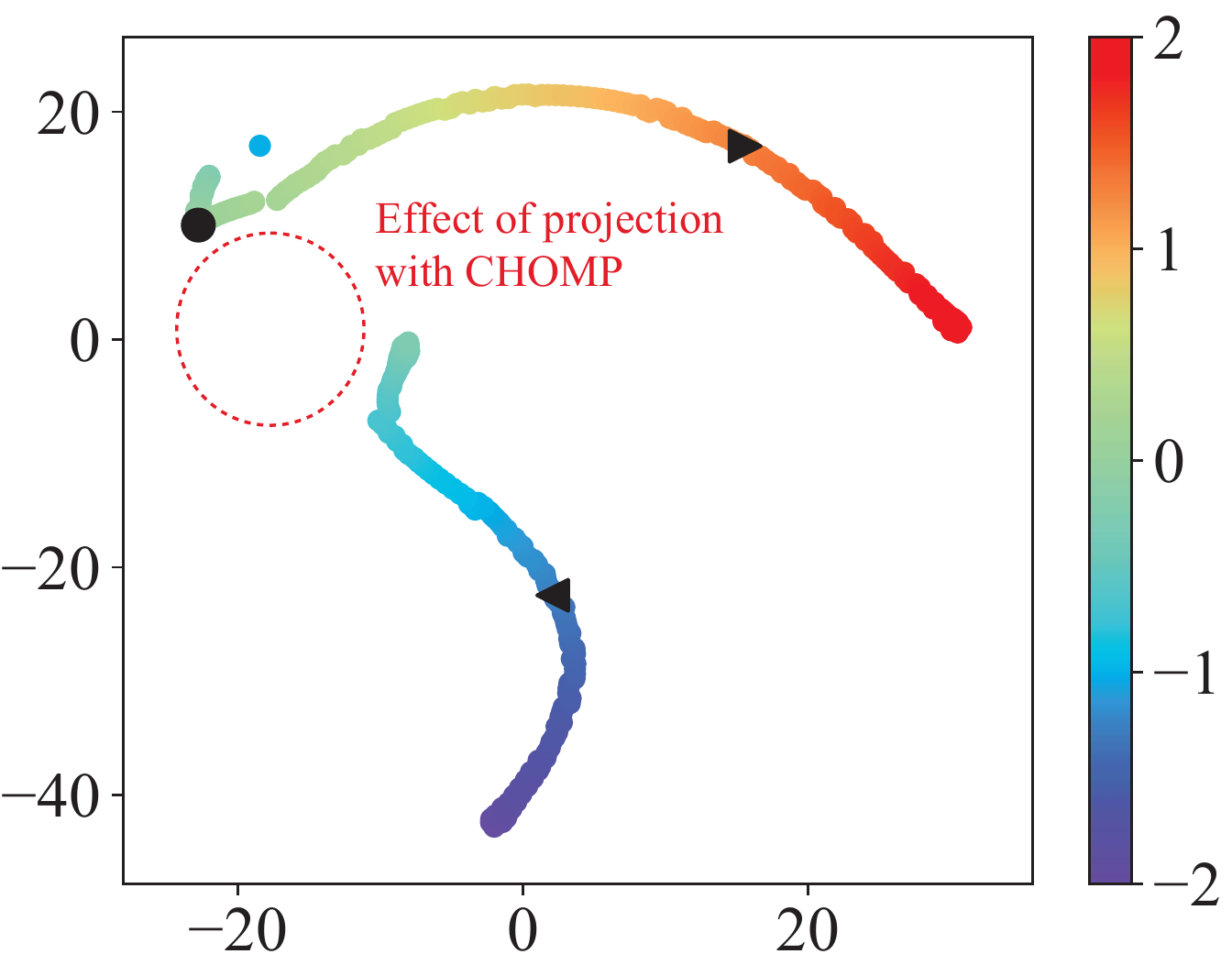}
		\caption{Distribution of solutions after fine-tuning. }
	\end{subfigure}
	\caption{Visualization of distribution of solutions on Task~1. Dimensionality is reduced using t-SNE. Color bar indicates value of $z$.  }
	\label{fig:task1_t-sne}
\end{figure}

The distributions of the outputs generated by the model for Task 1 is shown in Figure~\ref{fig:task1_t-sne} in which (a) and (b) show the distributions before and after fine-tuning, respectively.
For visualization, the latent variable is uniformly obtained from $[-2, 2]$, and the dimensionality of the trajectories is reduced using t-SNE~\citep{Maaten08}.
In Figure~\ref{fig:task1_t-sne}, the color bar indicates the value of $z$, and the black circles and triangles represent trajectories generated using $z=$-1.28, 0.0, and 1.28, respectively.
The distribution of outputs after fine-tuning with CHOMP is disconnected in some regions, as shown in Figure~\ref{fig:task1_t-sne}(b).
This result indicates that the model $p_{\vect{\theta}}(\vect{\xi}|\vect{z})$ includes solutions from different homotopy classes.

Prior studies show that a model with a continuous latent variable can represent samples from different classes/clusters.
For example, \cite{Kingma14} reported that a VAE with a continuous latent variable can model different hand-written digit images in the MNIST dataset, which contains multiple distinctive classes.
When samples from different classes are modeled using a VAE with a continuous latent variable, the datapoints are often interpolated, e.g., the model can generate an image between ``2'' and ``9'' when trained with the MNIST dataset. 
Likewise, the proposed method modeled trajectories from different homotopy classes, as shown in the experiments, and the trajectories were interpolated in the learned latent space.
Consequently, the distribution of the outputs is continuously connected, as shown in Figure~11(a).
However, in motion-planning problems, the interpolation of solutions from different homotopy classes can result in non-collision-free trajectories.
Therefore, we project the non-collision-free trajectories onto the collision-free solution space by performing fine-tuning using CHOMP. 
After fine-tuning, the distribution of the obtained solutions was disconnected, as shown in Figure~11(b).

\begin{figure}[tb]
	\centering
	\includegraphics[width=0.9\columnwidth]{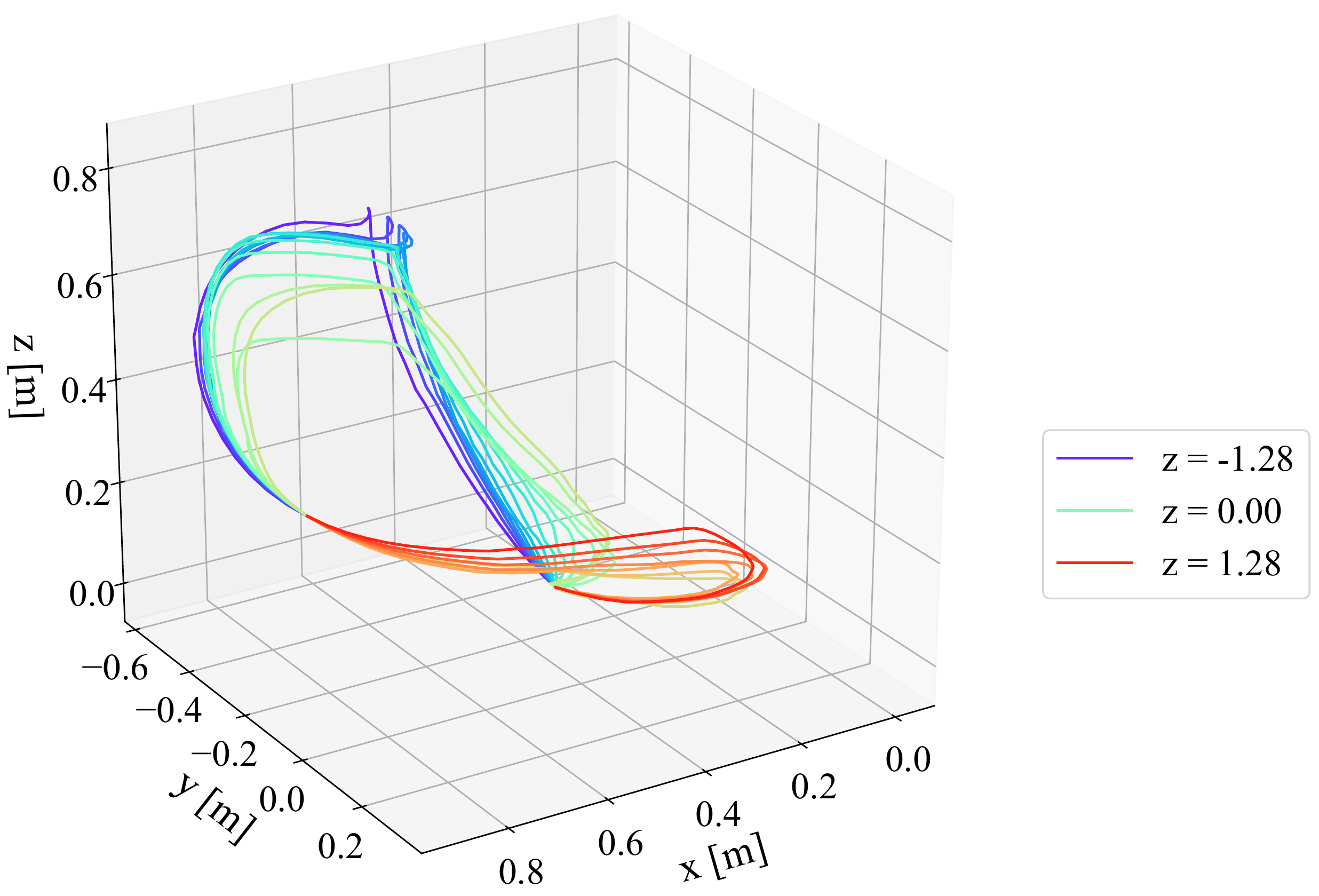}
	\caption{Trajectories in task space for Task~1. Result with one-dimensional latent variable. 20 trajectories are generated by linearly interpolating between $z = -1.28$ and $z = 1.28$. }
	\label{fig:task1_task}
\end{figure}

The task-space trajectories of the end-effector, which are shown in Figure~\ref{fig:task1_task}, imply that two clusters of solutions are available for Task~1.
These results indicate that the model trained with LSMO represents solutions from different homotopy classes.
Detailed results of Task~2 and 3 are provided in Appendix~\ref{sec:app_sim}; they support the discussion presented in this section.


\begin{figure*}[tb]
	\centering
	\begin{subfigure}[t]{0.32\columnwidth}
		\centering
		\includegraphics[width=\textwidth]{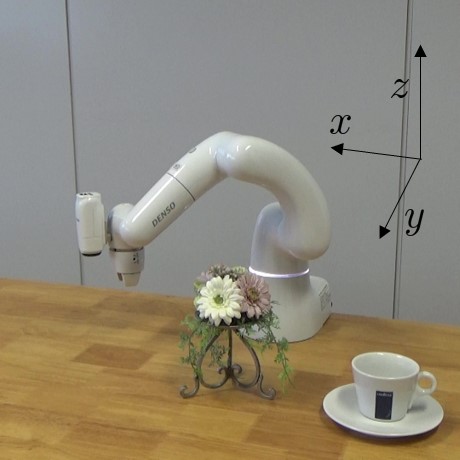}
		\caption{Start configuration for Task~4. }
	\end{subfigure}
	\begin{subfigure}[t]{0.32\columnwidth}
		\centering
		\includegraphics[width=\textwidth]{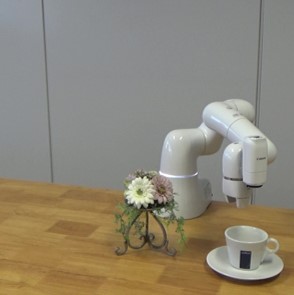}
		\caption{Goal configuration for Task~4. }
	\end{subfigure}
	\begin{subfigure}[t]{0.32\columnwidth}
		\centering
		\includegraphics[width=\textwidth]{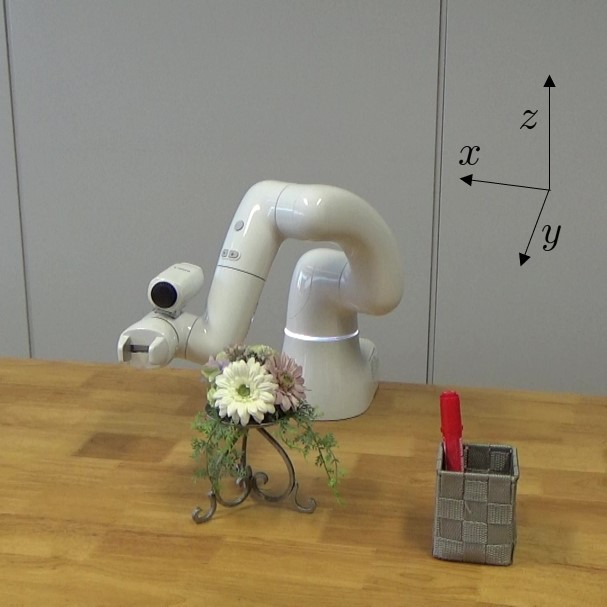}
		\caption{Start configuration for Task~5. }
	\end{subfigure}
	\begin{subfigure}[t]{0.32\columnwidth}
		\centering
		\includegraphics[width=\textwidth]{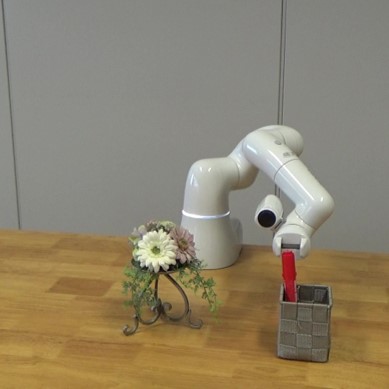}
		\caption{Goal configuration for Task~5. }
	\end{subfigure}
	\begin{subfigure}[t]{0.32\columnwidth}
		\centering
		\includegraphics[width=\textwidth]{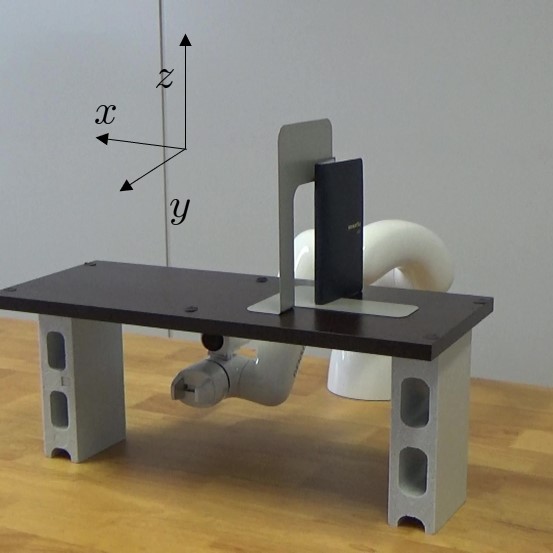}
		\caption{Start configuration for Task~6. }
	\end{subfigure}
	\begin{subfigure}[t]{0.32\columnwidth}
		\centering
		\includegraphics[width=\textwidth]{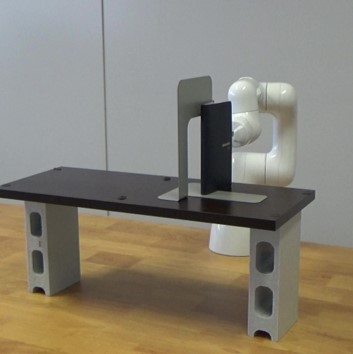}
		\caption{Goal configuration for Task~6. }
	\end{subfigure}
	\caption{Setting of Task~4, 5 and 6.  }
	\label{fig:cobotta_setting}
\end{figure*}

\begin{figure*}[tb]
	\centering
	\begin{subfigure}[t]{2\columnwidth}
		\centering
		\includegraphics[width=\textwidth]{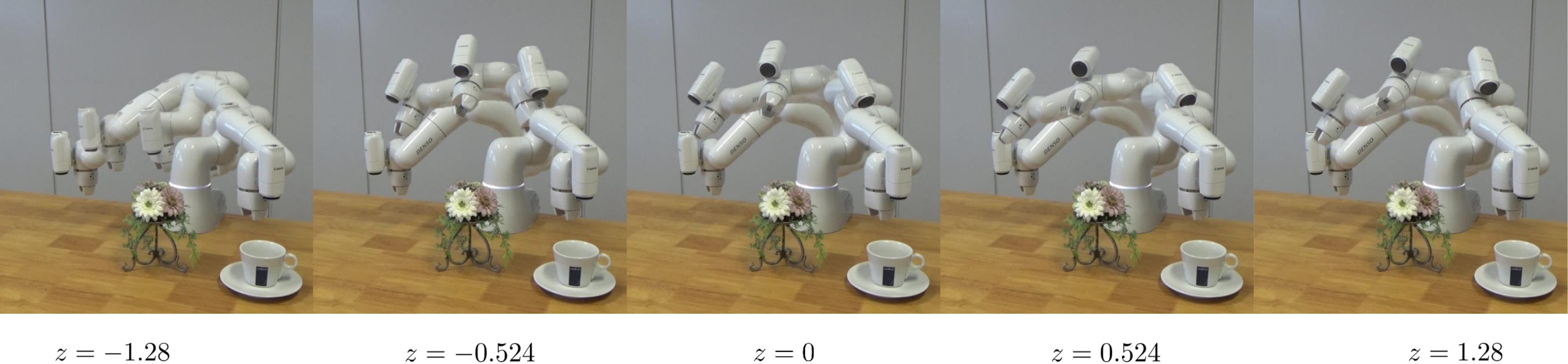}
		\caption{Solutions for Task~4. }
		\label{fig:cobotta2_solution}
	\end{subfigure}
	\begin{subfigure}[t]{2\columnwidth}
		\centering
		\includegraphics[width=\textwidth]{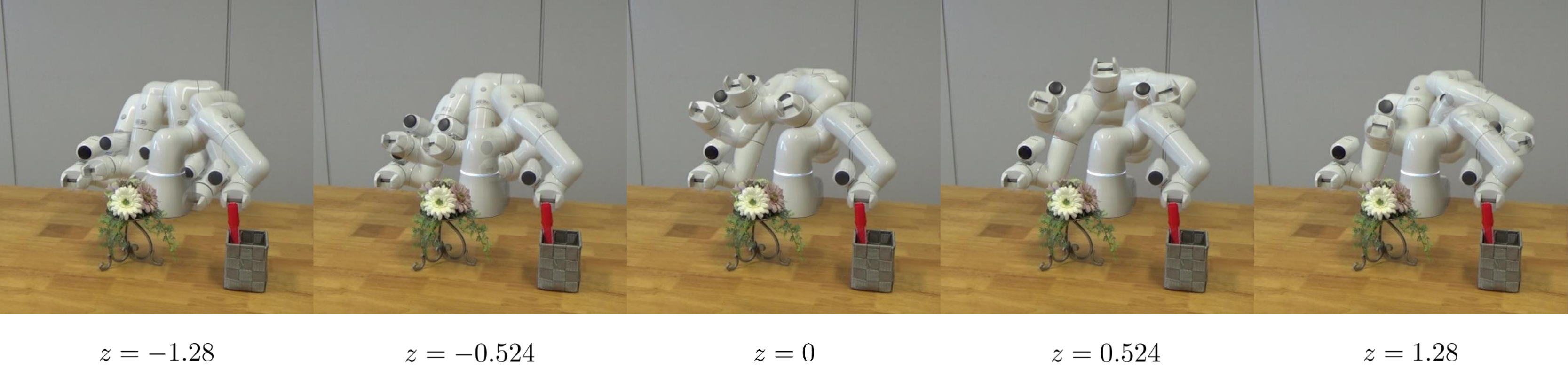}
		\caption{Solutions for Task~5. }
		\label{fig:cobotta3_solution}
	\end{subfigure}
	\begin{subfigure}[t]{2\columnwidth}
		\centering
		\includegraphics[width=\textwidth]{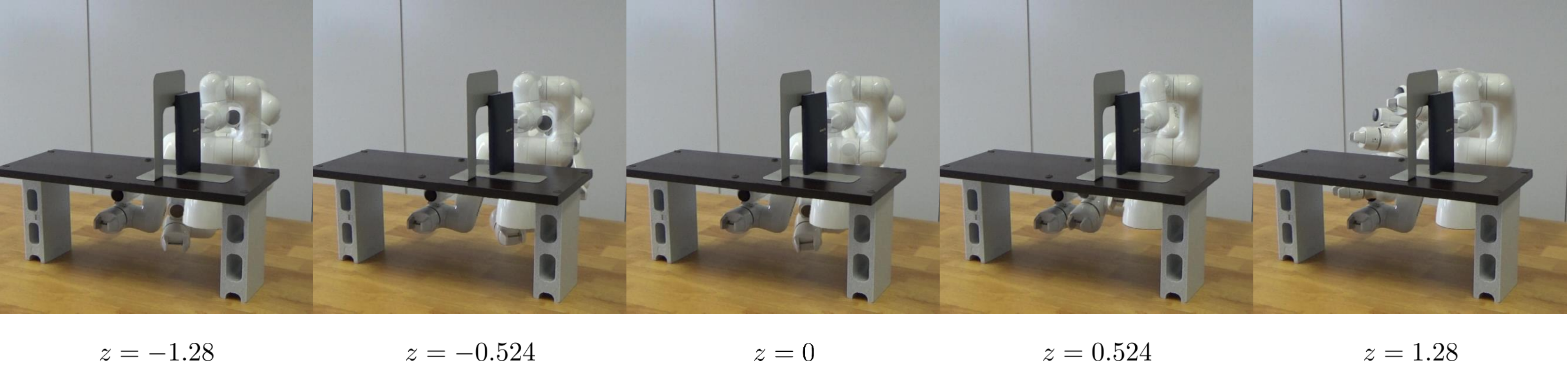}
		\caption{Solutions for Task~6. }
	\end{subfigure}
	\caption{Solutions for Task~4, 5 and 6.  Results with the one-dimensional latent variable. }
	\label{fig:cobotta_solutions}
\end{figure*}

\subsubsection{Computation Time and Effect of Hyperparameters}
The computation time for motion planning is summarized in Table~\ref{tbl:motion_time}.
When using the waypoint trajectory representation, the training time of MPSM takes about 25 min, 
whereas the training time of MPSM takes 5.5 min when using the parameterization based on RTPs.
The time required to generate a solution shown in Table~\ref{tbl:motion_time} indicates that these models exhibit comparable performance.
Therefore, the use of the parametrization based on RTPs can significantly reduce the computational cost.
When we use the way point parameterization, the value of each element may exhibit a large variance.
However, the variance can be reduced using RTPs because only the residual from the baseline trajectory is to be learned in RTPs. 

The results in Table~\ref{tbl:motion_time} also indicate that $\alpha=$20 outperforms $\alpha=$10 and 50 in the sense that the time required for fine-tuning was the minimum across the tasks.
The scores of the trajectories generated from the model shown in Table~\ref{tbl:motion_score_ablation} also indicate that $\alpha=$20 outperforms $\alpha=$10 and 50.
When the scaling factor $\alpha$ is larger, the relative importance of samples with higher scores becomes larger, which encourages the estimated density to be focused on the samples with high scores.
However, when the difference in the importance weight among samples is larger, the variance of estimating the loss function for training the neural network becomes larger, which may lead to unstable training.
Therefore, it is necessary to select an appropriate value for the scaling factor $\alpha$.

\begin{figure}[tb]
	\centering
		\includegraphics[width=\columnwidth]{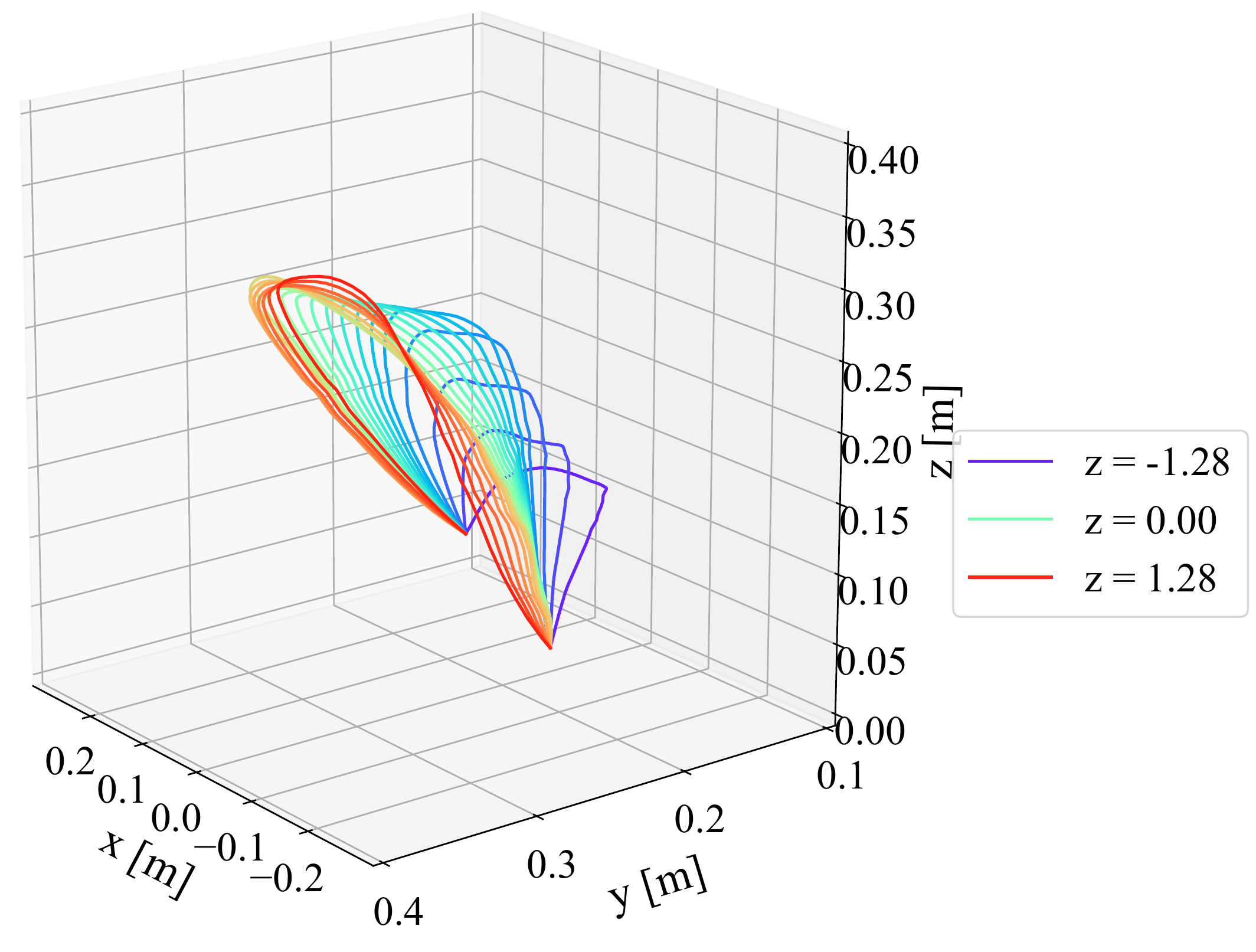}
		\caption{Solutions for Task~4 in task space. Results with the one-dimensional latent variable. }
	\label{fig:cobotta2_task}
\end{figure}

\begin{figure}[tb]
	\centering
	\includegraphics[width=\columnwidth]{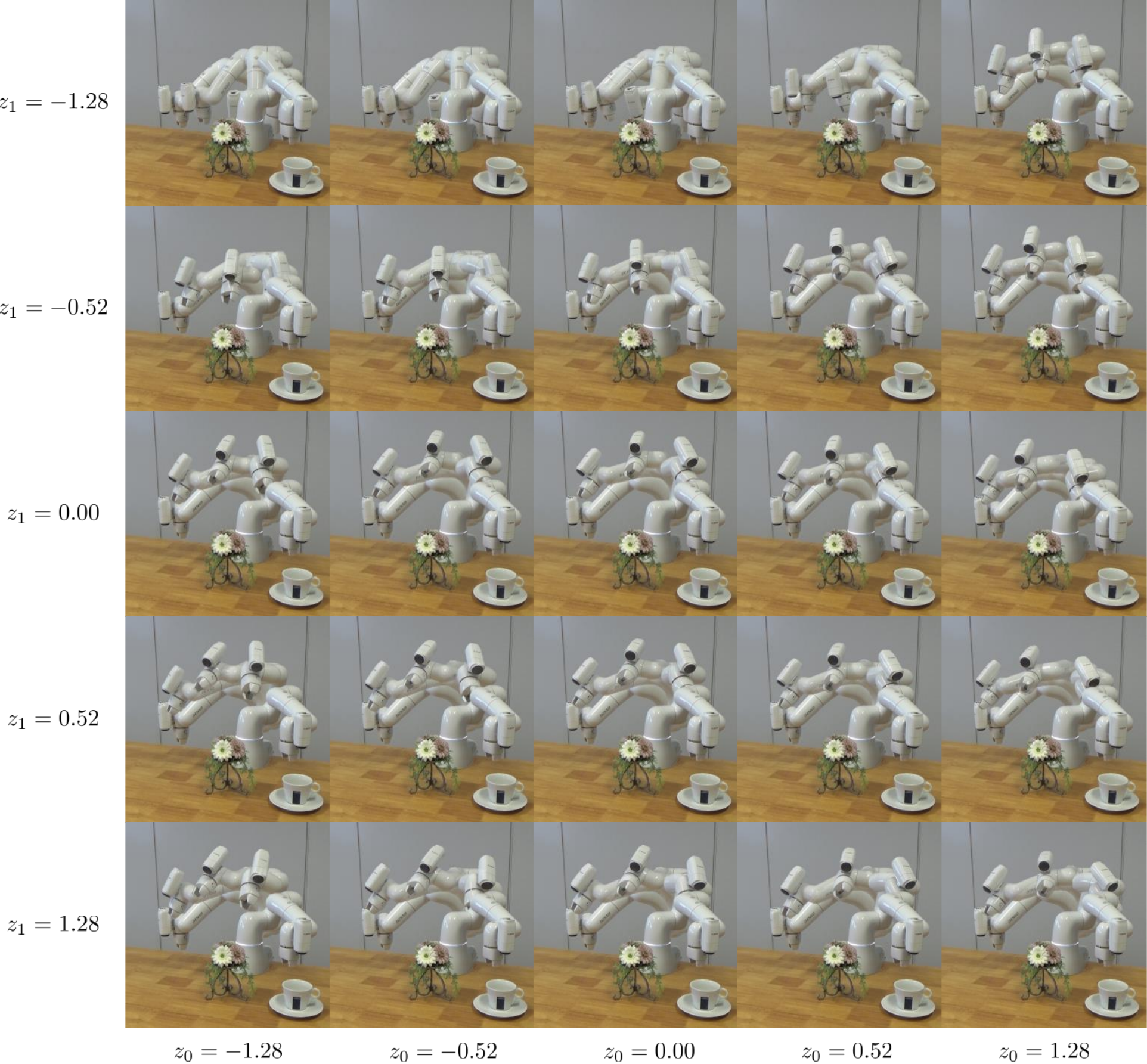}
	\caption{Result with two-dimensional latent variable. }
	\label{fig:cobotta2_2d}
\end{figure}

\begin{figure}[tb]
	\centering
	\begin{subfigure}[t]{0.48\columnwidth}
		\centering
		\includegraphics[width=\textwidth]{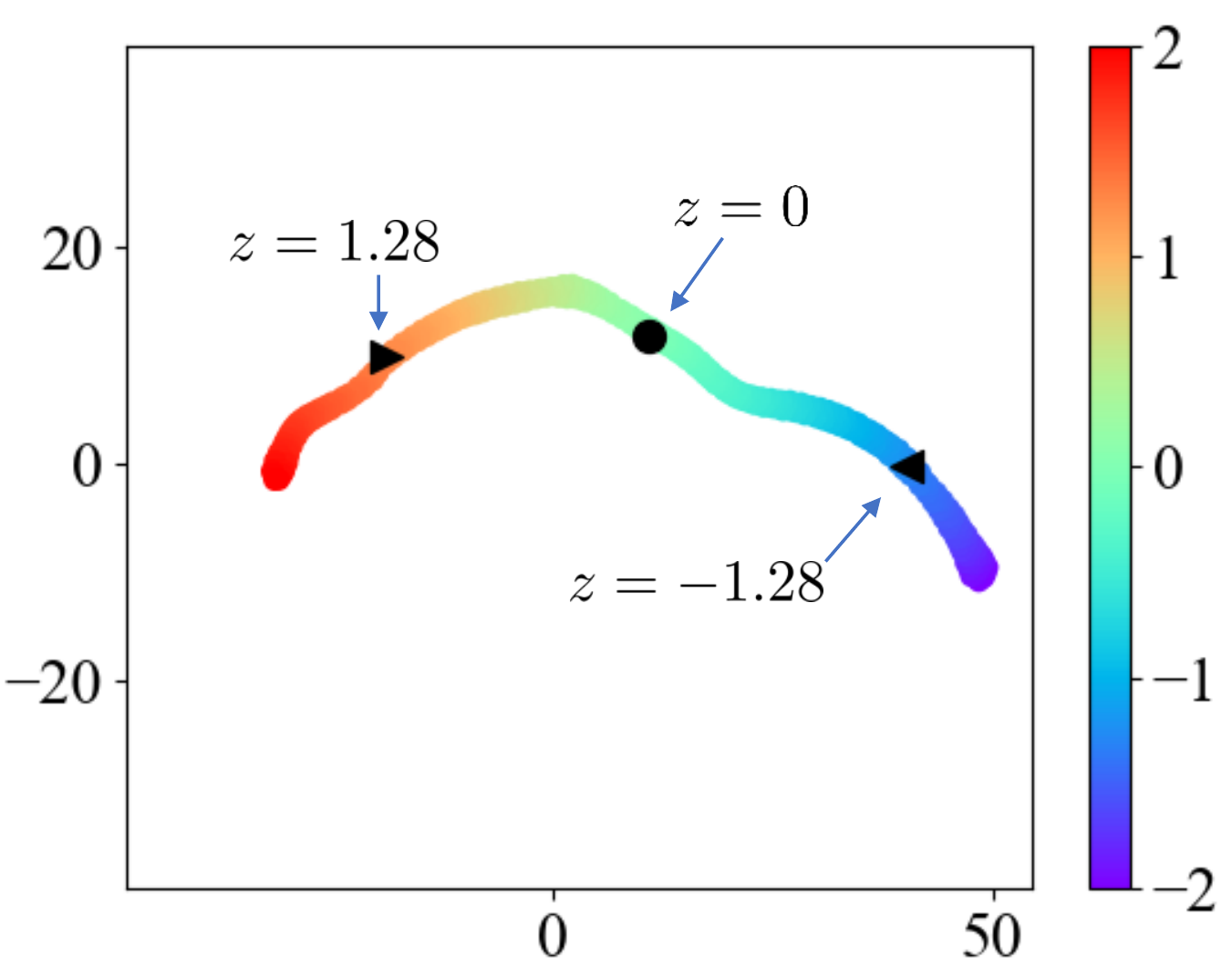}
		\caption{Result with one-dimensional latent variable. }
	\end{subfigure}
	\begin{subfigure}[t]{0.48\columnwidth}
		\centering
		\includegraphics[width=\textwidth]{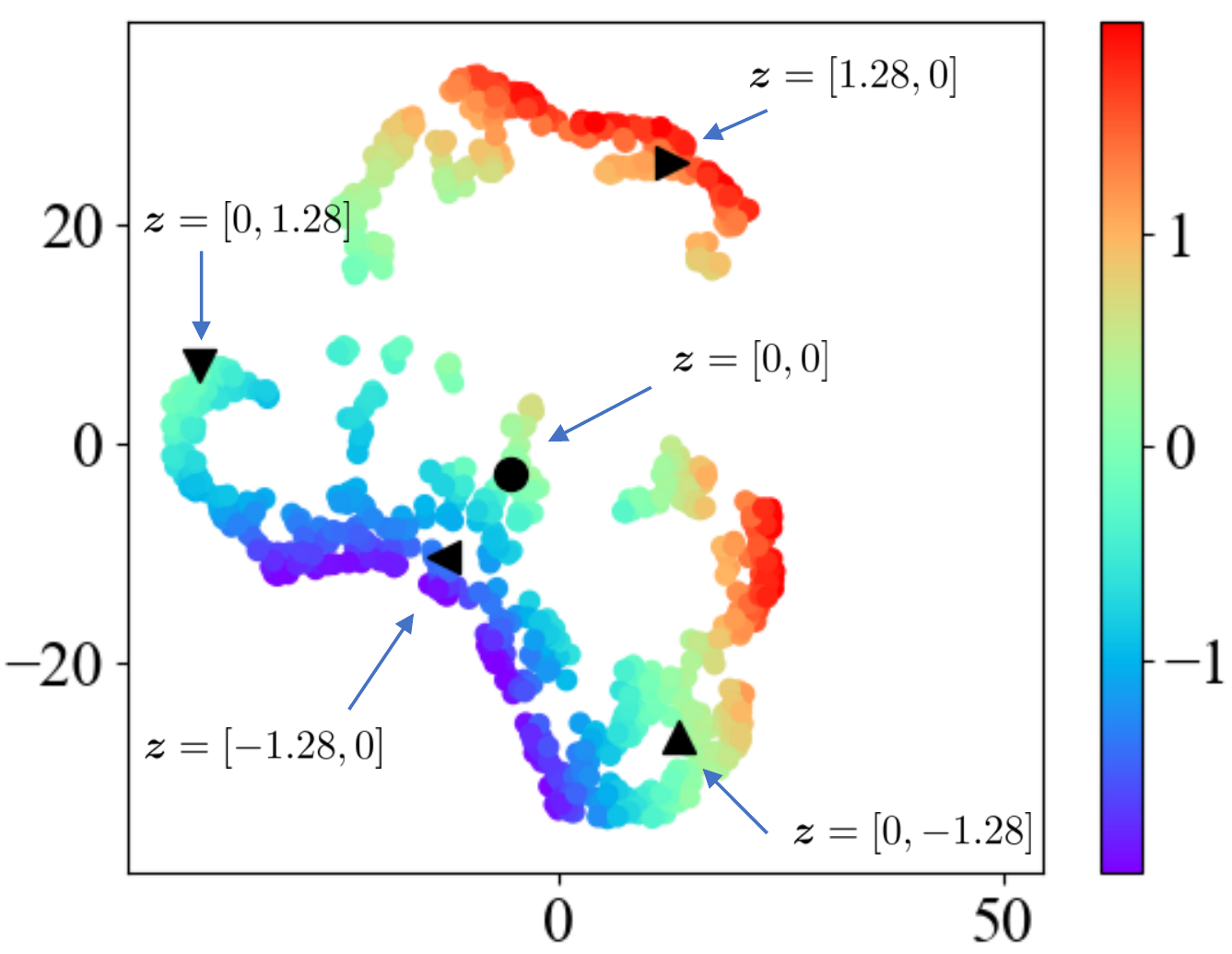}
		\caption{Result with two-dimensional latent variable.}
	\end{subfigure}
	\caption{Distribution of solutions on Task~4. Dimensionality is reduced using same transformation in (a) and (b).  The color bar indicates value of $z$ in (a) and $z_0$ in (b), respectively. }
	\label{fig:cobotta2_t-sne}
\end{figure}

\subsection{Experiments with Real Robot}

\subsubsection{Motion Planning for a Real Robot}
To verify that the proposed algorithm is applicable to real robots, we performed motion-planning experiments using a real robot.
We used Cobotta (Denso Wave Inc.), which has six DoFs, in the experiment.
The task settings are shown in Figure~\ref{fig:cobotta_setting}; in this study, we refer to the tasks as Tasks~4, 5, and 6, respectively.

The results with a one-dimensional latent variable are shown in Figure~\ref{fig:cobotta_solutions}.
For Task~4, the end-effector moves behind the obstacle when $z=-1.28$, whereas the end-effector moves over the obstacle when $z=1.28$, as shown in Figure~\ref{fig:cobotta_solutions}(a).
Figure~\ref{fig:cobotta2_task} shows the solutions obtained for Task~4 in the task space; as shown, the collision-free trajectories generated by the trained neural network changed continuously as the value of the latent variable was changed.
This result indicates that the neural network trained with MPSM captured a set of homotopic collision-free trajectories.
As shown in Figure~\ref{fig:cobotta_solutions}(b) and (c), MPSM also obtained diverse solutions for Tasks~5 and 6.
When we applied SMTO to these tasks, two or three solutions were obtained.
Therefore, it is evident that MPSM can obtain more diverse solutions than SMTO.
Please refer to Appendix~\ref{sec:app_real} for the solutions obtained by SMTO.

To demonstrate the effect of the dimensionality of the latent variable, we present the solutions obtained for Task~4 using the model with the two-dimensional latent variable in Figure~\ref{fig:cobotta2_2d}. 
The distributions of solutions obtained by MPSM for Task~4 were visualized using t-SNE, as shown in Figure~\ref{fig:cobotta2_t-sne}.
The results suggest that the model with the two-dimensional latent variable represents more diverse solutions than the model with the one-dimensional latent variable.
Regarding the fine-tuning of the output of the neural network, it was not necessary to fine-tune the outputs when the latent variable was one-dimensional for all tasks.
Meanwhile, when the latent variable was two-dimensional, fine-tuning was occasionally necessary.
This result indicates that there is a trade-off between the diversity and accuracy of solutions when selecting the dimensionality of the latent variable.
It can also be seen that the difference in the information encoded in the two channels are not clear when the latent variable is two-dimensional. 
For example, Figure~\ref{fig:cobotta2_2d} implies that both $z_0$ and $z_1$ encode the variation in the height of the end-effector.
Therefore, the results imply that increasing the dimensionality of the latent variable may complicate the process for the user to examine solutions captured by MPSM, although this may lead to a greater diversity of solutions.
The results of Tasks~5 and 6 are provided in Appendix~\ref{sec:app_real}; they also support the discussions presented in this section.

\subsubsection{Adaptation to Scene Change}
\begin{figure}[tb]
	\centering
	\includegraphics[width=\columnwidth]{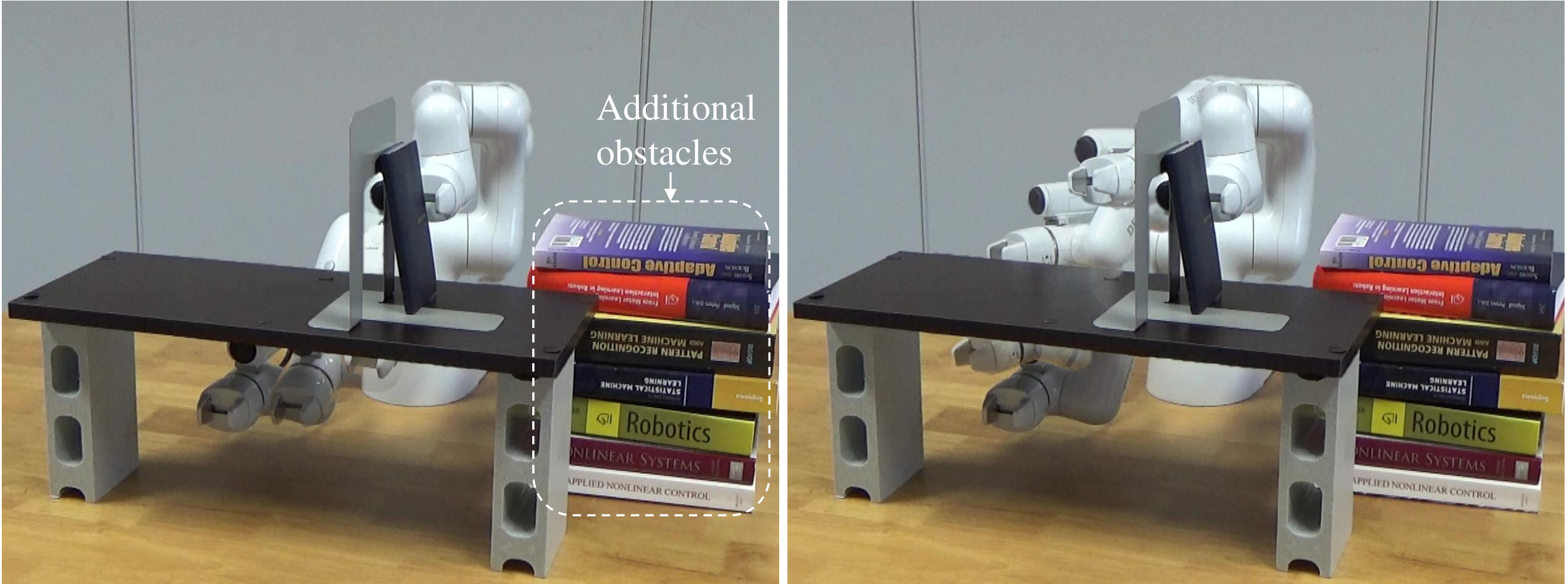}
	\caption{Solutions obtained for scenario changed from Task~6.} 
	\label{fig:cobotta4_adaptation}
\end{figure}

As discussed by \cite{Kumar20} in the context of reinforcement learning, obtaining diverse solutions can lead to robustness against scene changes. 
In our framework, diverse solutions can be obtained for a motion planning problem. 
When a few obstacles are added to the original scene,  
feasible solutions can be obtained in a set of the solutions for the original scenario, although a subset of solutions are disabled.
In the scenario shown in Figure~\ref{fig:cobotta4_adaptation}, obstacles were added in the scenario of Task~6, which is shown in Figure~\ref{fig:cobotta_setting}(e) and (f).
Our system obtained solutions shown in Figure~\ref{fig:cobotta4_adaptation} using the model trained for Task~6.
In our implementation, we can verify the collision of a trajectory in approximately 36~ms and quickly identify usable trajectories from those obtained for the original scenario.
Even if the trajectory identified for the original scenario is not directly usable for a new scenario, it can be serve as a descent initial trajectory for trajectory optimization when the change in the scenario is insignificant.
Therefore, the model need  not to be re-trained from scratch if the change in the scenario is insignificant.

\section{Discussion}
The motivation of this work is to allow the user to select the preferable solutions because the objective function used in motion planning may not properly reflect the user preference.
If the end-effector moves over the obstacle as in the left-most frame of Figure~\ref{fig:LSMO_sim_1d}(a), the user may find it scary.
However, programming such a preference is not trivial, and dealing with all such preferences is actually challenging.
Therefore, we provide diverse solutions for the user and let the user select one, rather than programming the user preferences.
We believe that our framework enhances the usability of a motion planner and helps make robots usable in daily life.
Although we focused on motion planning problems in this study, the concept of learning diverse solutions by learning latent representations is applicable to other domains.
For example, in our recent study, we developed a reinforcement learning method that obtains diverse solutions by learning latent representations; subsequently we applied it to continuous control tasks~\citep{Osa21}.

The tasks employed in this study involve large free space because our framework is beneficial in such circumstances.
Under the existence of a large free space, there are diverse ways to avoid obstacles and reach the desired position.
In such cases, the user will need to examine different solutions to perform the motion and select the type of motion.
In our framework, the variation of the trajectories is encoded in a low-dimensional latent variable, and this allows the user to intuitively examine different solutions.
If the free space is limited, then a possible variation of the solutions are limited.
In such cases, the user does not need to examine diverse solutions, and she/he is encouraged to employ sophisticated existing motion planning frameworks to find a single solution efficiently and robustly, such as TrajOpt~\citep{Schulman14}, GPMP~\citep{Mukadam18} and BIT*~\citep{Gammell20}.

Our method trains a generative model for solutions in motion planning.
Various methods for training deep generative models have been developed in recent studies~\citep{Goodfellow14,Chen17,Arjovsky17,Dupont18}; 
prior work such as \citep{Bengio13,Berthelot18,Verma19} investigated how to obtain meaningful latent representations in the context of unsupervised learning.
Although we investigated the manner in which techniques for deep generative models can be leveraged for motion planning, methods to obtain meaningful latent representations for motion-planning problems must be further investigated.
A limitation of our framework is that it takes approximately 5 min to train a neural network, which is not required for methods that do not use a neural network.
As obtaining an infinite set of diverse solutions is challenging, it is natural that there is a trade-off between the diversity of solutions and the computation time.
When using existing motion planning methods, it is necessary to manually tune the objective function or explore different random seeds in order to obtain different types of solutions.
Compared with such efforts, we think that the time required for training a neural network is negligible in practice.

A possible extension of this work is to learn both continuous and discrete latent variables to explicitly learn multiple separate sets of solutions.
This extension should be possible with the Gumbel-Softmax trick~\citep{Jang17,Maddison17}.
Regarding the objective function, we did not explicitly incorporate topology-based representations in the objective function, although the trained model captured diverse solutions from homotopic classes.
\cite{Ivan13} investigated the topology-based representations to describe the geometric relations between robots and objects.
Incorporating such representations with LSMO will be an interesting research topic.
Another possible extension is to make the neural network conditioned on scene features in such a way that the neural network can generate a trajectory for unseen situations. 
This is an important research direction to remove the necessity of training a neural work for each situation.
However, training a neural network to generate a collision-free trajectory for unseen situations is challenging~\citep{Srinivas18}, and training a neural network for generating diverse collision-free trajectories for unseen situations is even more challenging.
We will investigate this extension in future work.

\section{Conclusion}
In this study, we presented LSMO, which is an algorithm for learning an infinite set of solutions in optimization.
In our framework, diverse solutions are captured by learning latent representations of solutions.
We derived the proposed algorithm by considering the variational lower bound of the expected score of solutions.
We then adapted LSMO to motion planning problems and developed a novel motion planning algorithm, which we referred to as MPSM.
Our approach can be interpreted as training a deep generative model of collision-free trajectories for motion planning.
In our experiments, we show that a set of homotopic solutions can be obtained with MPSM on motion planning tasks, which involve hundreds of parameters.
Our framework for learning the solution manifold gives users an intuitive way to go through various solutions by changing the values of the latent variables.
We believe that the approach of learning the solution manifold in optimization can enhance the usability of motion planners in robotics by providing diverse solutions for the user.
In future work, we will investigate how to incorporate other deep generative models such as GANs in our framework.

\begin{funding}
	This work was partially supported by JSPS KAKENHI Grant Number 19K20370.
\end{funding}


\begin{thebibliography}{99}
	\bibitem[{{Agrawal} et~al.(2014){Agrawal}, {Shen} and v.~d.
		{Panne}}]{Agrawal14}
	{Agrawal} S, {Shen} S and v~d {Panne} M (2014) Diverse motions and character
	shapes for simulated skills.
	\newblock \emph{IEEE Transactions on Visualization and Computer Graphics}
	20(10): 1345--1355.	
	\bibitem[{Amari} (2016)]{Amari16}
	{Amari} S (2016) \emph{Information geometry and its applications}.
	\newblock Springer.	
%
	\bibitem[{Argall et~al.(2009)Argall, Chernova, Veloso and Browning}]{Argall09}
	Argall BD, Chernova S, Veloso M and Browning B (2009) A survey of robot
	learning from demonstration.
	\newblock \emph{Robotics and Autonomous Systems} 57(5): 469--483.
%
	\bibitem[{Arjovsky et~al.(2017)Arjovsky, Chintala and Bottou}]{Arjovsky17}
	Arjovsky M, Chintala S and Bottou L (2017) Wasserstein generative adversarial
	networks.
	\newblock In: \emph{Proceedings of the International Conference on Machine
		Learning (ICML)}.
%
	\bibitem[{Bacon et~al.(2017)Bacon, Harb and Precup}]{Bacon17}
	Bacon PL, Harb J and Precup D (2017) The option-critic architecture.
	\newblock In: \emph{Proceedings of the AAAI Conference on Artificial
		Intelligence (AAAI)}.
%
	\bibitem[{Bengio et~al.(2013)}]{Bengio13}
	Bengio Y, Mesnil G, Dauphin Y and Rifai S (2013) Better Mixing via Deep Representations.
	\newblock In: \emph{Proceedings of the International Conference on Machine
		Learning (ICML)}.
%
	\bibitem[{Berthelot et~al.(2018)}]{Berthelot18}
	Berthelot D, Raffel C, Roy A and Goodfellow I (2018) Understanding and Improving Interpolation in Autoencoders via an Adversarial Regularizer.
	\newblock In: \emph{Proceedings of the International Conference on Learning Representations (ICLR)}.
%
	\bibitem[{Boyd and Vandenberghe(2004)}]{Boyd04}
	Boyd S and Vandenberghe L (2004) \emph{Convex Optimization}.
	\newblock Cambridge University Press.
%
	\bibitem[{Chen et~al.(2020)Chen, Dai, Lin, Ye, Liu and Song}]{Chen20}
	Chen B, Dai B, Lin Q, Ye G, Liu H and Song L (2020) Learning to plan in high
	dimensions via neural exploration-exploitation trees.
	\newblock In: \emph{Proceedings of the International Conference on Learning
		Representations (ICLR)}.
	\bibitem[{Chen et~al.(2016)Chen, Duan, Huthooft, Schulman, Sutskever and
		Abbeel}]{Chen17}
	Chen X, Duan Y, Huthooft R, Schulman J, Sutskever I and Abbeel P (2016)
	Infogan: Interpretable representation learning by information maximizing
	generative adversarial nets.
	\newblock In: \emph{Advances in Neural Information Processing Systems (NIPS)}.
%
	\bibitem[{Cremer et~al.(2018)Cremer, Li and Duvenaud}]{Cremer18}
	Cremer C, Li X and Duvenaud D (2018) Inference suboptimality in variational
	autoencoders.
	\newblock In: \emph{Proceedings of the International Conference on Machine
		Learning (ICML)}.
%
	\bibitem[{Dayan and Hinton(1997)}]{Dayan97}
	Dayan P and Hinton G (1997) Using expectation-maximization for reinforcement
	learning.
	\newblock \emph{Neural Computation} 9: 271--278.
%
	\bibitem[{de~Boer et~al.(2005)de~Boer, Kroese, Mannor and Rubinstein}]{Boer05}
	de~Boer PT, Kroese DP, Mannor S and Rubinstein RY (2005) A tutorial on the
	cross-entropy method.
	\newblock \emph{Annals of Operations Research} 134: 19--67.
%
	\bibitem[{Deb and Saha(2010)}]{Deb10}
	Deb K and Saha A (2010) Finding multiple solutions for multimodal optimization
	problems using a multi-objective evolutionary approach.
	\newblock In: \emph{Proceedings of the 12th annual conference on Genetic and
		evolutionary computation}.
%
	\bibitem[{Dupont(2018)}]{Dupont18}
	Dupont E (2018) Learning disentangled joint continuous and discrete
	representations.
	\newblock In: \emph{Advances in Neural Information Processing Systems 31 (NIPS
		2018))}.
%
	\bibitem[{Eysenbach et~al.(2019)Eysenbach, Gupta, Ibarz and
		Levine}]{Eysenbach19}
	Eysenbach B, Gupta A, Ibarz J and Levine S (2019) Diversity is all you need:
	Learning skills without a reward function.
	\newblock In: \emph{Proceedings of the International Conference on Learning
		Representations (ICLR)}.
%
	\bibitem[{Florensa et~al.(2017)Florensa, Duan and Abbeel}]{Florensa17}
	Florensa C, Duan Y and Abbeel P (2017) Stochastic neural networks for
	hierarchical reinforcement learning.
	\newblock In: \emph{Proceedings of the International Conference on Learning
		Representations (ICLR)}.
%
	\bibitem[{Gammell et~al.(2020)Gammell, Barfoot, D. and Srinivasa}]{Gammell20}
	Gammell JD, Barfoot, D T and Srinivasa SS (2020) Batch informed trees (bit*):
	Informed asymptotically optimal anytime search.
	\newblock \emph{The International Journal of Robotics Research} 39(5):
	543--567.
%
	\bibitem[{Goldberg and Richardson(1987)}]{Goldberg87}
	Goldberg DE and Richardson J (1987) Genetic algorithms with sharing for
	multimodal function optimization.
	\newblock In: \emph{Proceedings of the Second International Conference on
		Genetic Algorithms}.
%
	\bibitem[{Goodfellow et~al.(2014)Goodfellow, Pouget-Abadie, Mirza, Xu,
		Warde-Farley, Ozair, Courville and Bengio}]{Goodfellow14}
	Goodfellow IJ, Pouget-Abadie J, Mirza M, Xu B, Warde-Farley D, Ozair S,
	Courville A and Bengio Y (2014) Generative adversarial nets.
	\newblock In: \emph{Advances in Neural Information Processing Systems (NIPS)}.
%
	\bibitem[{Hansen and Ostermeier(1996)}]{Hansen96}
	Hansen N and Ostermeier A (1996) Adapting arbitrary normal mutation
	distributions in evolution strategies: The covariance matrix adaptation.
	\newblock In: \emph{Proceedings of the IEEE International Conference on
		Evolutionary Computation}.
%
	\bibitem[{Hatcher(2002)}]{Hatcher02}
	Hatcher A (2002) \emph{Algebraic Topology}.
	\newblock Cambridge University Press.
	
	\bibitem[{Ivan(2013)}]{Ivan13}
	Ivan V, Zarubin D, Toussaint M and Vijayakumar S (2013) 
	Topology-based Representations for Motion Planning and Generalisation in Dynamic Environments with Interactions
	\newblock \emph{The International Journal of Robotics Research}, 32(9--10):1151--1163.
%
	\bibitem[{Jaillet and Simeon(2008)}]{Jaillet08}
	Jaillet L and Simeon T (2008) Path deformation roadmaps: Compact graphs with
	useful cycles for motion planning.
	\newblock \emph{The International Journal of Robotics Research} 27(11--12):
	1175--1188.
%
	\bibitem[{Jang et~al.(2017)Jang, Gu and Poole}]{Jang17}
	Jang E, Gu S and Poole B (2017) Categorical reparameterization with
	gumbel-softmax.
	\newblock In: \emph{Proceedings of the International Conference on Learning
		Representations (ICLR)}.
%
	\bibitem[{Jurgenson and Tamar(2019)}]{Jurgenson19}
	Jurgenson T and Tamar A (2019) Harnessing reinforcement learning for neural
	motion planning.
	\newblock In: \emph{Proceedings of Robotics: Science and Systems (R:SS)}.
%
	\bibitem[{Kalakrishnan et~al.(2011)Kalakrishnan, Chitta, Theodorou, Pastor and
		Schaal}]{Kalakrishnan11}
	Kalakrishnan M, Chitta S, Theodorou E, Pastor P and Schaal S (2011) Stomp:
	Stochastic trajectory optimization for motion planning.
	\newblock In: \emph{Proceedings of the IEEE International Conference on
		Robotics and Automation (ICRA)}. pp. 4569--4574.
%
	\bibitem[{{Kalakrishnan} et~al.(2013){Kalakrishnan}, {Pastor}, {Righetti} and
		{Schaal}}]{Kalakrishnan13}
	{Kalakrishnan} M, {Pastor} P, {Righetti} L and {Schaal} S (2013) Learning
	objective functions for manipulation.
	\newblock In: \emph{The IEEE International Conference on Robotics and
		Automation (ICRA)}. pp. 1331--1336.
%
	\bibitem[{Karaman and Frazzoli(2011)}]{Karaman11}
	Karaman S and Frazzoli E (2011) Sampling-based algorithms for optimal motion
	planning.
	\newblock \emph{The International Journal of Robotics Research} 30(7): 846--894.
%
	\bibitem[{Karasawa et~al.(2020)Karasawa, Kanemaki, Oomae, Fukui, Nakao and
		Osa}]{Karasawa20}
	Karasawa H, Kanemaki T, Oomae K, Fukui R, Nakao M and Osa T (2020) Hierarchical
	stochastic optimization with application to parameter tuning for
	electronically controlled transmissions.
	\newblock \emph{IEEE Robotics and Automation Letters} 5(2): 628--635.
%
	\bibitem[{Kavraki et~al.(1998)Kavraki, Kolountzakis and Latombe}]{Kavraki98}
	Kavraki LE, Kolountzakis MN and Latombe JC (1998) Analysis of probabilistic
	roadmaps for path planning.
	\newblock \emph{IEEE Transactions on Roborics and Automation} 14(1): 166--171.
%
	\bibitem[{Kavraki et~al.(1996)Kavraki, Svestka, Latombe and
		Overmars.}]{Kavraki96}
	Kavraki LE, Svestka P, Latombe JC and Overmars MH (1996) Probabilistic roadmaps 
	for path planning in high-dimensional configuration spaces.
	\newblock \emph{IEEE Transactions on Robotics and Automation} 12(4): 566--580.
%
	\bibitem[{Khatib(1986)}]{Khatib86}
	Khatib O (1986) Real-time obstacle avoidance for manipulators and mobile
	robots.
	\newblock \emph{The International Journal of Robotics Research} 5(1): 90--98.
%
	\bibitem[{Kim et~al.(2018)Kim, Wiseman, Miller, Sontag and Rush}]{Kim18}
	Kim Y, Wiseman S, Miller AC, Sontag D and Rush AM (2018) Semi-amortized
	variational autoencoders.
	\newblock In: \emph{Proceedings of the International Conference on Machine
		Learning (ICML)}.
%
	\bibitem[{Kingma and Welling(2014)}]{Kingma14}
	Kingma DP and Welling M (2014) Auto-encoding variational bayes.
	\newblock In: \emph{Proceedings of the International Conference on Learning
		Representations (ICLR)}.
%
	\bibitem[{Kober and Peters(2011)}]{Kober11}
	Kober J and Peters J (2011) Policy search for motor primitives in robotics.
	\newblock \emph{Machine Learning} 84: 171--203.
%
	\bibitem[{Koert et~al.(2016)Koert, Maeda, Lioutikov, Neumann and
		Peters}]{Koert16}
	Koert D, Maeda G, Lioutikov R, Neumann G and Peters J (2016) Demonstration
	based trajectory optimization for generalizable robot motions.
	\newblock In: \emph{Proceedings of the International Conference on Humanoid
		Robots (Humanoids)}.
%
	\bibitem[Kumar et~al.(2020)Kumar, Kumar, Levine, and Finn]{Kumar20}
	Kumar, S., Kumar, A., Levine, S., and Finn, C.
	\newblock One solution is not all you need:few-shot extrapolation via
	structured maxent rl.
	\newblock In \emph{Advances in Neural Information Processing Systems
		(NeurIPS)}, 2020.
%
	\bibitem[{LaValle and Kuffner(2001)}]{LaValle01}
	LaValle SM and Kuffner JJ (2001) Randomized kinodynamic planning.
	\newblock \emph{The International Journal of Robotics Research} .
%
	\bibitem[{Li et~al.(2017)Li, Song and Ermon}]{Li17}
	Li Y, Song J and Ermon S (2017) Infogail: Interpretable imitation learning
	fromvisual demonstrations.
	\newblock In: \emph{Advances in Neural Information Processing Systems (NIPS)}.
%
	\bibitem[{LeCun et~al.(2006)}]{LeCun06}
	LeCun Y, Chopra S, Hadsell R, Ranzato M, and Huang F (2006) A tutorial on energy-based
	learning. 
	\emph{Predicting structured data}
	\newblock MIT Press.
%
	\bibitem[{M.~A.~Rana and Ratliff(2020)}]{Rana20}
	M~A~Rana HRMMSCDFBB A~Li and Ratliff N (2020) Learning reactive motion policies
	in multiple task spaces from human demonstrations.
	\newblock In: \emph{Proceedings of the Conference on Robot Learning}. pp.
	1457--1468.
%
	\bibitem[{Maddison et~al.(2017)Maddison, Mnih and Teh}]{Maddison17}
	Maddison CJ, Mnih A and Teh YW (2017) The concrete distribution: A continuous
	relaxation of discrete random variables.
	\newblock In: \emph{Proceedings of the International Conference on Learning
		Representations (ICLR)}.
%
	\bibitem[{Merel et~al.(2019)Merel, Hasenclever, Galashov, Ahuja, Pham,
		G.~Wayne~and and Heess}]{Merel19}
	Merel J, Hasenclever L, Galashov A, Ahuja A, Pham V, G~Wayne~and YWT and Heess
	N (2019) Neural probabilistic motor primitives for humanoid control.
	\newblock In: \emph{Proceedings of the International Conference on Learning
		Representations (ICLR)}.
%
	\bibitem[{Mukadam et~al.(2020)Mukadam, Cheng, Fox, Boots and
		Ratliff}]{Mukadam20}
	Mukadam M, Cheng C, Fox D, Boots B and Ratliff N (2020) Riemannian motion
	policy fusion through learnable lyapunov function reshaping.
	\newblock In: \emph{Proceedings of the Conference on Robot Learning}. pp.
	204--219.
%
	\bibitem[{Mukadam et~al.(2018)Mukadam, Dong, Yan, Dellaert and
		Boots}]{Mukadam18}
	Mukadam M, Dong J, Yan X, Dellaert F and Boots B (2018) Continuous-time
	gaussian process motion planning via probabilistic inference.
	\newblock \emph{The International Journal of Robotics Research} .
%
	\bibitem[{Nachum et~al.(2018)Nachum, Gu, Lee and Levine}]{Nachum18}
	Nachum O, Gu S, Lee H and Levine S (2018) Data-efficient hierarchical
	reinforcement learning.
	\newblock In: \emph{Advances in Neural Information Processing Systems
		(NeurIPS)}.
%
	\bibitem[{Nachum et~al.(2019)Nachum, Gu, Lee and Levine}]{Nachum19}
	Nachum O, Gu S, Lee H and Levine S (2019) Near optimal representation learning
	for hierarchical reinforcement learning.
	\newblock In: \emph{Proceedings of the International Conference on Learning
		Representations (ICLR)}.
%
	\bibitem[{Orthey et~al.(2020)Orthey, Fr\'esz and Toussaint}]{Orthey20}
	Orthey A, Fr\'esz B and Toussaint M (2020) Motion planning explorer:
	Visualizing localminima using a local-minima tree.
	\newblock \emph{IEEE Robotics and Automation Letters} 5(2): 346--353.
%
	\bibitem[{Osa(2020)}]{osa20}
	Osa T (2020) Multimodal trajectory optimization for motion planning.
	\newblock \emph{The International Journal of Robotics Research} .
%
	\bibitem[{Osa et~al.(2017)Osa, Ghalamzan, Stolkin, Lioutikov, Peters and
		Neumann}]{Osa17}
	Osa T, Ghalamzan EAM, Stolkin R, Lioutikov R, Peters J and Neumann G (2017)
	Guiding trajectory optimization by demonstrated distributions.
	\newblock \emph{IEEE Robotics and Automation Letters} .
%
	\bibitem[{Osa et~al.(2018)Osa, Pajarinen, Neumann, Bagnell, Abbeel and
		Peters}]{Osa18}
	Osa T, Pajarinen J, Neumann G, Bagnell JA, Abbeel P and Peters J (2018) An
	algorithmic perspective on imitation learning.
	\newblock \emph{Foundations and Trends® in Robotics} 7(1-2): 1--179.
%
	\bibitem[{Osa et~al.(2019)Osa, Tangkaratt and Sugiyama}]{Osa19}
	Osa T, Tangkaratt V and Sugiyama M (2019) Hierarchical reinforcement learning
	via advantage-weighted information maximization.
	\newblock In: \emph{Proceedings of the International Conference on Learning
		Representations (ICLR)}.
%
	\bibitem[{Osa et~al.(2021)Osa, Tangkaratt and Sugiyama}]{Osa21}
	Osa T, Tangkaratt V and Sugiyama M (2021) Discovering Diverse Solutions in Deep Reinforcement Learning.
	\newblock \emph{arXiv}.
%
	\bibitem[{Rana et~al.(2017)Rana, Mukadam, Ahmadzadeh, Chernova and
		Boots}]{Rana17}
	Rana MA, Mukadam M, Ahmadzadeh SR, Chernova S and Boots B (2017) Towards robust
	skill generalization: Unifying learning from demonstration and motion
	planning.
	\newblock In: \emph{Proceedings of the Conference on Robot Learning (CoRL)}.
%
	\bibitem[{Schaul et~al.(2015)Schaul, Horgan, Gregor,  and Silver}]{Schaul15}
	Schaul T, Horgan D, Gregor K,  and Silver D (2015) Universal value function
	approximators.
	\newblock In: \emph{Proceedings of the InInternational Conference on Machine
		Learning (ICML)}.
%
	\bibitem[{Schulman et~al.(2014)Schulman, Duan, Ho, Lee, Awwal, Bradlow, Pan,
		Patil, Goldberg and Abbeel}]{Schulman14}
	Schulman J, Duan Y, Ho J, Lee A, Awwal I, Bradlow H, Pan J, Patil S, Goldberg K
	and Abbeel P (2014) Motion planning with sequential convex optimization and
	convex collision checking.
	\newblock \emph{The International Journal of Robotics Research} 33(9):
	1251--1270.
%
	\bibitem[{Sharma et~al.(2019)Sharma, Sharma, Rhinehart and Kitani}]{Sharma19}
	Sharma M, Sharma A, Rhinehart N and Kitani KM (2019) Directed-info gail:
	Learning hierarchical policies from unsegmented demonstrations using directed
	information.
	\newblock In: \emph{Proceedings of the International Conference on Learning
		Representations (ICLR)}.
%
	\bibitem[{Srinivas et~al.(2018)Srinivas, Jabri, Abbeel, Levine and
		Finn}]{Srinivas18}
	Srinivas A, Jabri A, Abbeel P, Levine S and Finn C (2018) Universal planning
	networks.
	\newblock In: \emph{Proceedings of the International Conference on Machine
		Learning (ICML)}.
%
	\bibitem[{Stoean et~al.(2010)Stoean, Preuss, Stoean and Dumitrescu}]{Stoean10}
	Stoean C, Preuss M, Stoean R and Dumitrescu D (2010) Multimodal optimization by
	means of a topological speciesconservation algorithm.
	\newblock \emph{IEEE Transactions on EvolutionaryComputation} 14(6): 842--864.
%
	\bibitem[{Toussaint(2015)}]{Toussaint15}
	Toussaint M (2015) Logic-geometric programming: Anoptimization-based approach
	to combined task and motionplanning.
	\newblock In: \emph{Proceedings of the International Joint Conference on
		ArtificialIntelligence (IJCAI)}.
%
	\bibitem[{Toussaint et~al.(2018)Toussaint, Allen, Smith and
		Tenenbaum}]{Toussaint18}
	Toussaint M, Allen KR, Smith KA and Tenenbaum JB (2018) Differentiable physics
	and stable modes for tool-use and manipulation planning.
	\newblock In: \emph{Proceedings of Robotics: Sciences and Systems (R:SS)}.
%
	\bibitem[{Toussaint et~al.(2020)Toussaint, Ha and Driess}]{Toussaint20}
	Toussaint M, Ha J and Driess D (2020) Describing physics for physical
	reasoning: Force-based sequential manipulation planning.
	\newblock \emph{IEEE Robotics and Automation Letters} .
%
	\bibitem[{van~der Maaten and Hinton(2008)}]{Maaten08}
	van~der Maaten L and Hinton G (2008) Visualizing data using {t-SNE}.
	\newblock \emph{Journal of Machine Learning Research} 9: 2579--2605.
	%
	\bibitem[{Vezhnevets et~al.(2017)Vezhnevets, Osindero, Schaul, Heess,
		Jaderberg, Silver and Kavukcuoglu}]{Vezhnevets17}
	Vezhnevets AS, Osindero S, Schaul T, Heess N, Jaderberg M, Silver D and
	Kavukcuoglu K (2017) Fe{U}dal networks for hierarchical reinforcement
	learning.
	\newblock In: \emph{Proceedings of the International Conference on Machine
		Learning (ICML)}.	
%
	\bibitem[{Verma et~al.(2019)}]{Verma19}
	Verma V, Lamb A, Beckham C, Najafi A, Mitliagkas I, Lopez-Paz D and Bengio Y (2019) 
	Manifold Mixup: Better Representations by Interpolating Hidden States.
	\newblock In: \emph{Proceedings of the International Conference on Machine
		Learning (ICML)}.	
%
	\bibitem[{Zucker et~al.(2013)Zucker, Ratliff, Dragan, Pivtoraiko, Klingensmith,
		Dellin, Bagnell and Srinivasa}]{Zucker13}
	Zucker M, Ratliff N, Dragan A, Pivtoraiko M, Klingensmith M, Dellin C, Bagnell
	JA and Srinivasa S (2013) Chomp: Covariant hamiltonian optimization for
	motion planning.
	\newblock \emph{The International Journal of Robotics Research} 32: 1164--1193.
\end{thebibliography}

\appendix


\section{Condition Number of Feature Matrix} 
\label{sec:cond_num}
To make the paper self-contained, we describe how the condition number of $\vect{\Phi}$ indicates the numerical stability of modeling trajectories with a neural network.
The condition number of a matrix $\vect{A}$ is defined as
\begin{align}
\kappa(\vect{A}) = \left\| \vect{A} \right\| \cdot \left\| \vect{A}^{-1} \right\|
\end{align}
for which the value depends on the norm we use.
In the following discussion, we assume that we use the $\ell_2$-norm.

We consider the problem of estimating $\vect{w}$ and compute $\vect{\xi} = \Phi \vect{w}$.
This computation appears in trajectory generation with DMPs and ProMPs.
When the error of estimating $\vect{w}$ is given by $\Delta\vect{w}$, 
the relative error of estimating $\vect{w}$, which is the ratio of estimation error $\Delta \vect{w}$ and the true value of $\vect{w}$,  is given by 
\[
r_w = \frac{ \left\| \Delta \vect{w} \right\|_2}{  \left\| \vect{w} \right\|_2} .
\]
Likewise, we denote by $\Delta \vect{\xi}$ the error of estimating $\vect{\xi}$, and the relative error of estimating $\vect{\xi}$ is given by
\[
r_\vect{\xi} = \frac{ \left\| \Delta \vect{\xi} \right\|_2}{  \left\| \vect{\xi} \right\|_2} .
\]
The ratio of $r_w$ and $r_{\vect{\xi}}$ indicates how the error of estimating $\vect{w}$ is propagated to the error of estimating $\vect{\xi}$.
We can obtain the following relation using $\vect{\xi} = \vect{\Phi} \vect{w}$:
\begin{align}
\frac{ r_{\vect{\xi}}}{r_{\vect{w}}} 
& = \frac{ \left\| \Delta \vect{\xi} \right\|_2}{  \left\| \vect{\xi} \right\|_2}
\frac{  \left\| \vect{w} \right\|_2} { \left\| \Delta \vect{w} \right\|_2} 
= \frac{ \left\| \vect{\Phi} \Delta \vect{w} \right\|_2}{ \left\| \Delta \vect{w} \right\|_2} \frac{  \left\| \vect{w} \right\|_2}{  \left\| \vect{\Phi} \vect{w} \right\|_2} \\
&  \leq  \max  \frac{ \left\|  \vect{\Phi} \Delta \vect{w} \right\|_2}{ \left\| \Delta \vect{w} \right\|_2}
\max \frac{  \left\| \vect{\Phi}^{-1} \vect{w} \right\|_2}{  \left\| \vect{w} \right\|_2} \\
& =  \left\|  \vect{\Phi} \right\|_2 \cdot \left\| \vect{\Phi}^{-1}  \right\|_2  = \kappa(\Phi).
\end{align}
Therefore, condition number $\kappa(\Phi)$ is the upper bound of $r_{\vect{\xi}} / r_{\vect{w}}$.

\begin{table*}[tb]
	\caption{Condition number of $\vect{\Phi}$.}
	\label{tbl:cond}
	\begin{center}
		\begin{tabular}{ccccc}
			\toprule
			\makecell{$b_{\textrm{exp}}(t)$} & \makecell{1.40e+04 \\ ($h=0.005$)} & \makecell{6.86e+7 \\ ($h=0.01$)} & \makecell{ 1.22e+17 \\ ($h=0.1$) } &   \makecell{ 2.17e+17 \\ ($h=0.25$) }   \\
			\midrule
			\makecell{$b_{\textrm{log}}(t)$} & \makecell{3.06e+17 \\ ($\alpha$=5)} & \makecell{4.57e+11 \\ ($\alpha$=10)} &  \makecell{ 1100 \\ ($\alpha$=50) } &  \makecell{140 \\ ($\alpha$=100)} \\
			\bottomrule
		\end{tabular}
	\end{center}
	\hspace{5.5cm}\footnotesize{We used $T=50$ and $B=30$. $\kappa(\vect{\Phi})$ depends on $T$ and $B$.}
\end{table*}

Table~\ref{tbl:cond} shows a comparison of the condition number $\kappa(\vect{\Phi})$ between the two basis functions.
When using the basis function $b_{\textrm{exp}}(t)$ for $\vect{\Phi}$, the condition number is $\kappa(\vect{\Phi}) \approx 10^{17}$ for $h=0.1$.
This large condition number indicates that an error in $w$ can be significantly magnified when estimating $\vect{\xi}$.
For example, suppose that the order of the value is given as $\left\| \vect{\xi} \right\|_2 \approx 10$ and $\left\| \vect{w} \right\|_2 \approx 10$. 
If the necessary precision for estimating $\vect{\xi}$ is $\left\| \Delta \vect{\xi} \right\|_2 \approx 0.1$, 
then the order of the estimation error of $\vect{w}$ should be $\left\| \Delta \vect{w} \right\|_2 \approx 10^{-18}$, which is problematic when we train a neural network to estimate $\vect{w}$ for planning $\vect{\xi}$.
The large condition number of $\vect{\Phi}$ may not be problematic when the solution can be obtained using a closed form as in DMPs and ProMPs 
because we can obtain a solution with high accuracy in a single matrix calculation.
However, when we train a neural network that estimates $\vect{w}$, minimizing the loss function with a stochastic gradient descent will require numerous iterations to achieve such high precision.

When using the basis function in \eqref{eq:logistic_basis} with $\alpha=50$ for the feature matrix, the condition number is $\kappa(\phi) \approx 1000$.
In the above example, the necessary precision is $\left\| \Delta \vect{w} \right\|_2 \approx 10^{-4}$, which is achievable in the training of a neural network.
We employed the basis function in \eqref{eq:logistic_basis} in our experiments with RTP, although other forms of the basis function can be used as long as the condition number is sufficiently small.

It is worth noting that scaling the feature matrix $\vect{\Phi}$ does not change the condition number as $\kappa(\vect{\Phi}) = \kappa(a\vect{\Phi})$ where $a$ is an arbitrary real number.
Therefore, the numerical instability caused by the large condition number $\kappa(\vect{\Phi})$ cannot be resolved by scaling the feature matrix $\vect{\Phi}$.

\section{Experiment Details}
\subsection{Conditions for training a neural network for tasks with synthetic test functions}

We provide the network architecture and training parameters for the tasks with synthetic test functions in Table~\ref{tbl:param_test}.
The implementation of VAE is based on the implementation provided by \cite{Dupont18}.

\begin{table}[b]
	\caption{Network architecture and training parameters for tasks with synthetic test functions.}
	\centering
	\begin{tabular}{lcl}
		\toprule
		Description     & Symbol     & Value \\
		\midrule
		\makecell[l]{Number of samples drawn \\ from the proposal distribution} & $N$ & 20000\\
		\makecell[l]{Coefficient for the \\ information capacity} & $\gamma$ & 0.1\\
		Learning rate & & 0.001 \\
		Batch size & & 250 \\
		Number of training epoch && 350 \\
		\makecell[l]{Number of units in hidden \\ layers in $q_{\vect{\psi}}(\vect{\xi}|z)$ }    &        & (64, 64)  \\
		\makecell[l]{Number of units in hidden \\ layers in $p_{\vect{\theta}}(z|\vect{\xi})$ }   &        & (64, 64)  \\
		Activation function     &        & Relu, Relu  \\
		optimizer & & Adam\\
		\bottomrule
	\end{tabular}
	\label{tbl:param_test}
\end{table}

\subsection{Definitions of test functions}
\label{app:test_func}
The definitions of the test functions used in the experiment are as follows.
The figures plot the range $x_1 \in [0, 2]$ and  $x_2 \in [0, 2]$.

The test function~1 is given by
\begin{align}
R(x_1, x_2) = \exp( - 2d )
\end{align}
where
{\small
\begin{align}
d = 
\left\{
\begin{array}{cll}
((x_2 - 1.05)^2 + (x_1 - 0.5)^2)^{0.5}, & \textrm{if} & x_1 < 0.5, \\
\frac{| -0.3 x_1 - x_2 + 1.2 |}{( 0.09 + 1 )^2}   , & \textrm{if} & 0.5 < x_1 < 1.5, \\
((x_2 - 0.75)^2 + (x_1 - 1.5)^2)^{0.5},& \textrm{if} & x_1 \geq 1.5.
\end{array}
\right. 
\end{align}
}

The toy function~2 is given by
\begin{align}
R(x_1, x_2) = \exp( - 2d )
\end{align}
where
\begin{align}
d = | (x_2 - 1.5)^2 + (x_1 + 1)^2 - 2.5 |.
\end{align}

The toy function~3 is given by
\begin{align}
R(x_1, x_2) = \exp\big( - 2(d + 0.2x_2 + 0.14) \big)
\end{align}
where
{\small
\begin{align}
d = 
\left\{
\begin{array}{cll}
((x_2 - 0.94)^2 + (x_1 - 0.7)^2)^{0.5}, & \textrm{if} & x_1 < 0.7, \\
\frac{| 0.2 x_1 - x_2 + 0.8 |}{( 0.04 + 1 )^2}   , & \textrm{if} & 0.7 < x_1 < 1.4, \\
((x_2 - 1.08)^2 + (x_1 - 1.4)^2)^{0.5},& \textrm{if} & x_1 \geq 1.4.
\end{array}
\right. 
\end{align}
}

The toy function~4 is given by
\begin{align}
R(x_1, x_2) = \exp( - 2d )
\end{align}
where
\begin{align}
d = | (x_2 - 1)^2 + (x_1 - 1)^2 - 0.5 |.
\end{align}

\begin{table}[tb]
	\caption{Network architecture and training parameters for motion planning tasks.}
	\centering
	\begin{tabular}{lcl}
		\toprule
		Description     & Symbol     & Value \\
		\midrule
		\makecell[l]{Number of time steps \\ in a trajectory} & $T$ & 50\\
		\makecell[l]{Number of basis funcs. \\ in RTP} & $B$ & 20\\
		\makecell[l]{{\small Number of samples drawn} \\ {\small from the proposal distribution} } & $N$ & \makecell[l]{4e3 (w/ RTP) \\ 2e4 (w/o RTP) }\\
		\makecell[l]{Coefficient for the \\ information capacity} & $\gamma$ & 10\\
		Learning rate & & 0.001 \\
		Batch size & & 250 \\
		Number of training epoch && 700 \\
		\makecell[l]{Number of units in hidden \\ layers in $q_{\vect{\psi}}(\vect{\xi}|z)$ }    &        & (300, 200)  \\
		\makecell[l]{Number of units in hidden \\ layers in $p_{\vect{\theta}}(z|\vect{\xi})$ }   &        & (200, 300)  \\
		Activation function     &        & Relu, Relu  \\
		optimizer & & Adam\\
		\bottomrule
	\end{tabular}
	\label{tbl:param_mp}
\end{table}

\subsection{Conditions for training a neural network for motion planning tasks}

We provide the network architecture and training parameters for the motion planning tasks in Table~\ref{tbl:param_test}.
The information capacity $C_{\vect{z}}$ in \eqref{eq:joint_loss} is linearly increased from 0 to 5 during the training, using the implementation of joint VAE provided by \cite{Dupont18}.


\section{Additional results on motion planning tasks}
\label{sec:app_exp}
\subsection{Solutions Found for Tasks~2 and 3}
\label{sec:app_sim}
Additional results on simulated environments are shown in Figure~\ref{fig:LSMO_task1_app}--\ref{fig:task3_task}.
The results of Tasks~2 and 3 support the discussion in the main manuscript.
For example, MPSM learned trajectories from different homotopic classes on Task~2.
As shown in Figure~\ref{fig:LSMO_task2}(a), the end-effector moves over the table if $z=-1.28$, whereas the end-effector moves behind the obstacle if $z=1.28$.
The trajectory changes continuously as the value of $z$ changes, but a discontinuous change in the trajectory occurs between $z=0.34$ and $z=0.72$, as shown in Fig.~\ref{fig:LSMO_task2}(a).
Actually, the distribution of solutions is disconnected after fine-tuning as shown in Fig.~\ref{fig:task2_t-sne}.
Figure~\ref{fig:task2_task} also implies that there are two clusters of solutions for Task~2.
Meanwhile, it was not necessary to fine-tune the output of $p_{\vect{\theta}}(\vect{\xi}|\vect{z})$ for Task~3, and the distribution of the solutions was continuously connected, as shown in Fig.~\ref{fig:task3_t-sne}.

\subsection{Effect of Dimensionality of Latent Variable}
\begin{figure*}[tb]
			\centering
			\begin{subfigure}[t]{2\columnwidth}
				\centering
				\includegraphics[width=\textwidth]{task1_1d}
				\caption{Result with one-dimensional latent variable. }
			\end{subfigure}
			\begin{subfigure}[t]{2\columnwidth}
				\centering
				\includegraphics[width=\textwidth]{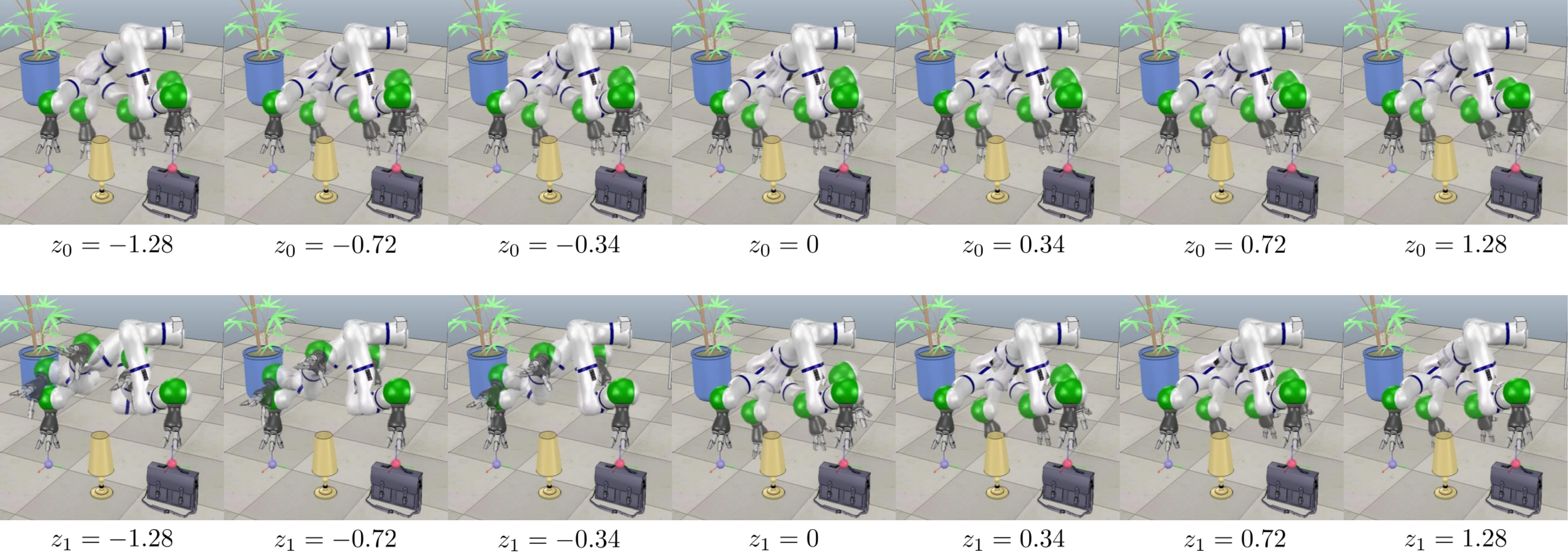}
				\caption{Result with two-dimensional latent variable. $z_1=0$ in top row, and $z_0=0$ in bottom row.}
			\end{subfigure}
			\caption{Solutions generated from $p_{\vect{\theta}}(\vect{\xi}|\vect{z})$ with different values of latent variable $\vect{z}$ on Task~1.  }
			\label{fig:LSMO_task1_app}
		\end{figure*}

\begin{figure}[tb]
			\centering
			\begin{subfigure}[t]{0.48\columnwidth}
				\centering
				\includegraphics[width=\textwidth]{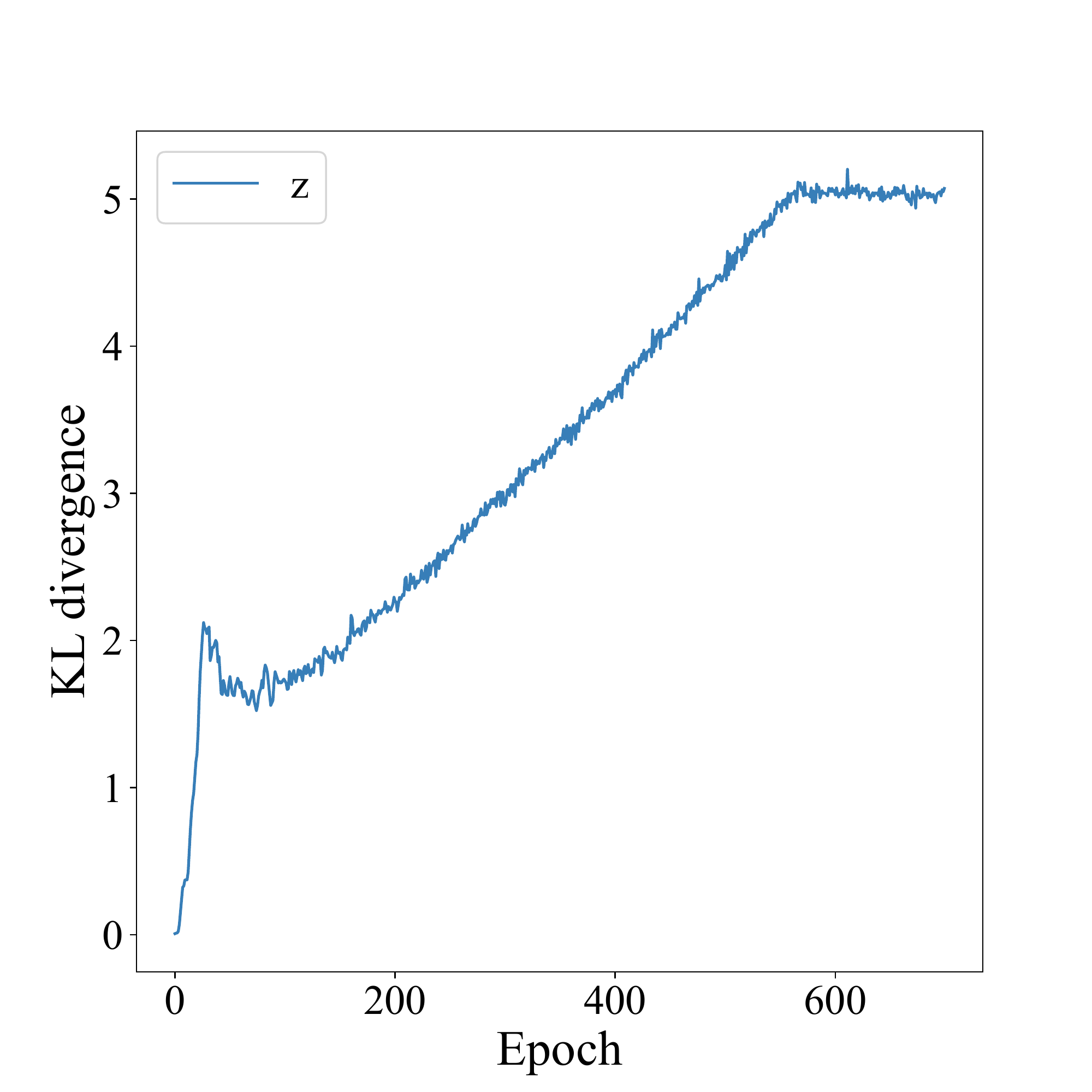}
				\caption{Results with one-dimensional latent variable. }
			\end{subfigure}
			\begin{subfigure}[t]{0.48\columnwidth}
				\centering
				\includegraphics[width=\textwidth]{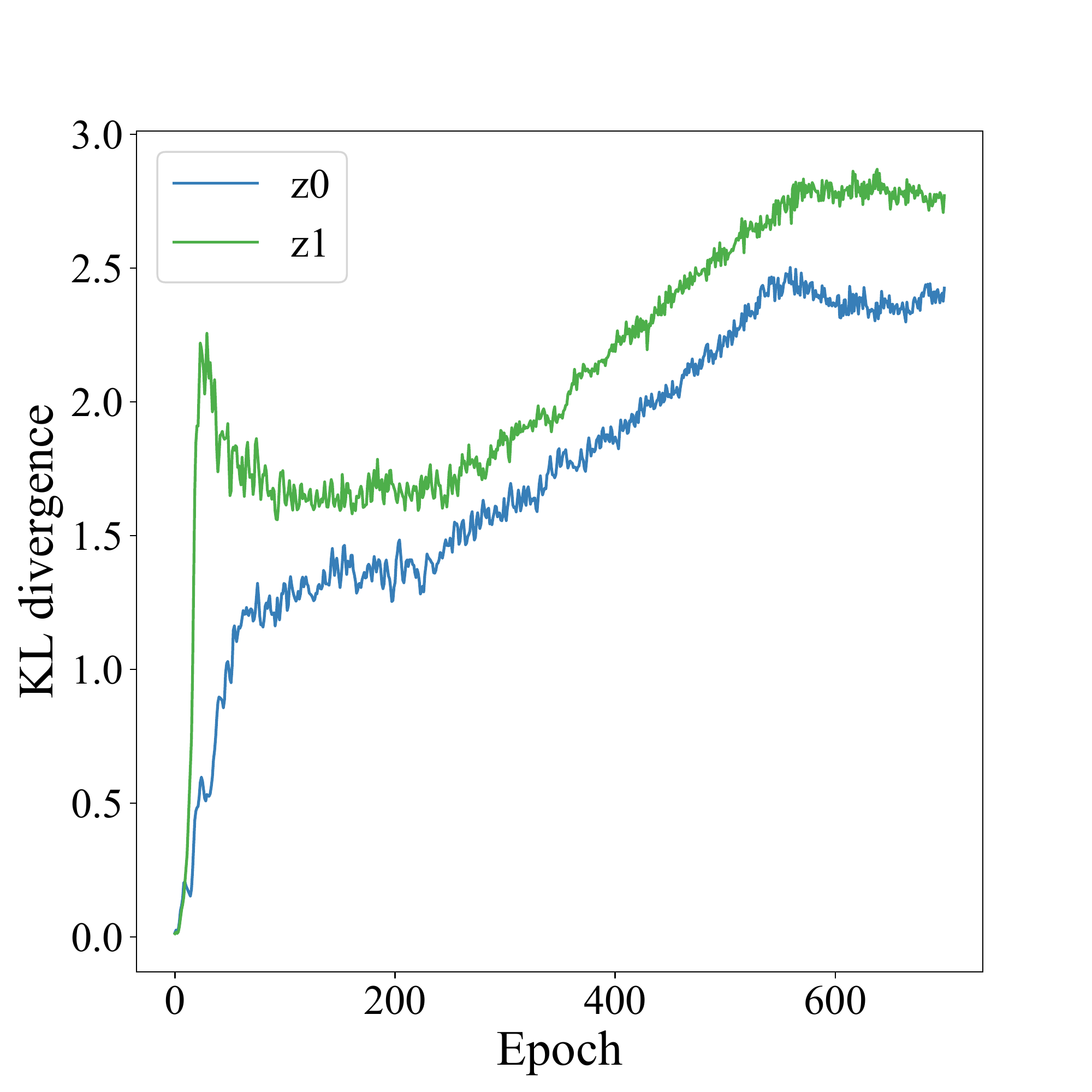}
				\caption{Results with two-dimensional latent variable. }
			\end{subfigure}
			\caption{KL divergence during training on Task~1. }
			\label{fig:task1_KL}
		\end{figure}

\begin{figure}[tb]
			\centering
			\begin{subfigure}[t]{0.48\columnwidth}
				\centering
				\includegraphics[width=\textwidth]{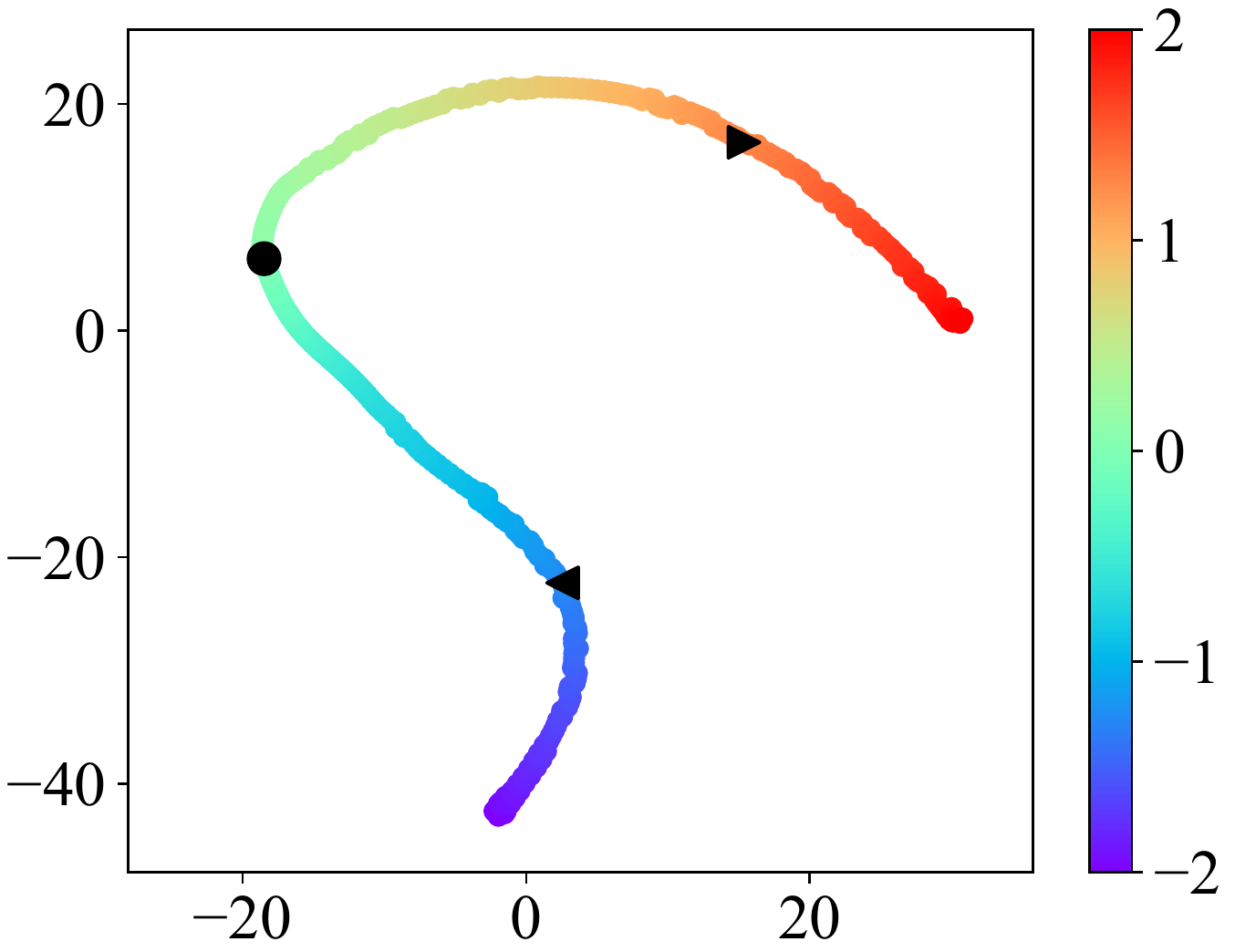}
				\caption{Distribution of solutions before fine-tuning. }
			\end{subfigure}
			\begin{subfigure}[t]{0.48\columnwidth}
				\centering
				\includegraphics[width=\textwidth]{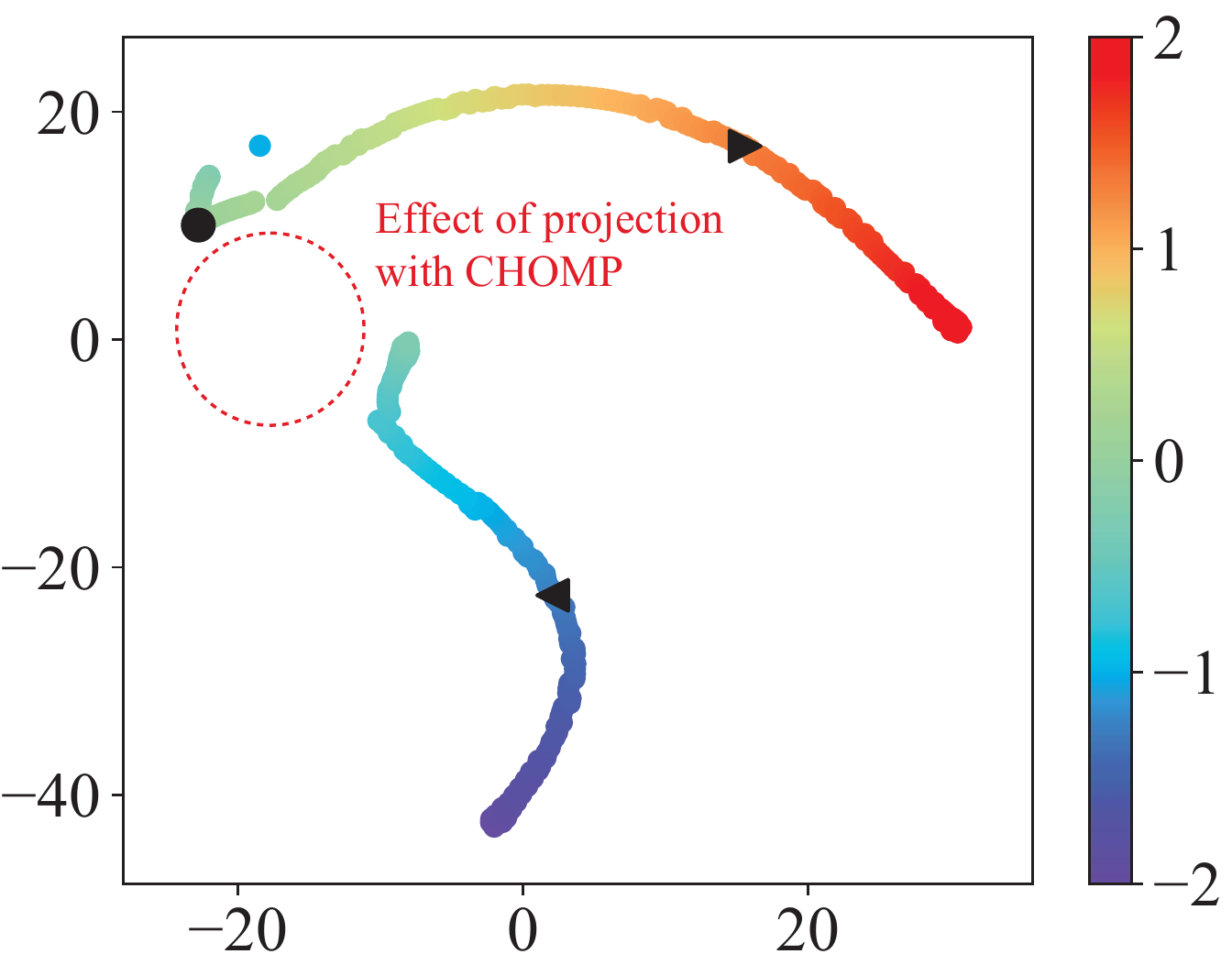}
				\caption{Distribution of solutions after fine-tuning. }
			\end{subfigure}
			\caption{Visualization of distribution of solutions on Task~1. Dimensionality is reduced using t-SNE. Color bar indicates value of $z$.  }
			\label{fig:task1_t-sne_app}
		\end{figure}

\begin{figure}[tb]
			\centering
			\begin{subfigure}[t]{0.32\columnwidth}
				\centering
				\includegraphics[width=\textwidth]{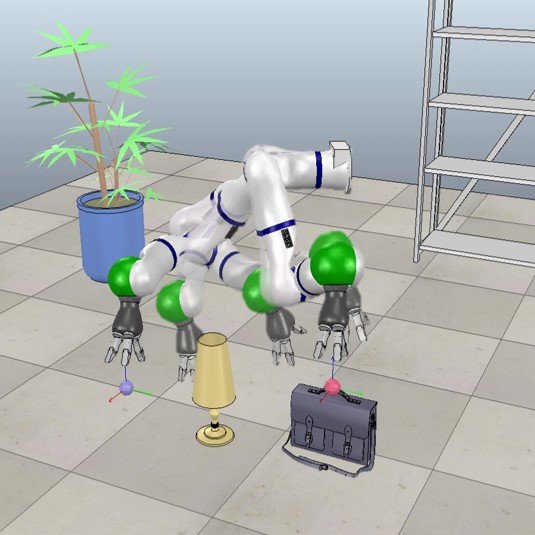}
			\end{subfigure}
			\begin{subfigure}[t]{0.32\columnwidth}
				\centering
				\includegraphics[width=\textwidth]{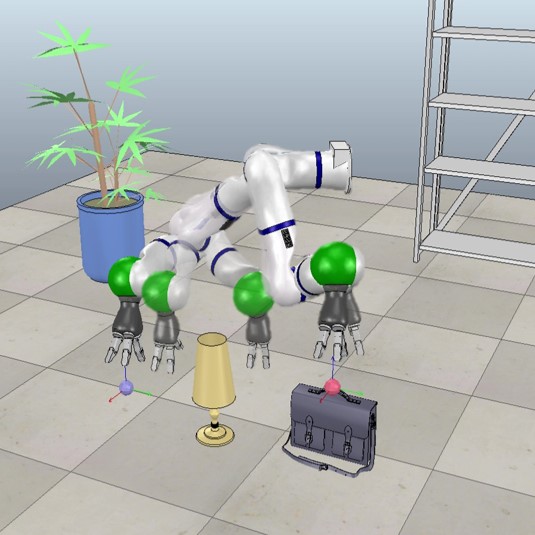}
			\end{subfigure}
			\begin{subfigure}[t]{0.32\columnwidth}
				\centering
				\includegraphics[width=\textwidth]{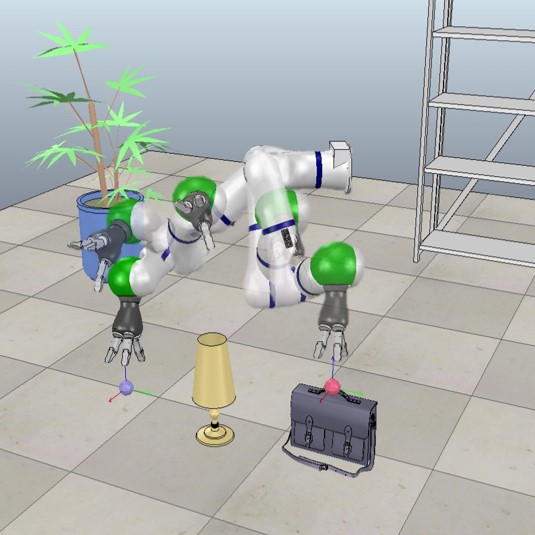}
			\end{subfigure}
			\caption{Three solutions found by SMTO for Task~1.  }
			\label{fig:task1_smto}
		\end{figure}

\begin{figure}[tb]
			\includegraphics[width=0.9\columnwidth]{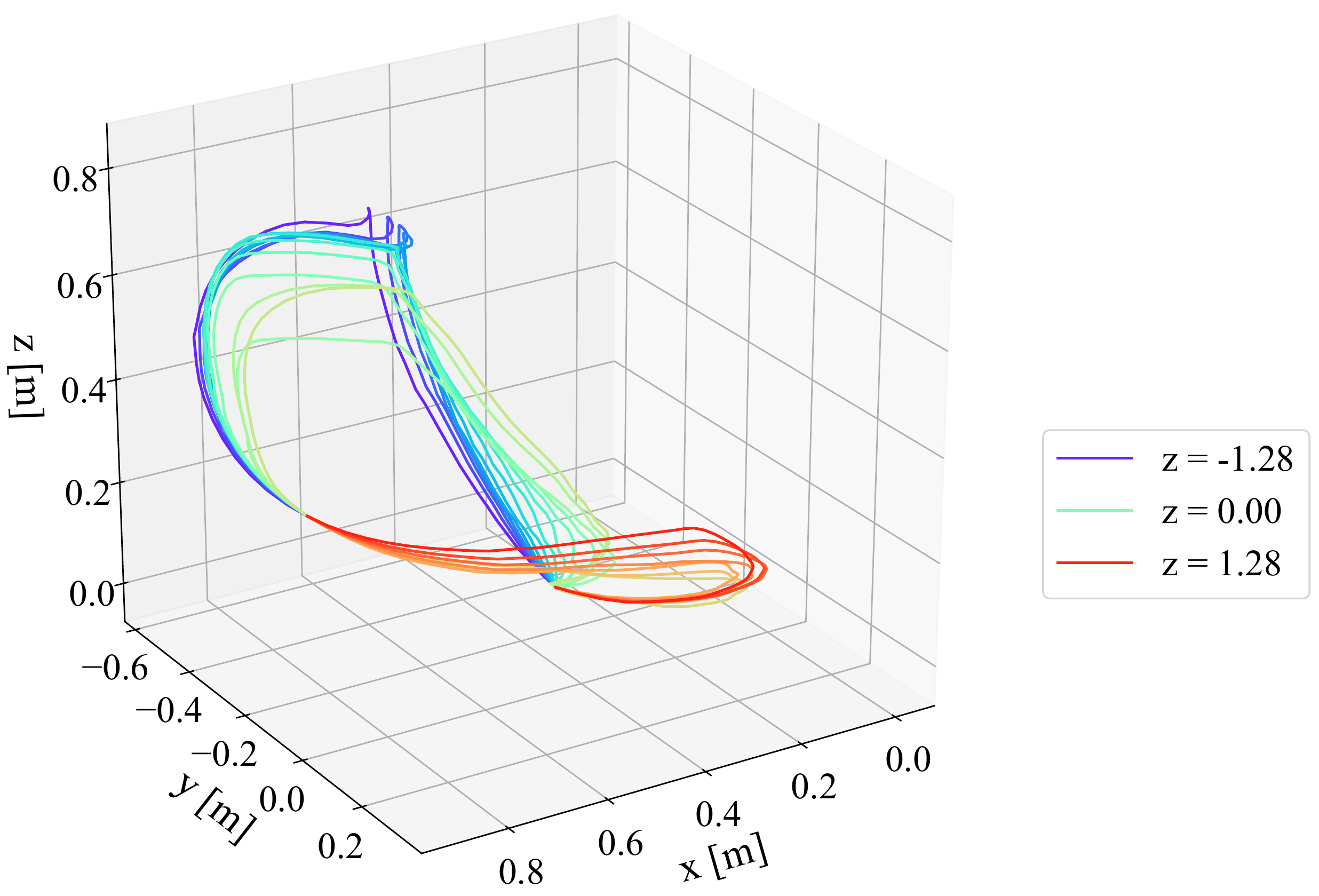}
			\caption{Trajectories in task space for Task~1. Result with one-dimensional latent variable. 20 trajectories are generated by linearly interpolating between $z = -1.28$ and $z = 1.28$. }
			\label{fig:task1_task_app}
		\end{figure}

\begin{figure*}[tb]
			\centering
			\begin{subfigure}[t]{2\columnwidth}
				\centering
				\includegraphics[width=\textwidth]{task2_1d}
				\caption{Result with one-dimensional latent variable. }
			\end{subfigure}
			\begin{subfigure}[t]{2\columnwidth}
				\centering
				\includegraphics[width=\textwidth]{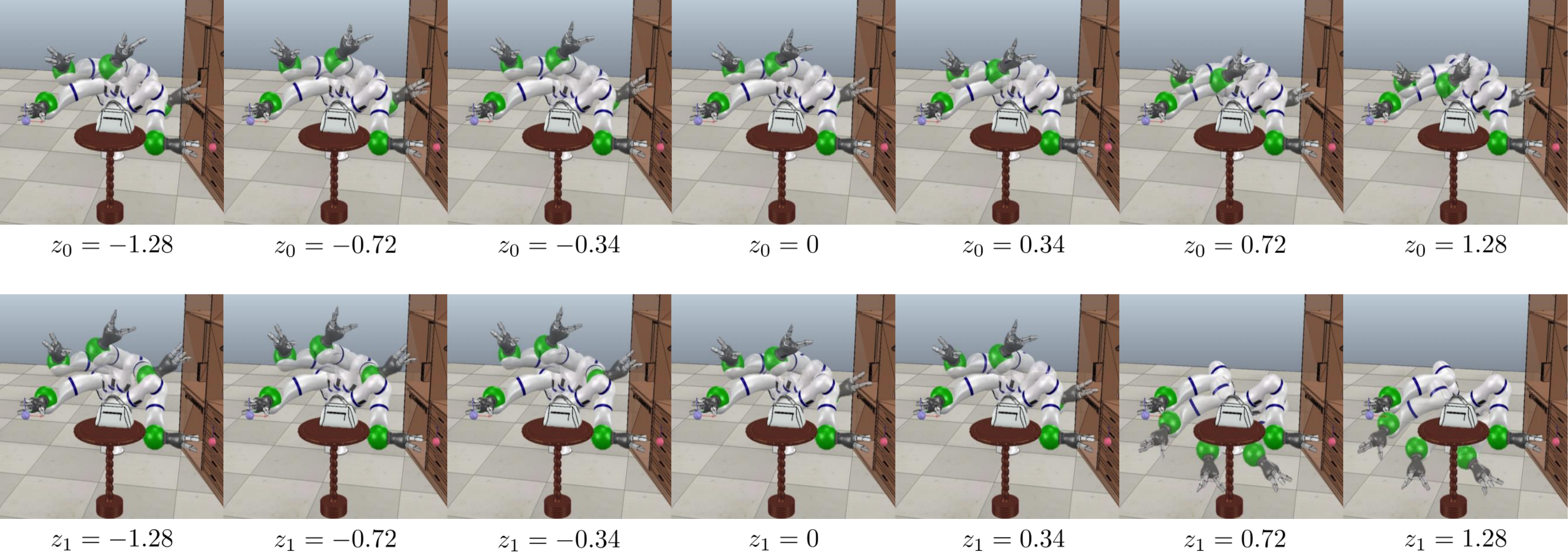}
				\caption{Result with two-dimensional latent variable. $z_1=0$ in top row, and $z_0=0$ in bottom row.}
			\end{subfigure}
			\caption{Solutions generated from $p_{\vect{\theta}}(\vect{\xi}|\vect{z})$ with different values of latent variable $\vect{z}$ on Task~2.  }
			\label{fig:LSMO_task2}
		\end{figure*}

\begin{figure}[tb]
			\centering
			\begin{subfigure}[t]{0.48\columnwidth}
				\centering
				\includegraphics[width=\textwidth]{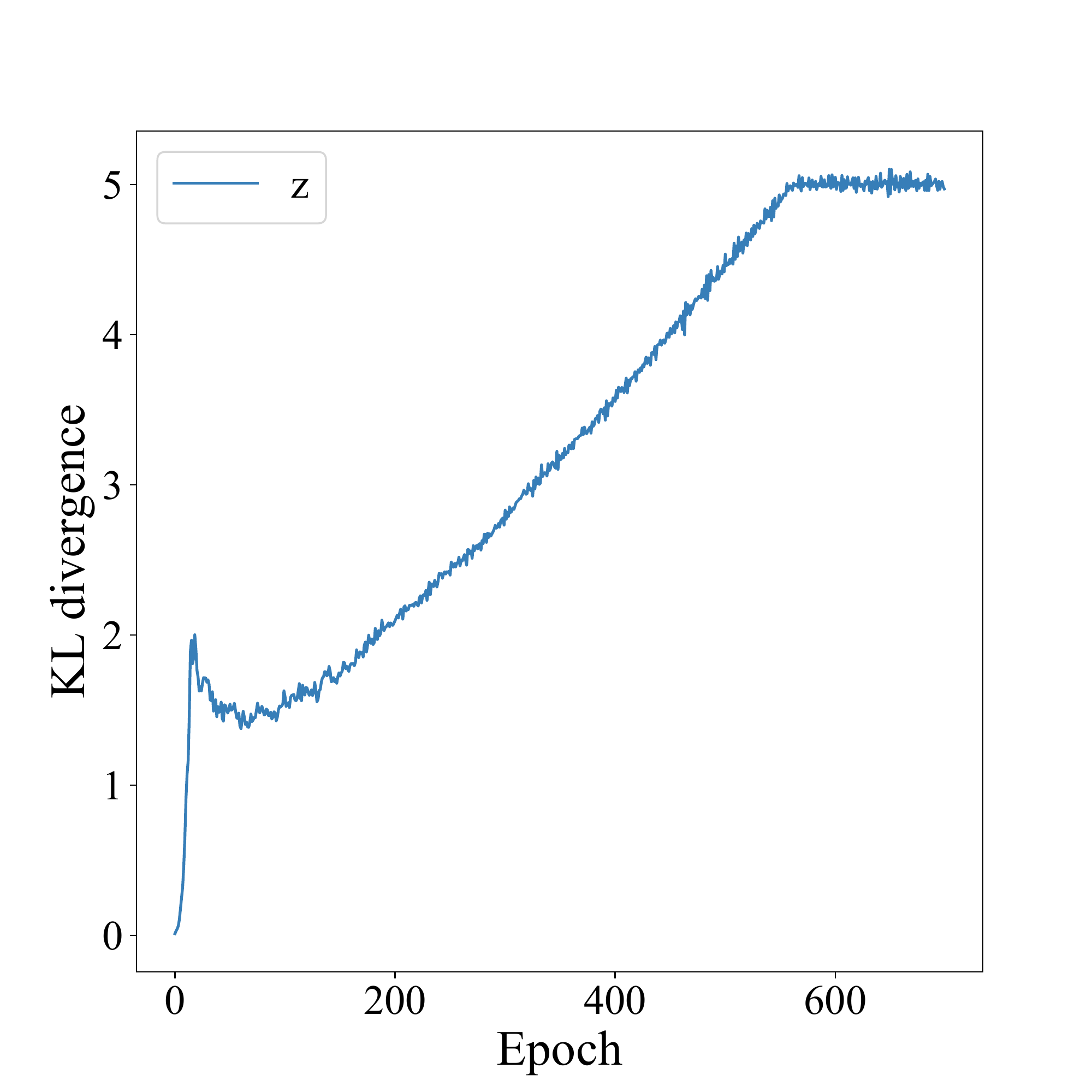}
				\caption{Results with one-dimensional latent variable. }
			\end{subfigure}
			\begin{subfigure}[t]{0.48\columnwidth}
				\centering
				\includegraphics[width=\textwidth]{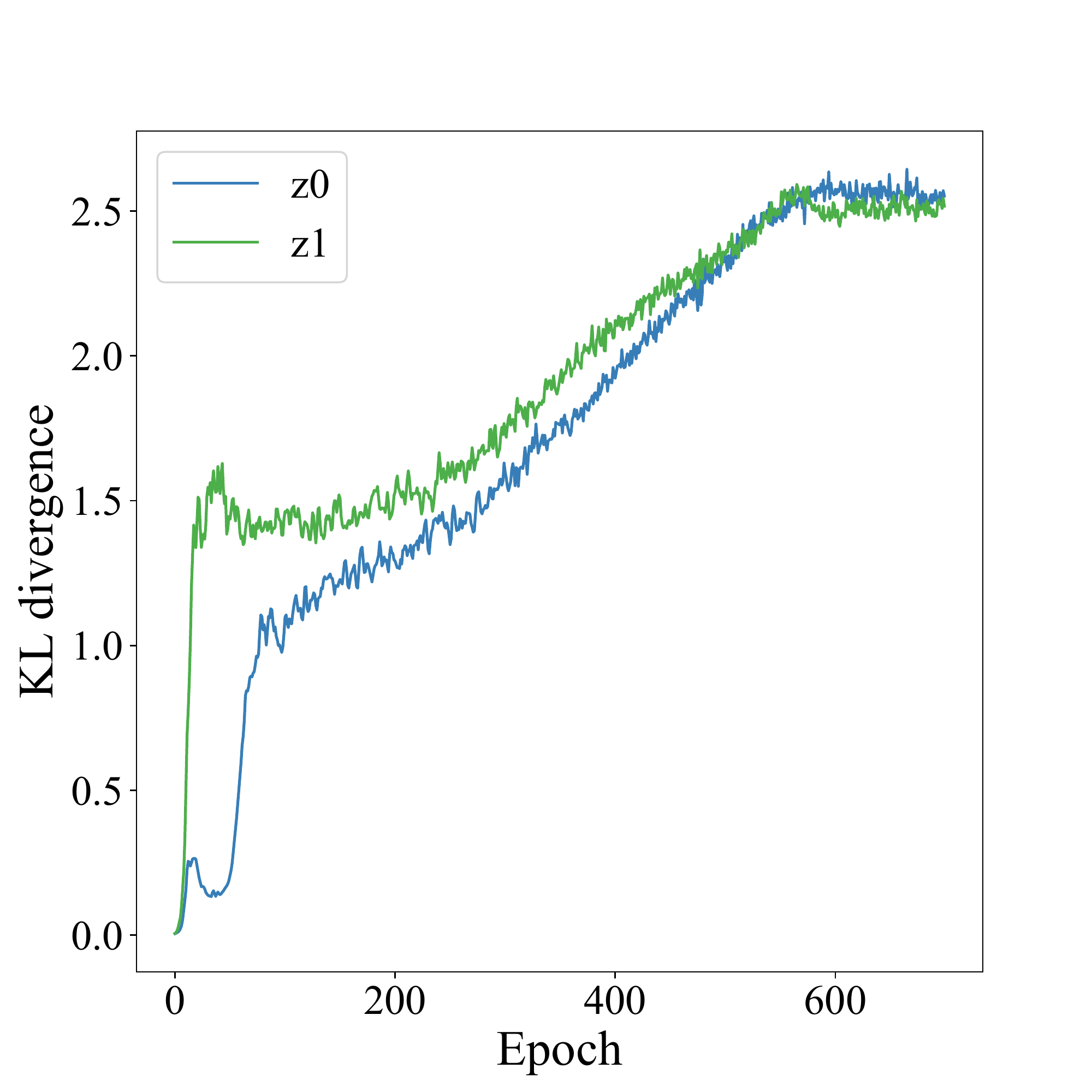}
				\caption{Results with two-dimensional latent variable.  }
			\end{subfigure}
			\caption{KL divergence during training on Task~2. }
			\label{fig:task2_KL}
		\end{figure}

\begin{figure}[tb]
			\centering
			\begin{subfigure}[t]{0.48\columnwidth}
				\centering
				\includegraphics[width=\textwidth]{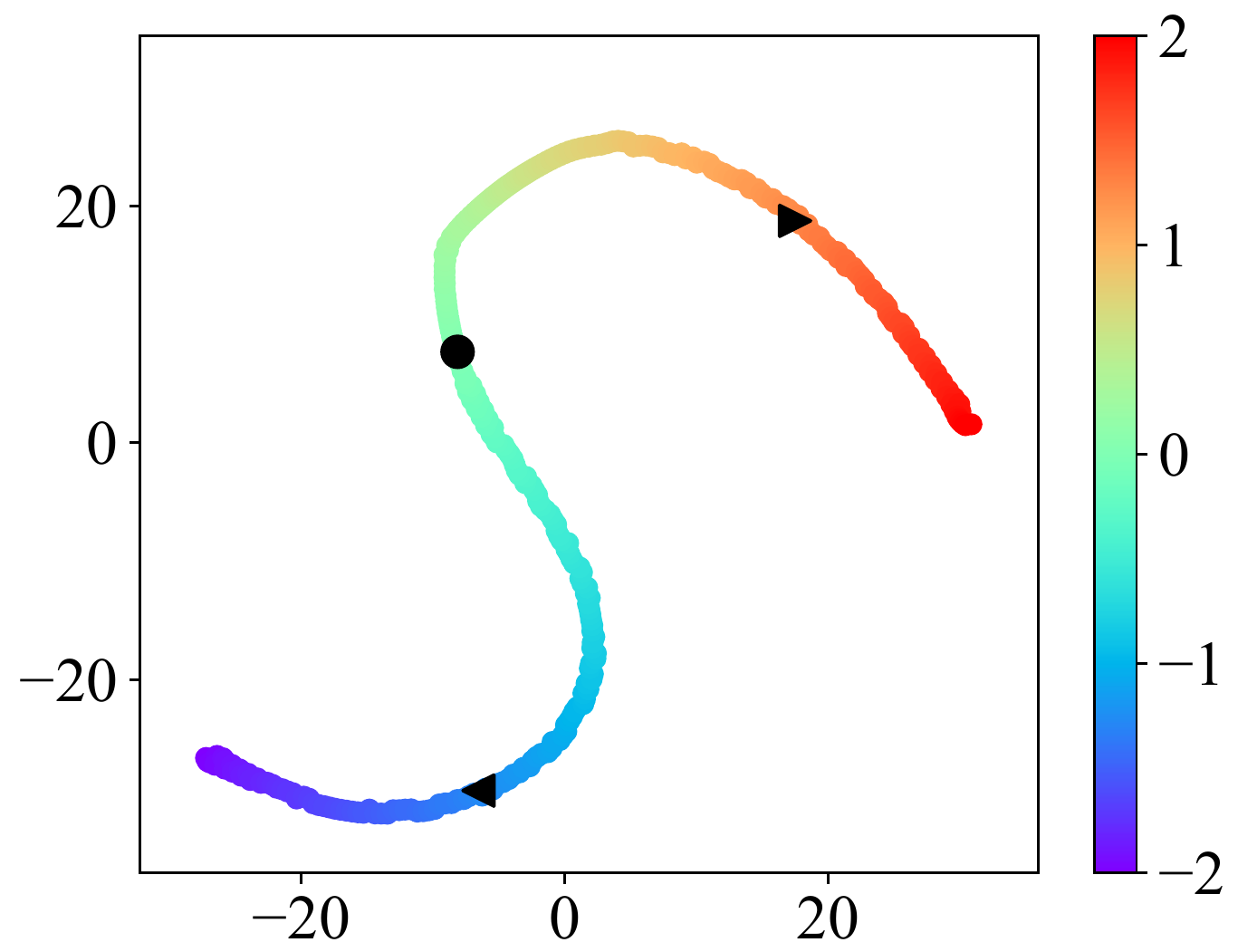}
				\caption{Distribution of solutions before fine-tuning. }
			\end{subfigure}
			\begin{subfigure}[t]{0.48\columnwidth}
				\centering
				\includegraphics[width=\textwidth]{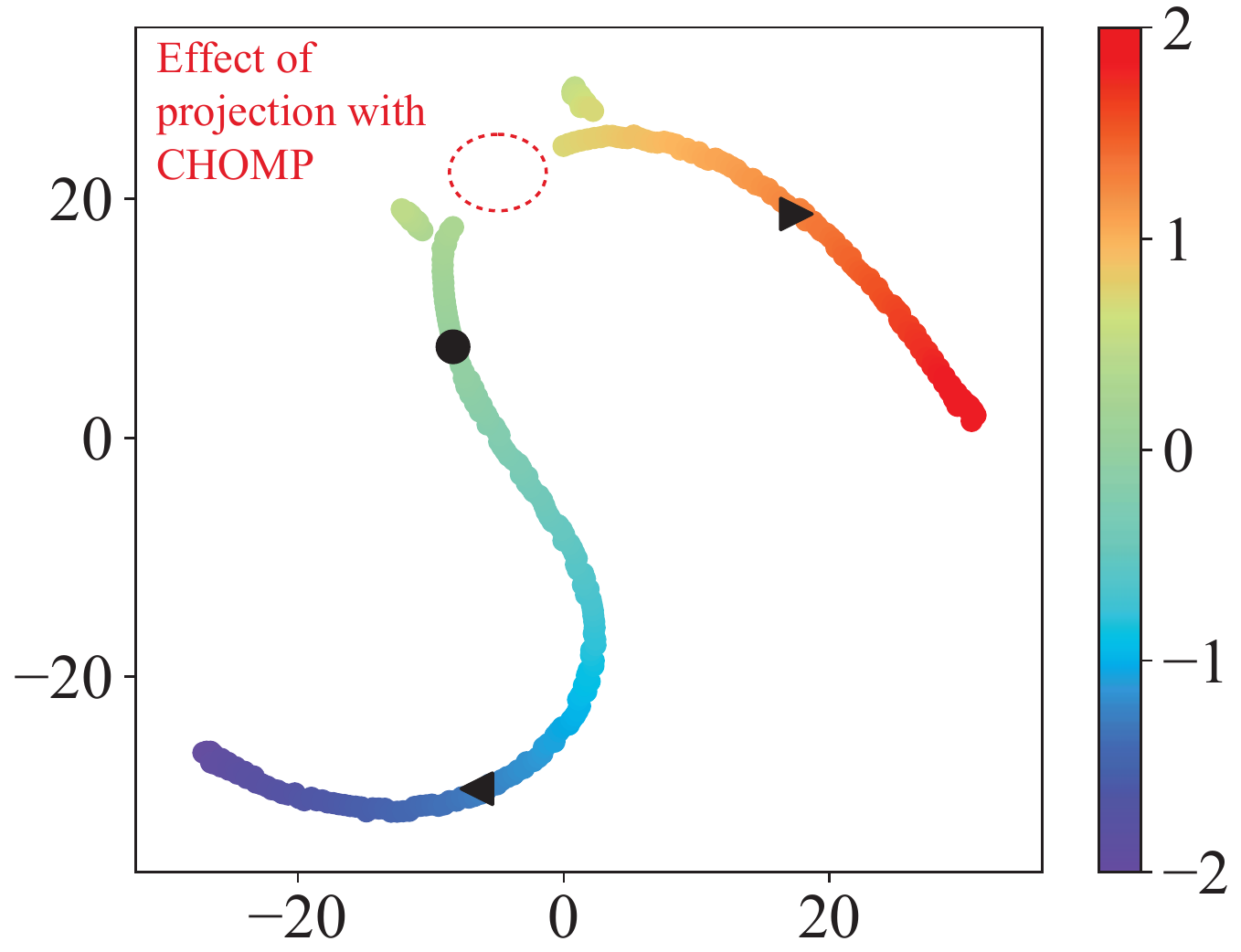}
				\caption{Distribution of solutions after fine-tuning. }
			\end{subfigure}
			\caption{Visualization of distribution of solutions on Task~2. Dimensionality is reduced using t-SNE. Color bar indicates value of $z$.  }
			\label{fig:task2_t-sne}
		\end{figure}

\begin{figure}[tb]
			\centering
			\begin{subfigure}[t]{0.42\columnwidth}
				\centering
				\includegraphics[width=\textwidth]{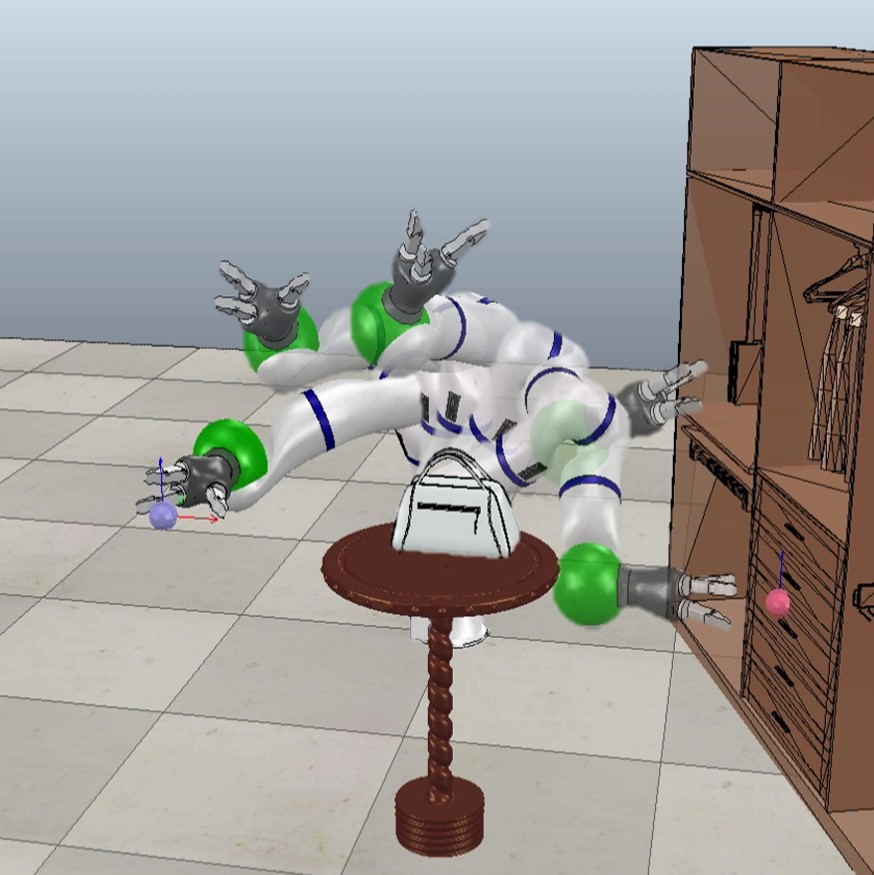}
			\end{subfigure}
			\begin{subfigure}[t]{0.42\columnwidth}
				\centering
				\includegraphics[width=\textwidth]{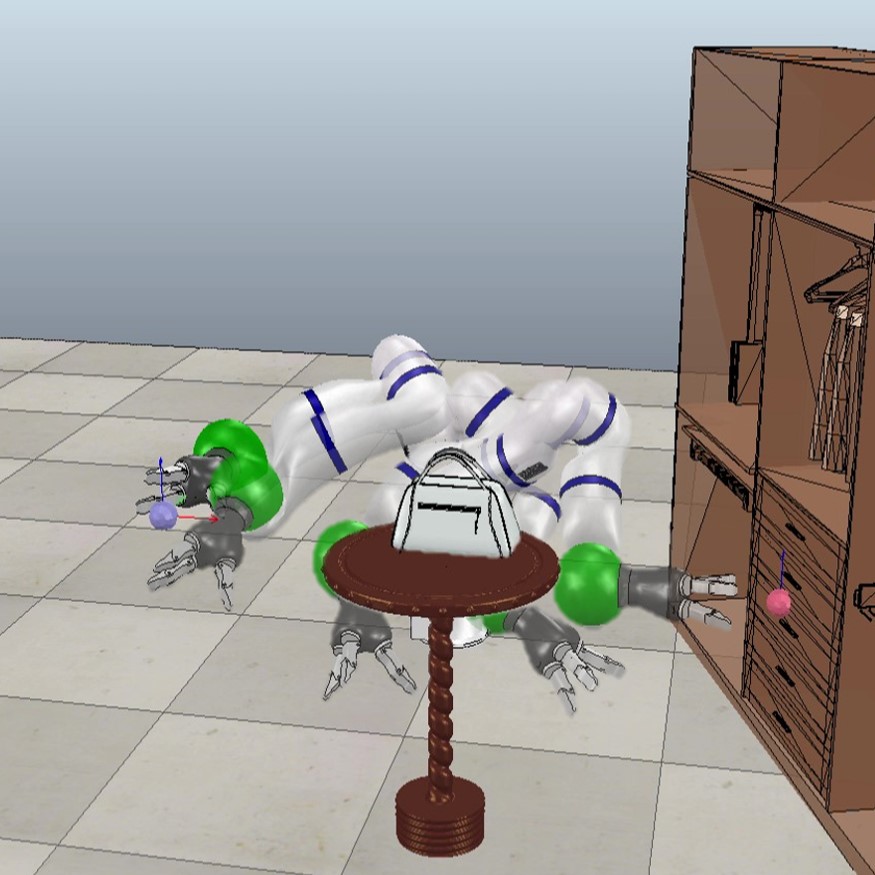}
			\end{subfigure}
			\caption{Two solutions found by SMTO for Task~2.  }
			\label{fig:task2_smto}
		\end{figure}

\begin{figure}[tb]
			\hfill
			\includegraphics[width=0.9\columnwidth]{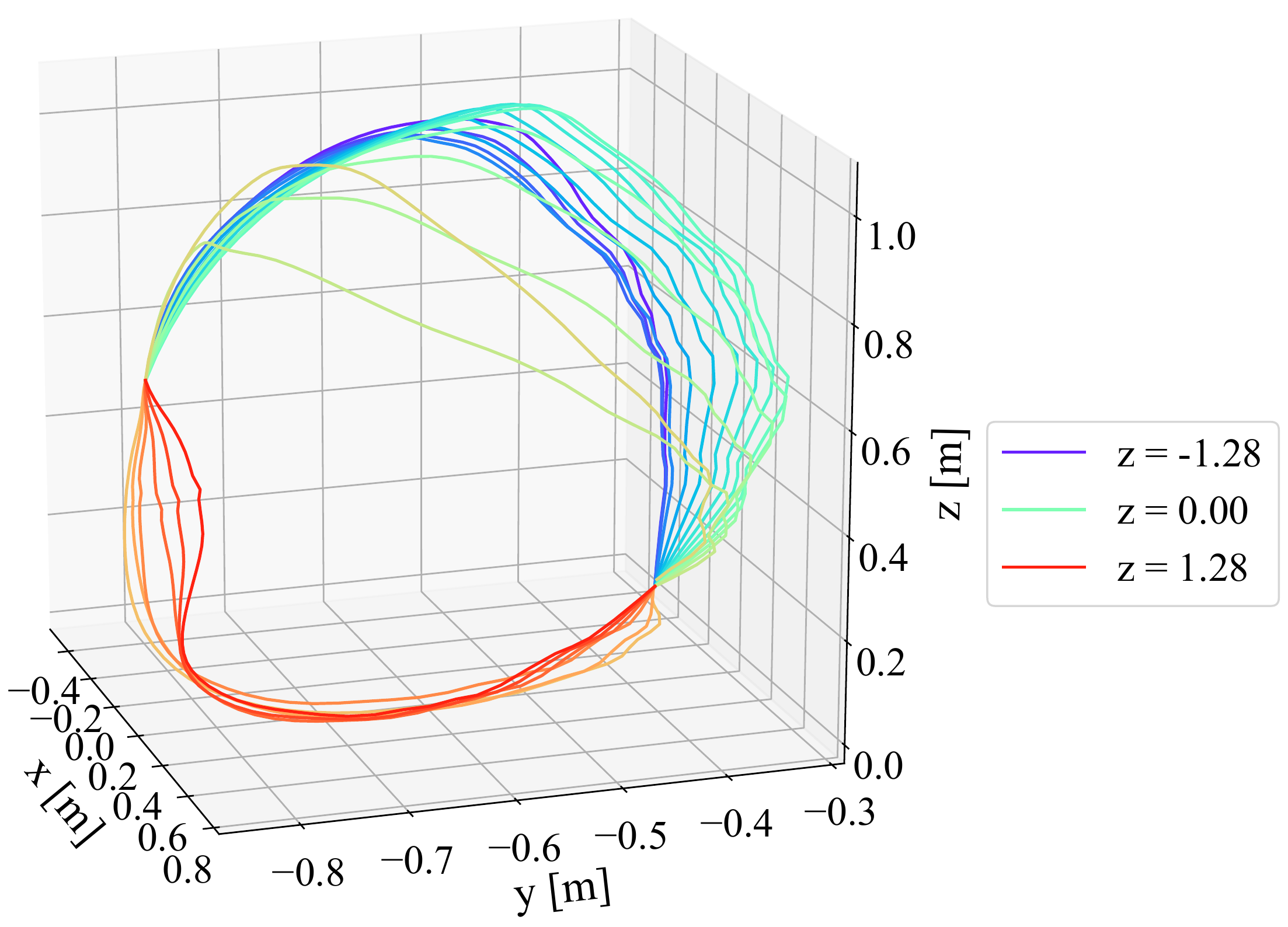}
			\caption{Trajectories in task space for Task~2. Result with one-dimensional latent variable. 20 trajectories are generated by linearly interpolating between $z = -1.28$ and $z = 1.28$.  }
			\label{fig:task2_task}
		\end{figure}

\begin{figure*}[tb]
			\centering
			\begin{subfigure}[t]{2\columnwidth}
				\centering
				\includegraphics[width=\textwidth]{task3_1d}
				\caption{Result with one-dimensional latent variable. }
			\end{subfigure}
			\begin{subfigure}[t]{2\columnwidth}
				\centering
				\includegraphics[width=\textwidth]{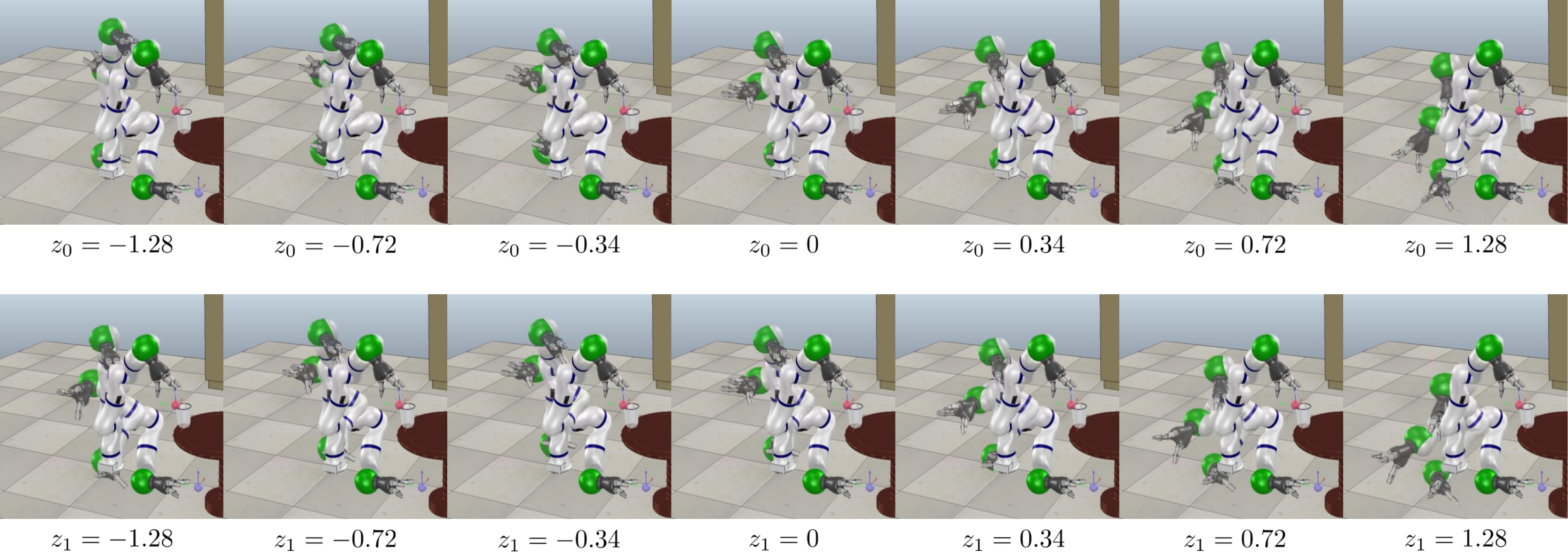}
				\caption{Result with two-dimensional latent variable. $z_1=0$ in top row, and $z_0=0$ in bottom row.}
			\end{subfigure}
			\caption{Solutions generated from $p_{\vect{\theta}}(\vect{\xi}|\vect{z})$ with different values of latent variable $\vect{z}$ on Task~3.  }
			\label{fig:LSMO_task3}
		\end{figure*}

\begin{figure}[tb]
			\centering
			\begin{subfigure}[t]{0.48\columnwidth}
				\centering
				\includegraphics[width=\textwidth]{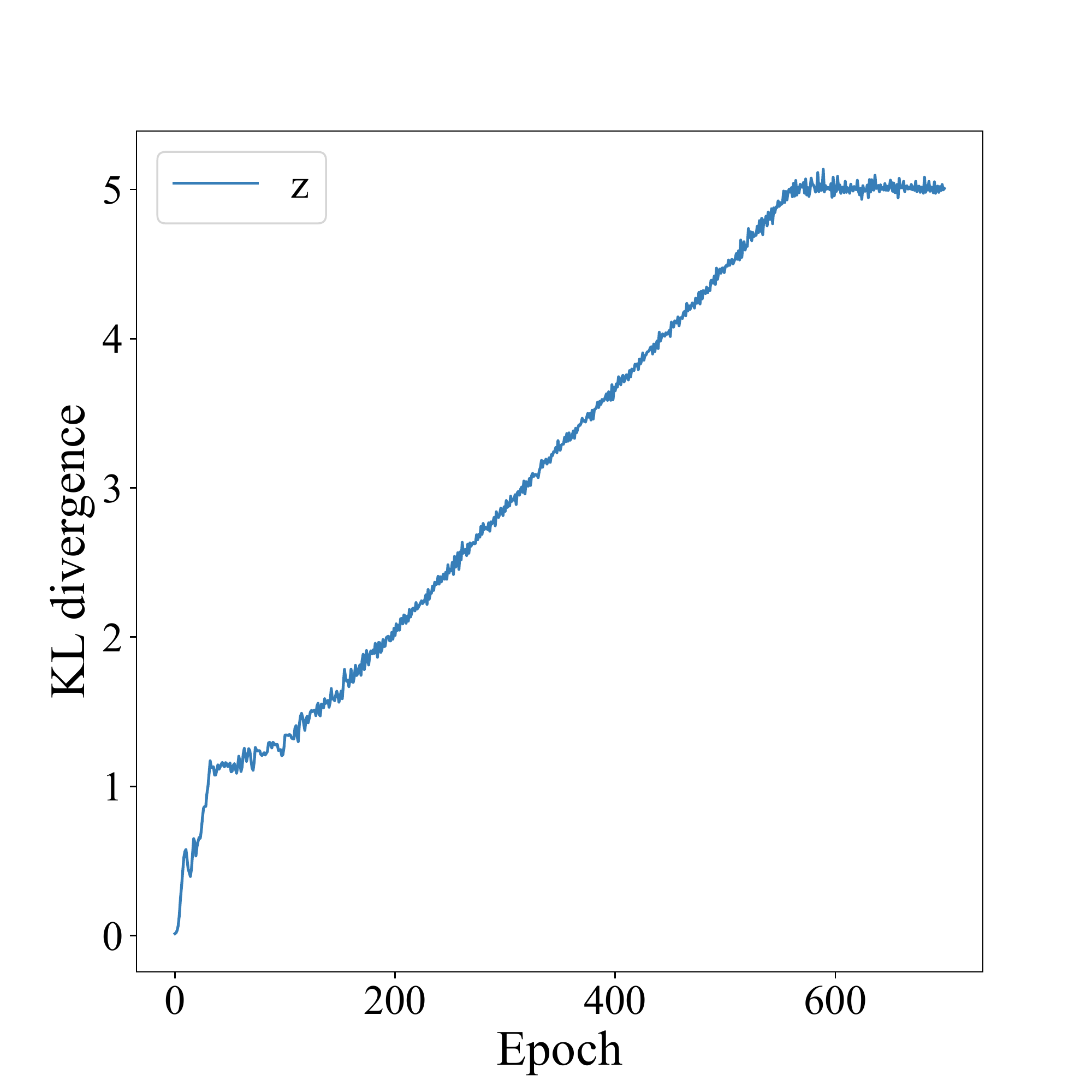}
				\caption{Results with one-dimensional latent variable. }
			\end{subfigure}
			\begin{subfigure}[t]{0.48\columnwidth}
				\centering
				\includegraphics[width=\textwidth]{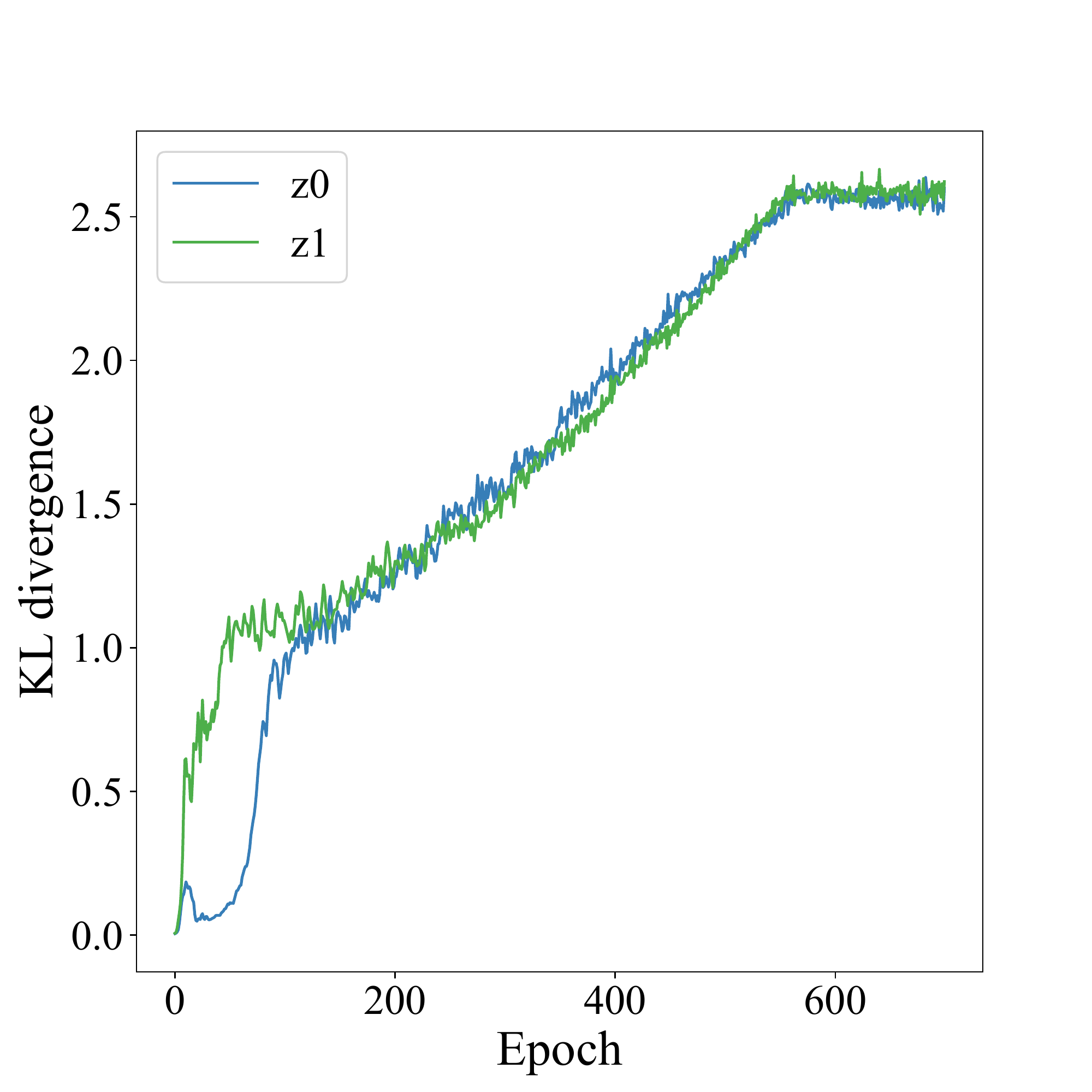}
				\caption{Results with two-dimensional latent variable. }
			\end{subfigure}
			\caption{KL divergence during the training on Task~3. }
			\label{fig:task3_KL}
		\end{figure}

\begin{figure}[tb]
			\centering
			\begin{subfigure}[t]{0.48\columnwidth}
				\centering
				\includegraphics[width=\textwidth]{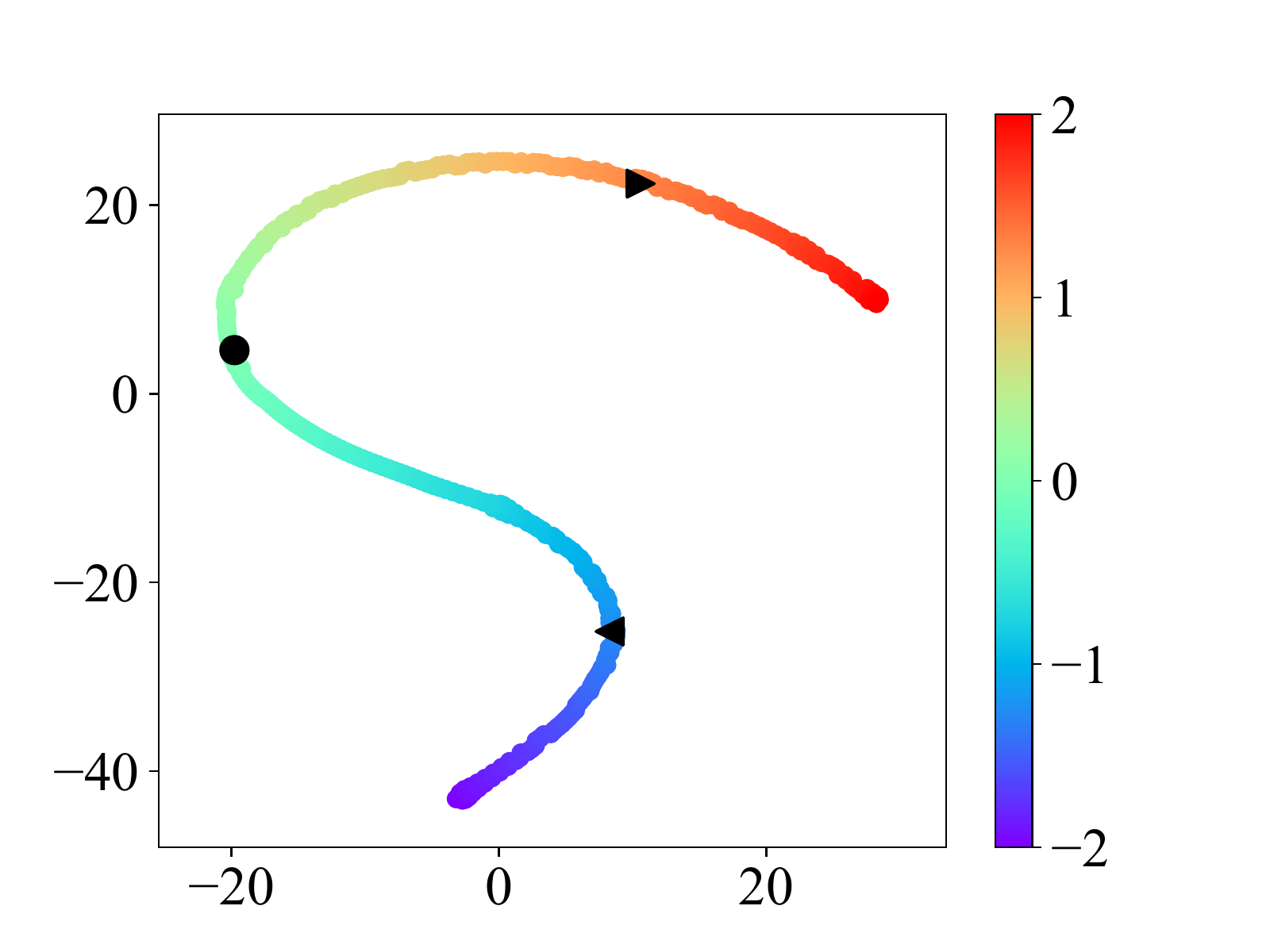}
				\caption{Distribution of solutions before fine-tuning. }
			\end{subfigure}
			\begin{subfigure}[t]{0.48\columnwidth}
				\centering
				\includegraphics[width=\textwidth]{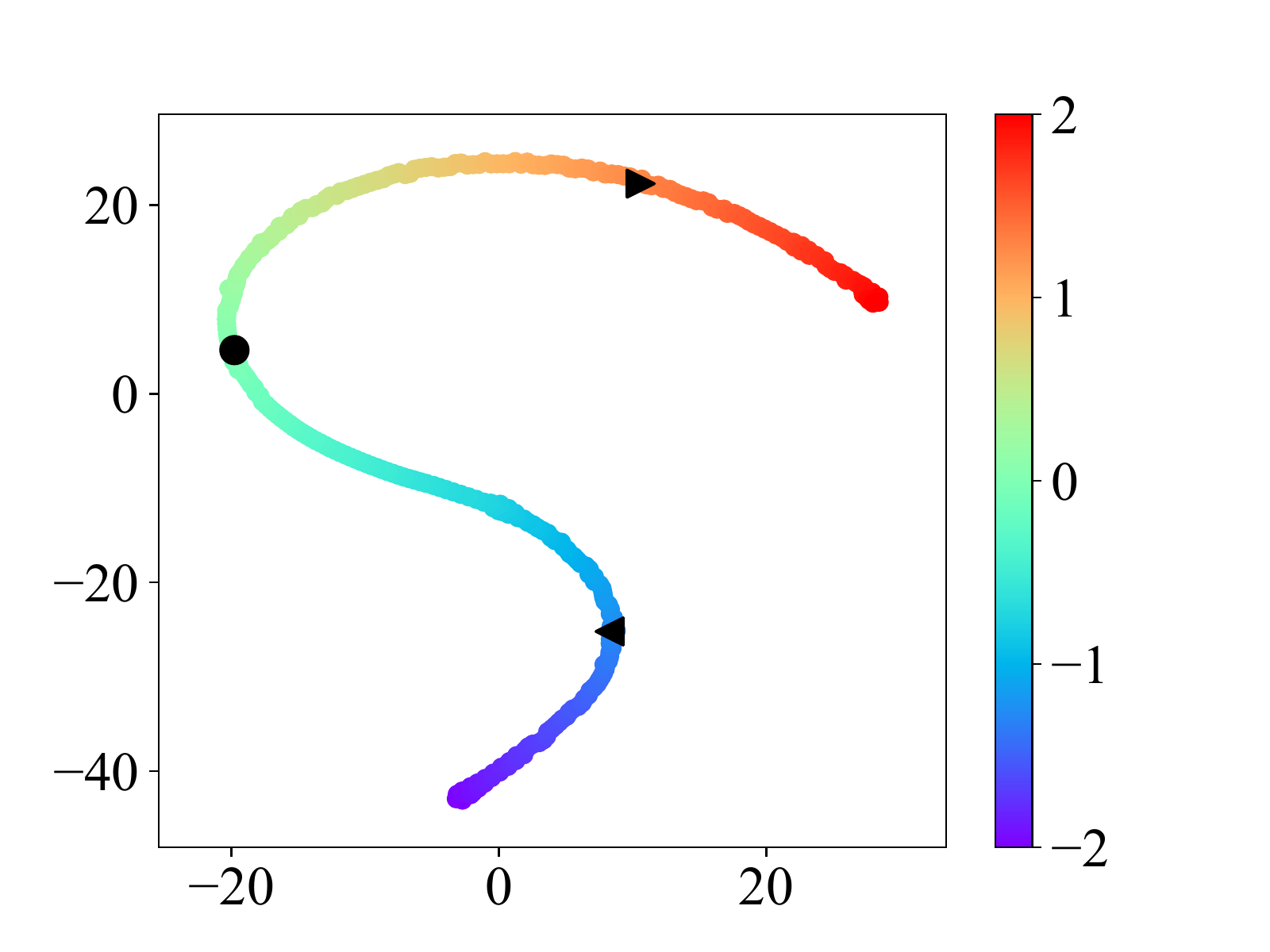}
				\caption{Distribution of solutions after fine-tuning. }
			\end{subfigure}
			\caption{Visualization of distribution of solutions on Task~3. Dimensionality is reduced using t-SNE. The color bar indicates the value of $z$. As fine-tuning was not necessary, distribution of solution did not change between before and after the fine-tuning.  }
			\label{fig:task3_t-sne}
		\end{figure}

\begin{figure}[tb]
			\centering
			\begin{subfigure}[t]{0.32\columnwidth}
				\centering
				\includegraphics[width=\textwidth]{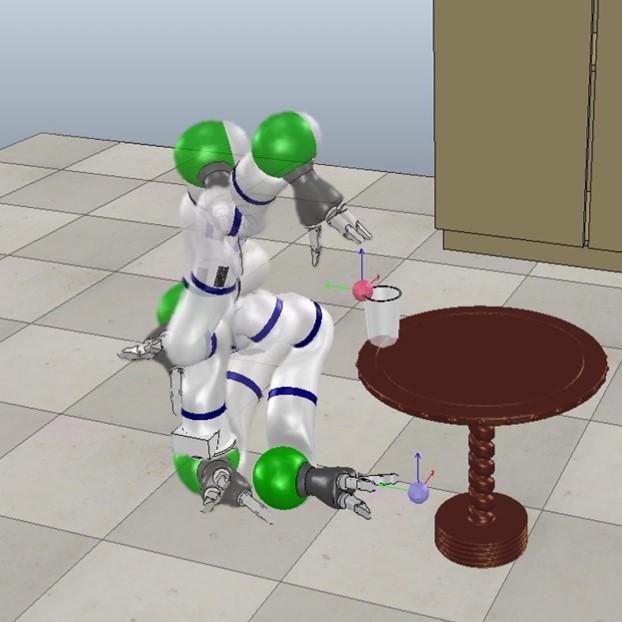}
			\end{subfigure}
			\begin{subfigure}[t]{0.32\columnwidth}
				\centering
				\includegraphics[width=\textwidth]{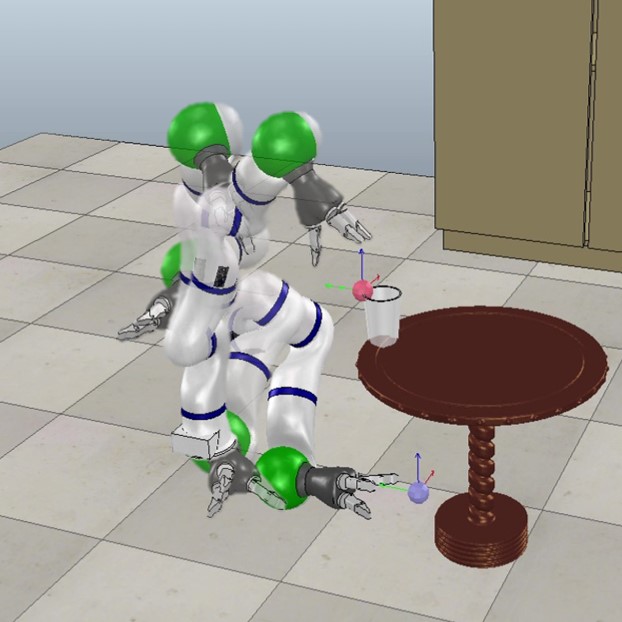}
			\end{subfigure}
			\begin{subfigure}[t]{0.32\columnwidth}
				\centering
				\includegraphics[width=\textwidth]{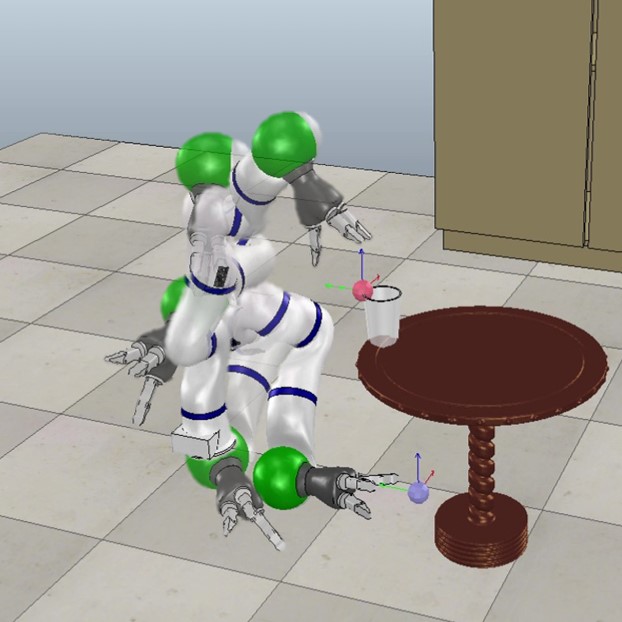}
			\end{subfigure}
			\caption{Three solutions found by SMTO for Task~3.  }
			\label{fig:task3_smto}
		\end{figure}

\begin{figure}[tb]
			\hfill
			\includegraphics[width=0.9\columnwidth]{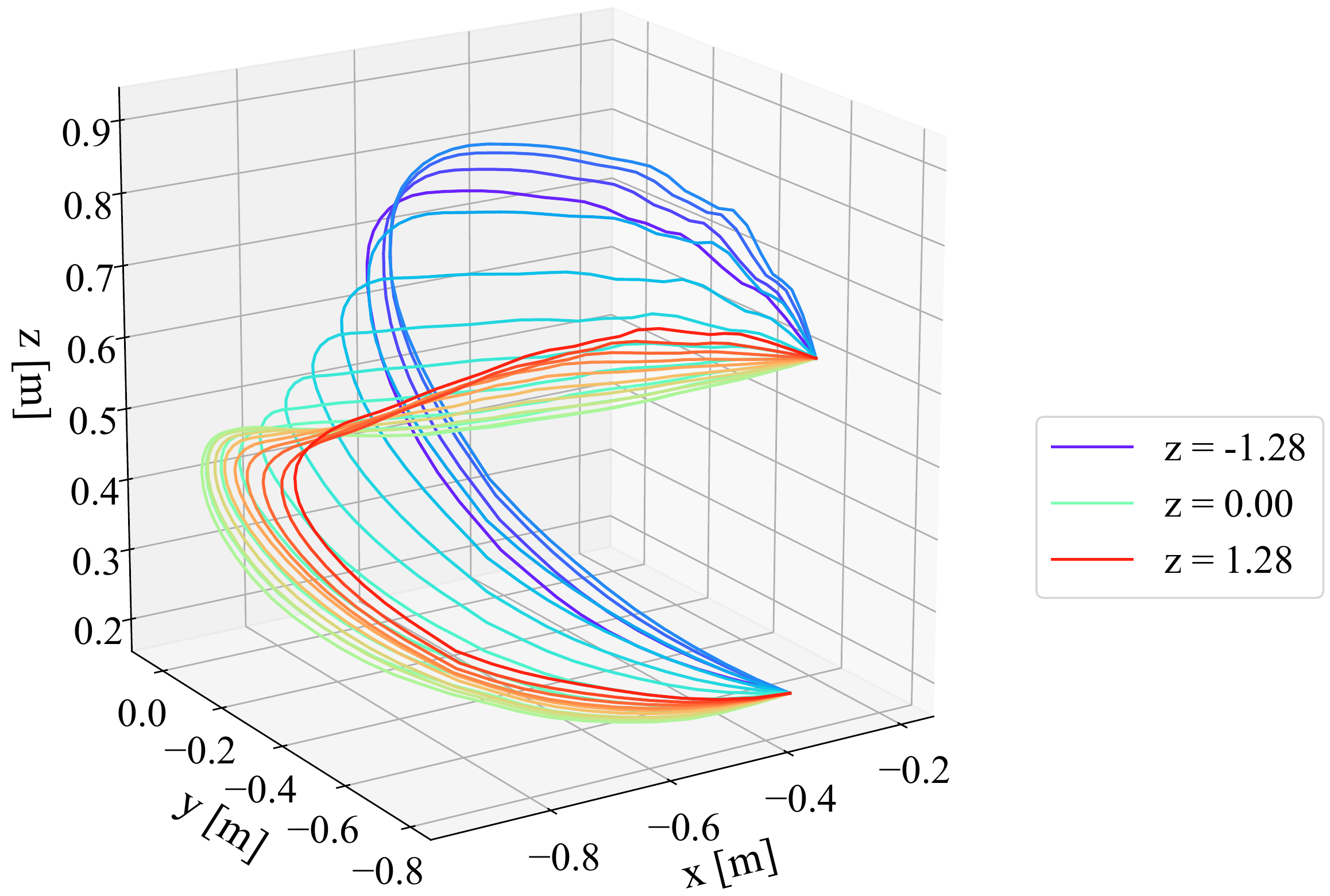}
			\caption{Trajectories in task space for Task~3. Result with one-dimensional latent variable. 20 trajectories are generated by linearly interpolating between $z = -1.28$ and $z = 1.28$. }
			\label{fig:task3_task}
		\end{figure}

Regarding the dimensionality of the latent variable, increasing the dimensionality of the latent variable did not necessarily lead to an increase in the diversity of solutions. 
Although the use of a two-dimensional latent variable led to more diverse solutions than the one-dimensional latent variable for Task~2, this was not the case for Tasks~1 and~3. 
For Task 2, the trajectories shown in Figure~\ref{fig:LSMO_task2}(b) indicate that the two channels $z_0$ and $z_1$ encode different trajectories. 
When the value of $z_0$ is varied, the posture of the manipulator changes while maintaining the height of the end-effector during the motion.
In contrast, the height of the end-effector during the motion changes when the value of $z_1$ is varied. 
When examining the trajectories found for Task~1, which are visualized in Figure~\ref{fig:LSMO_task1_app}(b), the effect of changing the value of $z_0$ is not clear. 
In Figure~\ref{fig:task1_KL}, we show the KL divergence $\KL\big( q_{\vect{\psi}}(\vect{z} | \vect{x}) || p(\vect{z}) \big)$, which is the upper bound of the mutual information between $\vect{z}$ and $\vect{x}$~\citep{Dupont18}.
The plot of the KL divergence also indicates that channel $z_1$ encodes more information than channel $z_0$ on Task~1.
For Task~3, although two channels encode variations of trajectories, as shown in Figure~\ref{fig:LSMO_task3}(b),  the difference of variations encoded in $z_0$ and $z_1$ is not clear.
Figure~\ref{fig:task3_KL} indicates that comparable amount of information is encoded in $z_0$ and $z_1$.
These observations suggest that the information encoded in $z_0$ and $z_1$ is entangled for Task~3.
Therefore, the one-dimensional latent variable was sufficient to model the diversity of solutions for these tasks.




\subsection{Additional Results with Real Robot Experiments}
\label{sec:app_real}

\begin{figure*}
	\begin{minipage}{\textwidth}
		\begin{figure}[H]
			\centering
			\begin{subfigure}[t]{0.7\columnwidth}
				\centering
				\includegraphics[width=\textwidth]{cobotta2_1d}
				\caption{Result with one-dimensional latent variable. }
			\end{subfigure}
			\begin{subfigure}[t]{0.28\columnwidth}
				\centering
				\includegraphics[width=\textwidth]{cobotta2_taskspace}
				\caption{Trajectories in task space. Results with the one-dimensional latent variable. }
			\end{subfigure}
			\begin{subfigure}[t]{0.85\columnwidth}
				\centering
				\includegraphics[width=\textwidth]{cobotta2_2d}
				\caption{Result with two-dimensional latent variable. }
			\end{subfigure}
			\caption{Solutions generated from $p_{\vect{\theta}}(\vect{\xi}|\vect{z})$ with different values of latent variable $\vect{z}$ on Task~4.  }
			\label{fig:cobotta2_app}
		\end{figure}
	\end{minipage}
	\begin{minipage}{0.5\textwidth}
		\begin{figure}[H]
			\centering
			\begin{subfigure}[t]{0.32\columnwidth}
				\centering
				\includegraphics[width=\textwidth]{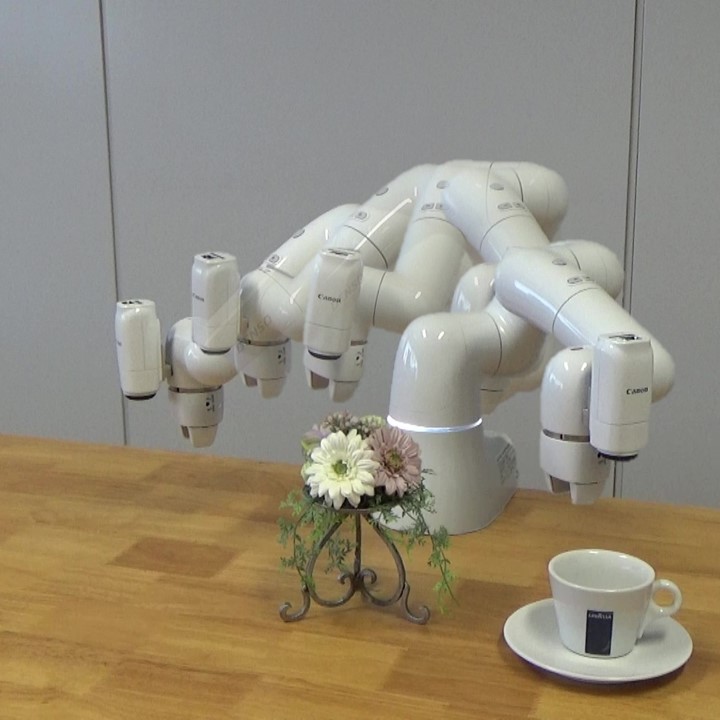}
			\end{subfigure}
			\begin{subfigure}[t]{0.32\columnwidth}
				\centering
				\includegraphics[width=\textwidth]{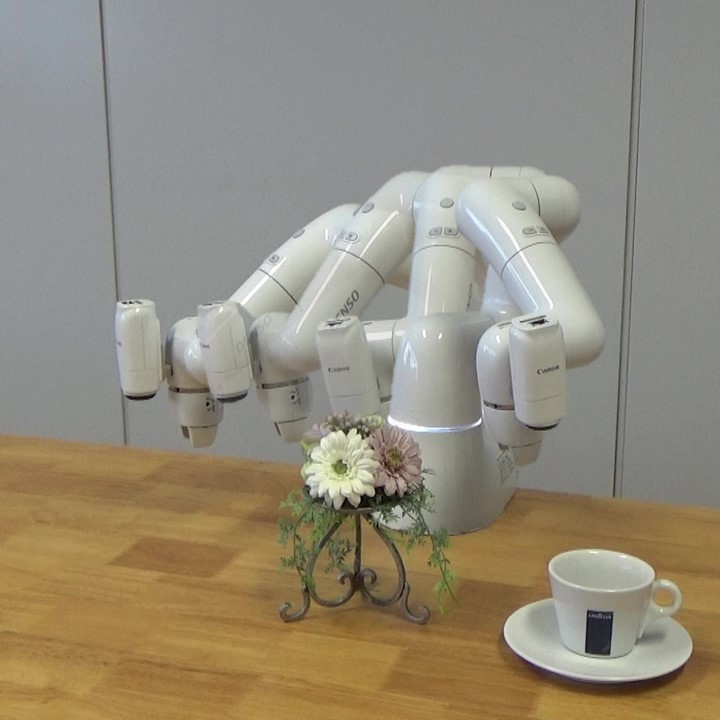}
			\end{subfigure}
			\begin{subfigure}[t]{0.32\columnwidth}
				\centering
				\includegraphics[width=\textwidth]{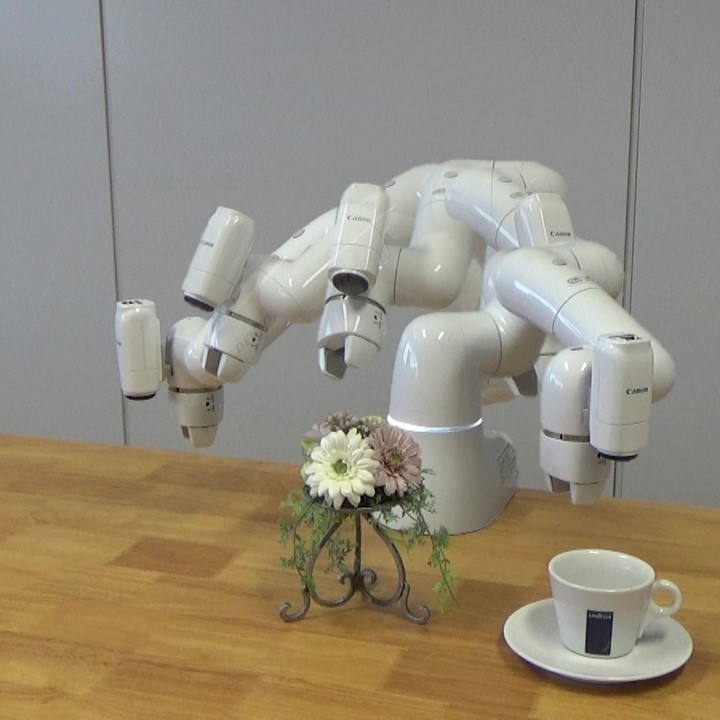}
			\end{subfigure}
			\caption{Three solutions generated by SMTO for Task~4.  }
			\label{fig:cobotta2_smto}
			\vspace{1cm}
		\end{figure}
	\end{minipage}
	\begin{minipage}{0.5\textwidth}
		\begin{figure}[H]
			\centering
			\begin{subfigure}[t]{0.48\columnwidth}
				\centering
				\includegraphics[width=\textwidth]{t-sne_LSMO_cobotta2_1d}
				\caption{Result with one-dimensional latent variable. }
			\end{subfigure}
			\begin{subfigure}[t]{0.48\columnwidth}
				\centering
				\includegraphics[width=\textwidth]{t-sne_LSMO_cobotta2_2d}
				\caption{Result with two-dimensional latent variable.}
			\end{subfigure}
			\caption{Distribution of solutions on Task~4. Dimensionality is reduced using same transformation in (a) and (b).  The color bar indicates value of $z$ in (a) and $z_0$ in (b), respectively. }
			\label{fig:cobotta2_t-sne_app}
		\end{figure}
	\end{minipage}
\end{figure*}

\begin{figure*}
	\begin{minipage}{\textwidth}
		\begin{figure}[H]
			\centering
			\begin{subfigure}[t]{0.7\columnwidth}
				\centering
				\includegraphics[width=\textwidth]{cobotta3_1d}
				\caption{Result with one-dimensional latent variable. }
			\end{subfigure}
			\begin{subfigure}[t]{0.28\columnwidth}
				\centering
				\includegraphics[width=\textwidth]{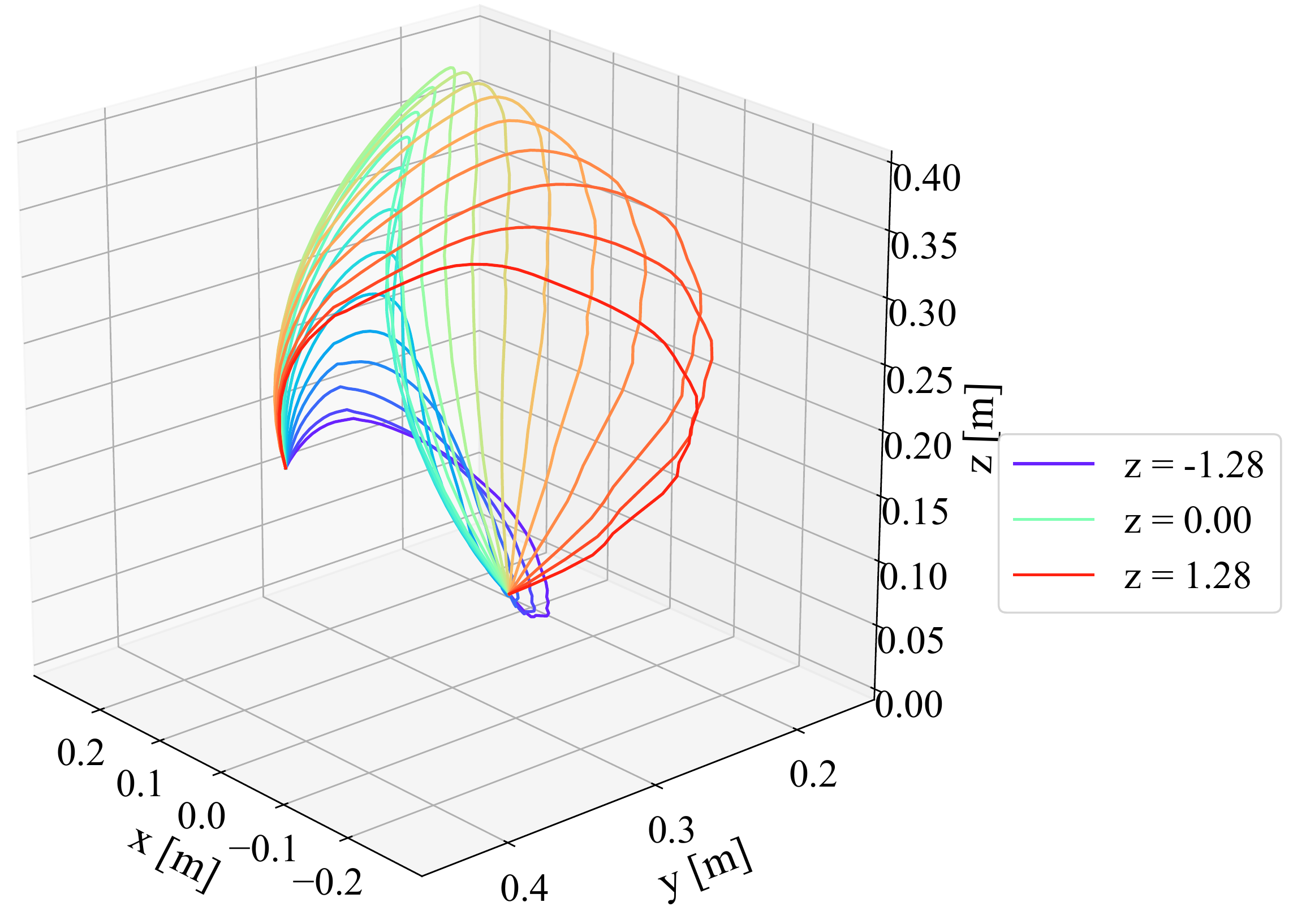}
				\caption{Trajectories in task space. Results with the one-dimensional latent variable. }
			\end{subfigure}
			\begin{subfigure}[t]{0.85\columnwidth}
				\centering
				\includegraphics[width=\textwidth]{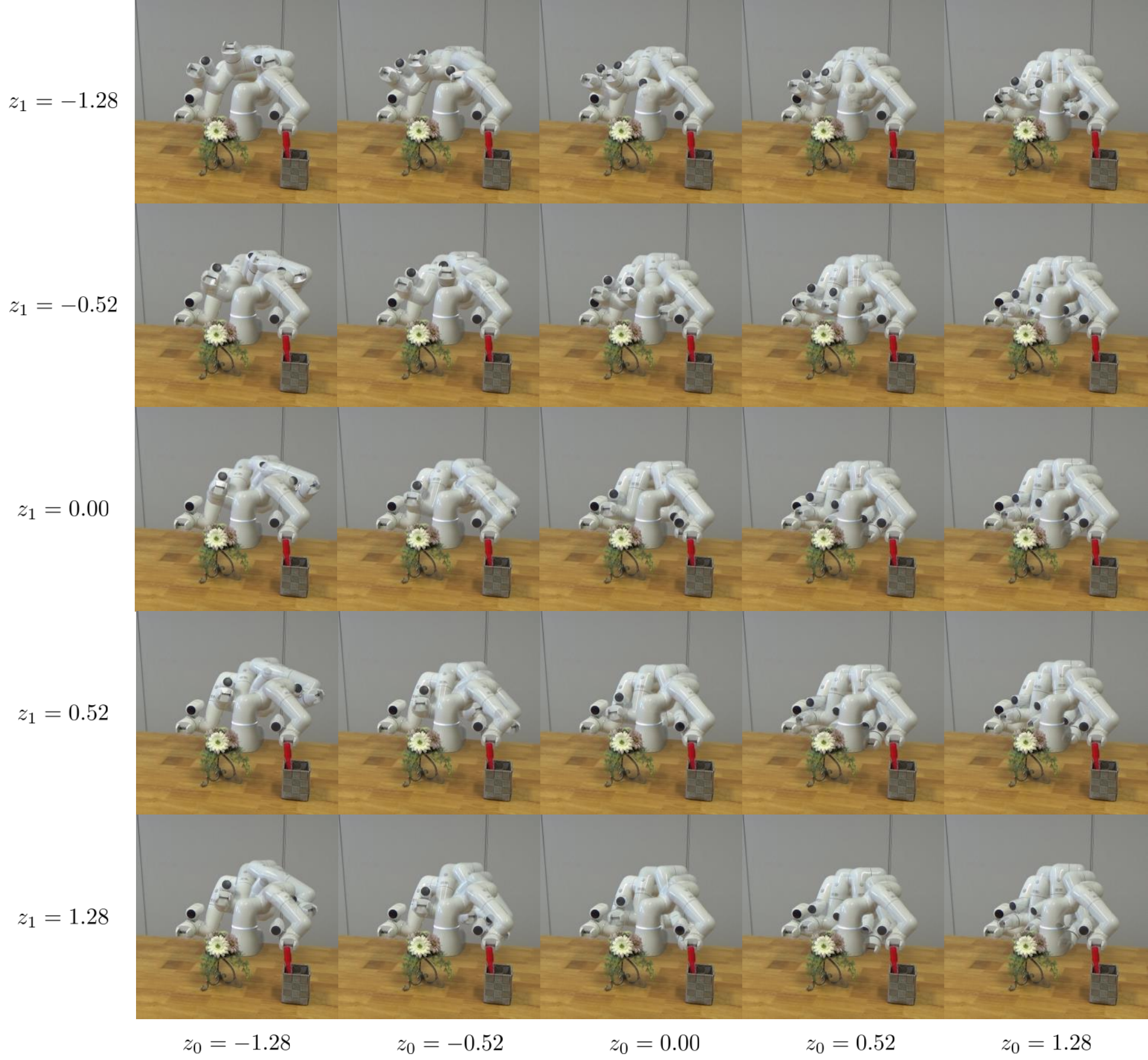}
				\caption{Result with two-dimensional latent variable. }
			\end{subfigure}
			\caption{Solutions generated from $p_{\vect{\theta}}(\vect{\xi}|\vect{z})$ with different values of latent variable $\vect{z}$ on Task~5.  }
			\label{fig:cobotta3}
		\end{figure}
	\end{minipage}
	\begin{minipage}{0.5\textwidth}
		\begin{figure}[H]
			\centering
			\begin{subfigure}[t]{0.32\columnwidth}
				\centering
				\includegraphics[width=\textwidth]{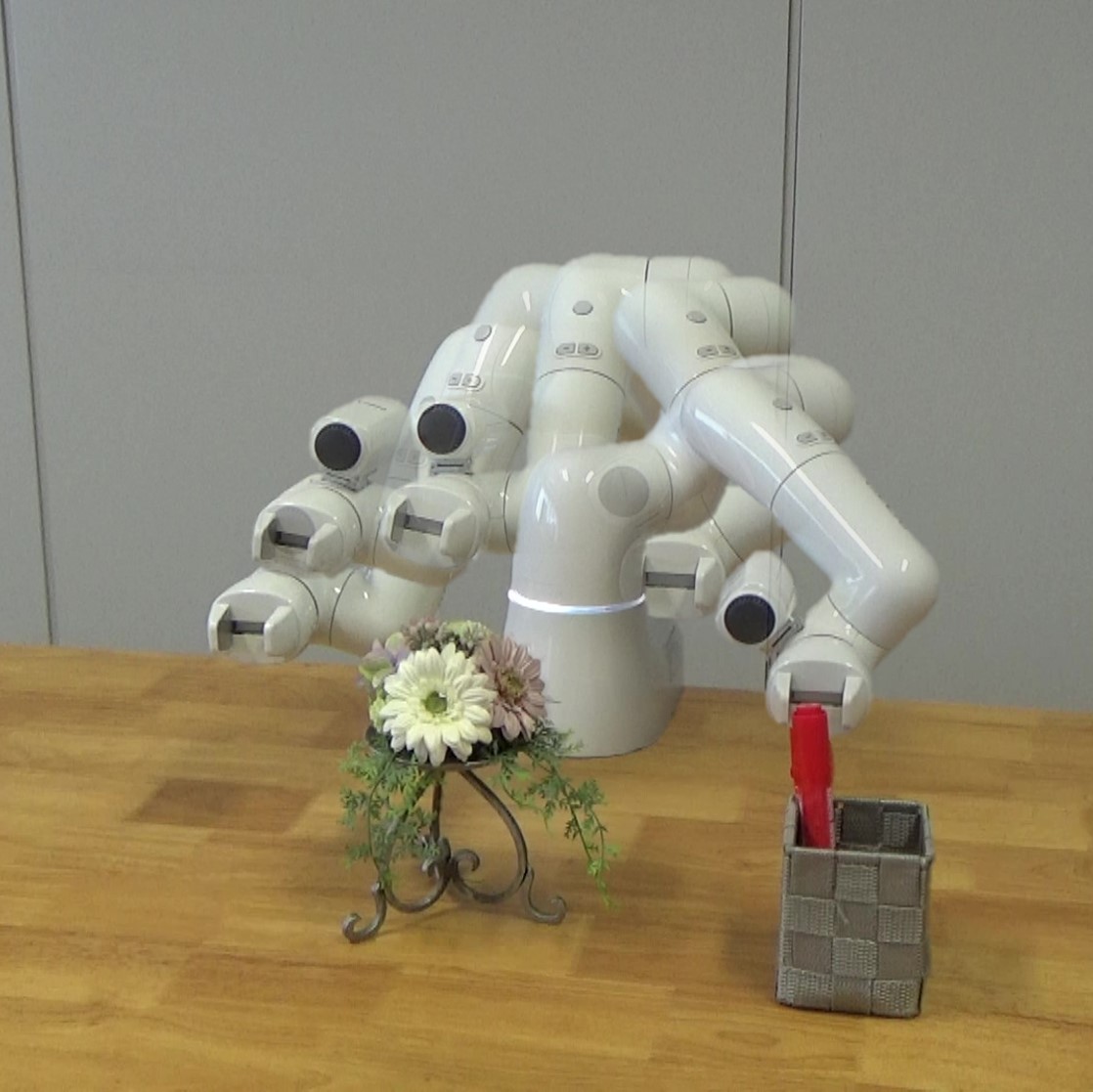}
			\end{subfigure}
			\begin{subfigure}[t]{0.32\columnwidth}
				\centering
				\includegraphics[width=\textwidth]{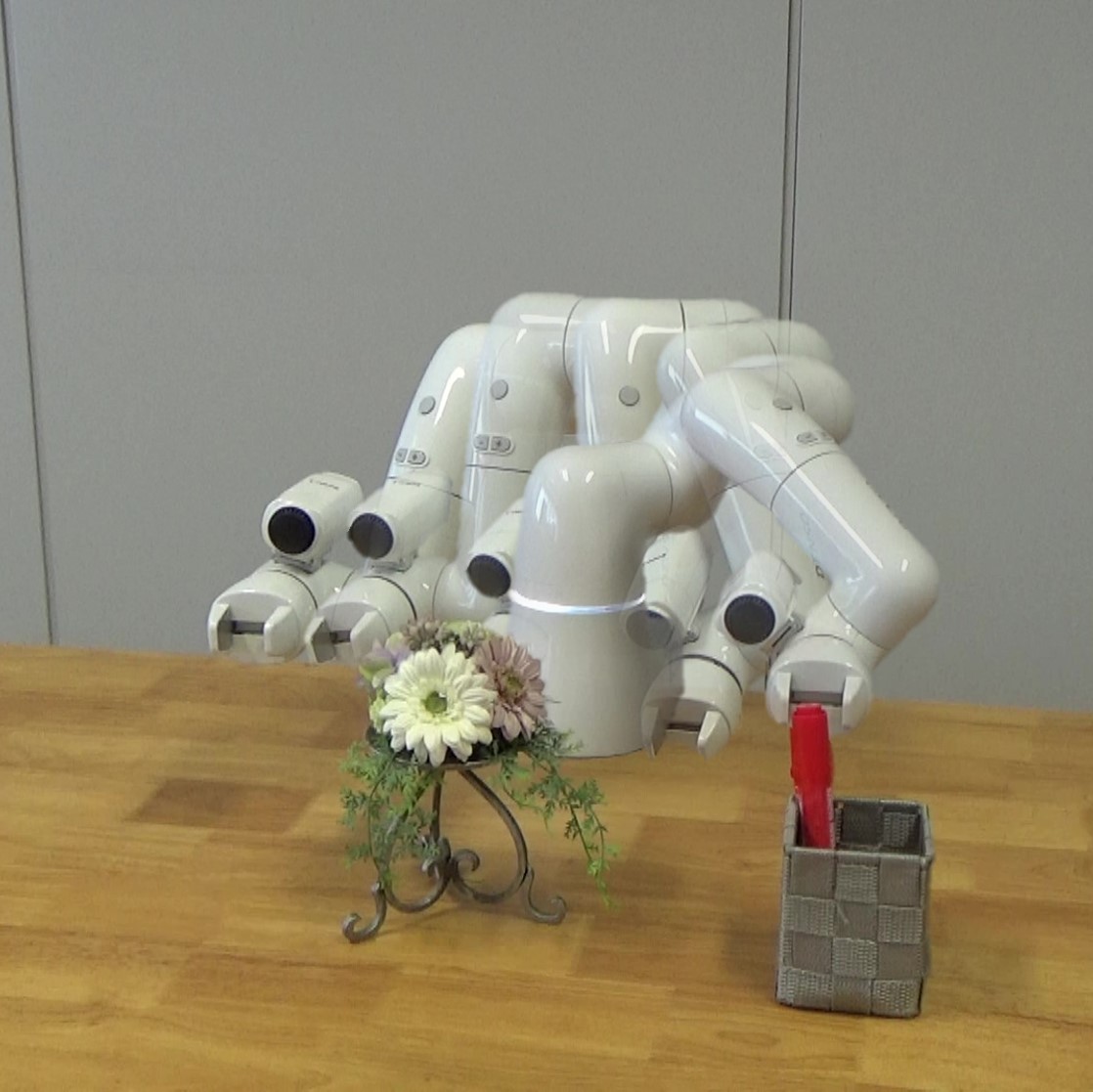}
			\end{subfigure}
			\begin{subfigure}[t]{0.32\columnwidth}
				\centering
				\includegraphics[width=\textwidth]{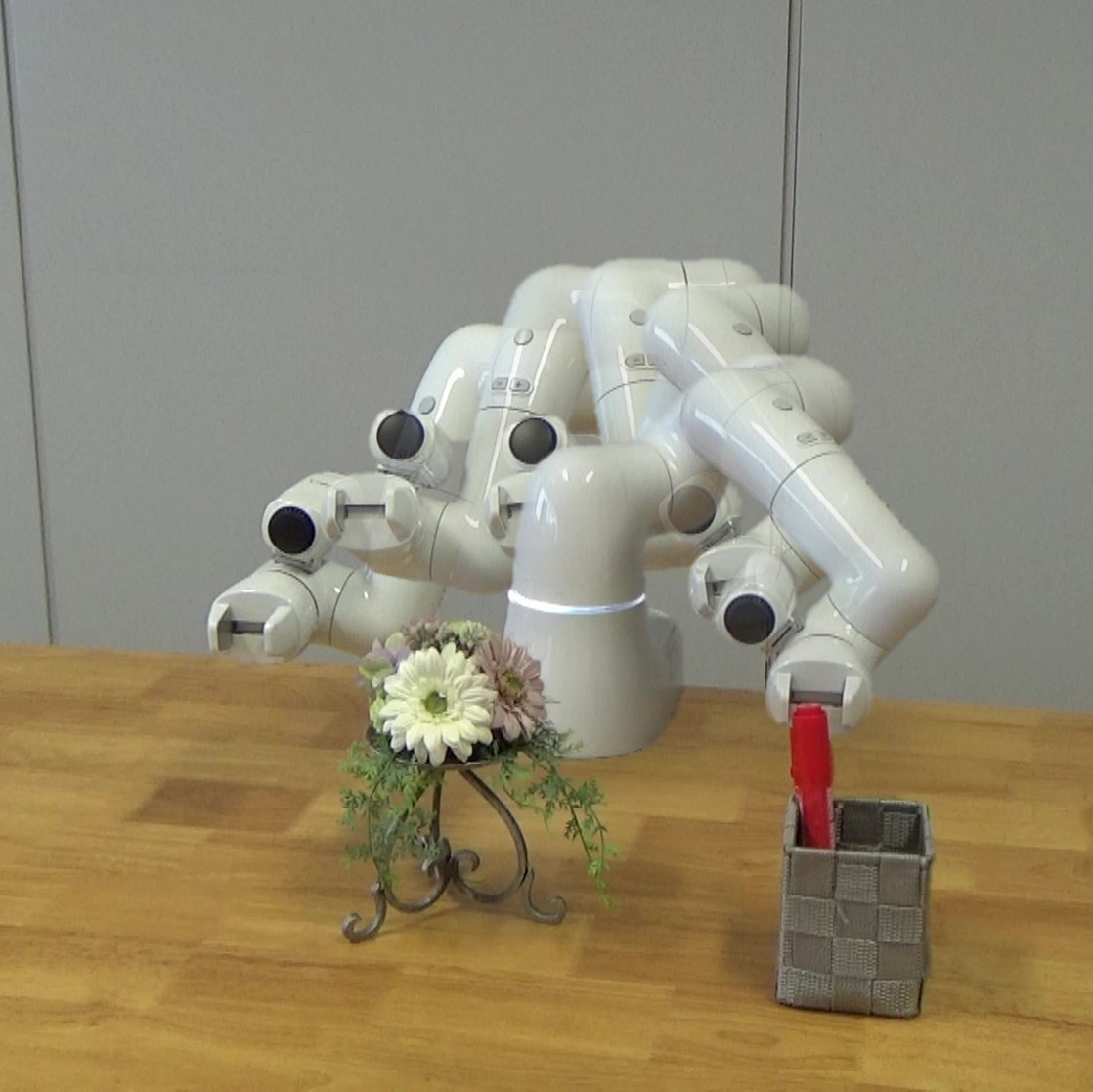}
			\end{subfigure}
			\caption{Three solutions generated by SMTO for Task~5.  }
			\label{fig:cobotta3_smto}
			\vspace{1cm}
		\end{figure}
	\end{minipage}
	\begin{minipage}{0.5\textwidth}
		\begin{figure}[H]
			\centering
			\begin{subfigure}[t]{0.48\columnwidth}
				\centering
				\includegraphics[width=\textwidth]{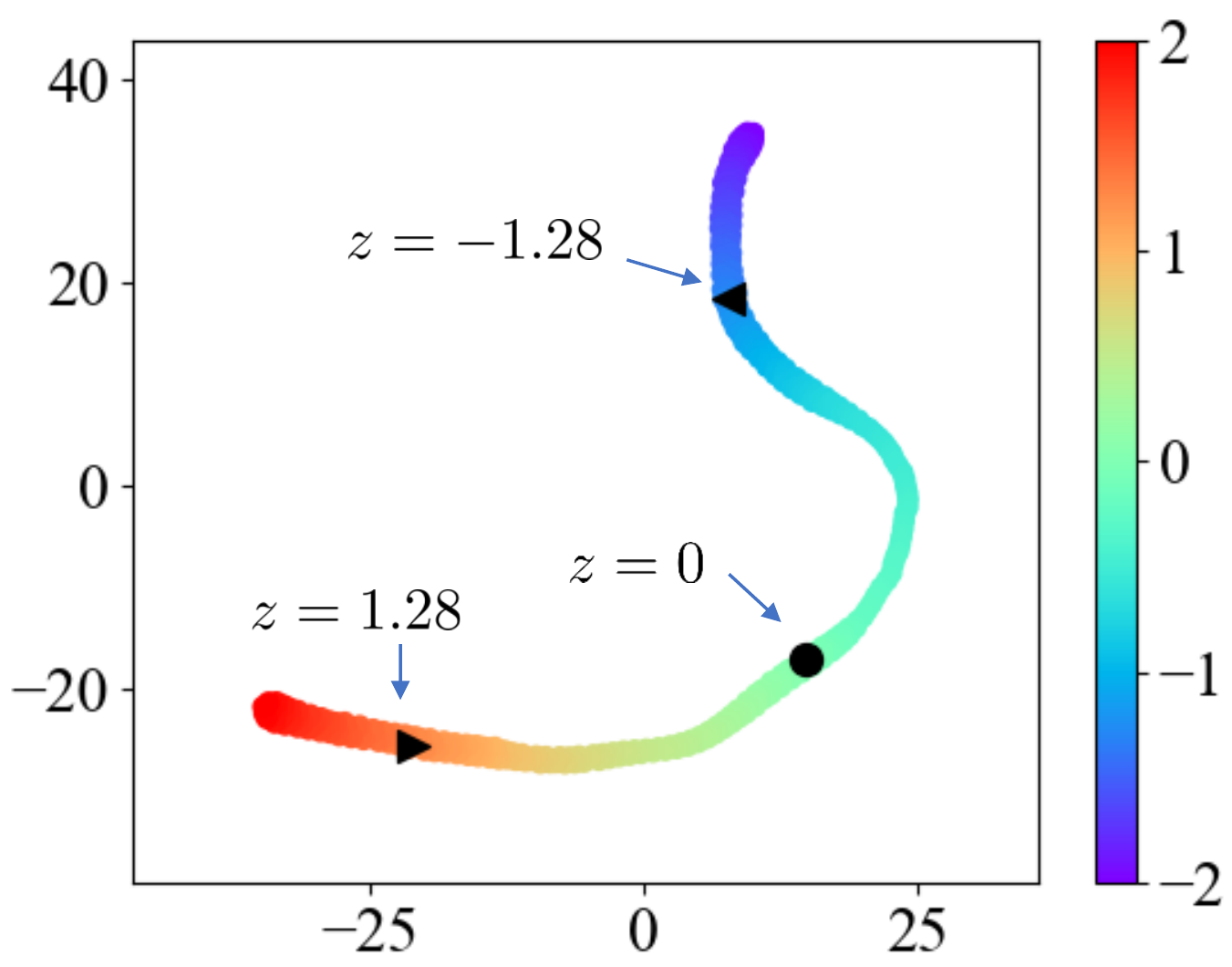}
				\caption{Result with one-dimensional latent variable. }
			\end{subfigure}
			\begin{subfigure}[t]{0.48\columnwidth}
				\centering
				\includegraphics[width=\textwidth]{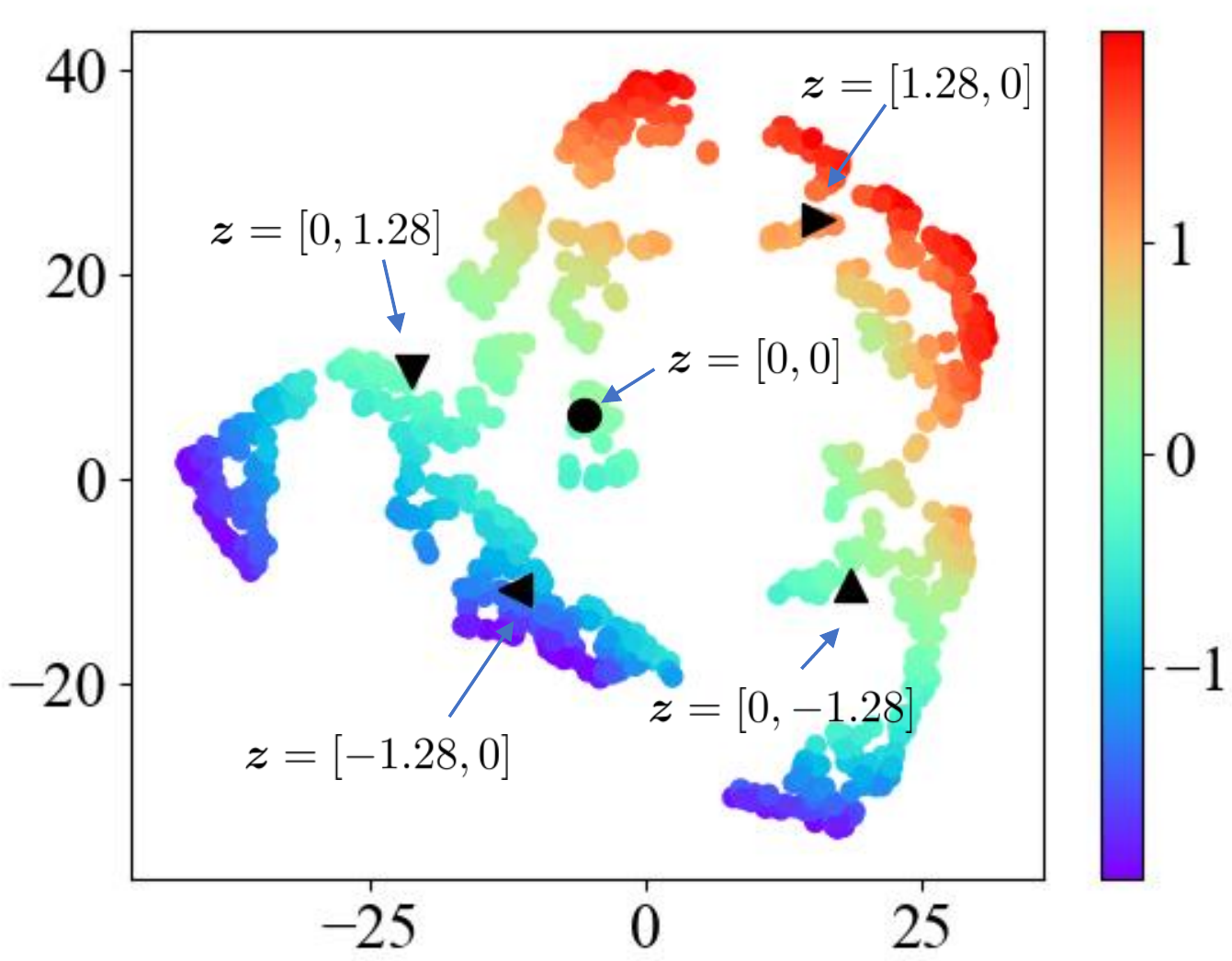}
				\caption{Result with two-dimensional latent variable.}
			\end{subfigure}
			\caption{Distribution of solutions on Task~5. Dimensionality is reduced using same transformation in (a) and (b).  Color bar indicates value of $z$ in (a) and $z_0$ in (b), respectively. }
			\label{fig:cobotta3_t-sne}
		\end{figure}
	\end{minipage}
\end{figure*}

\begin{figure*}
	\begin{minipage}{\textwidth}
		\begin{figure}[H]
			\centering
			\begin{subfigure}[t]{0.7\columnwidth}
				\centering
				\includegraphics[width=\textwidth]{cobotta4_1d}
				\caption{Result with one-dimensional latent variable. }
			\end{subfigure}
			\begin{subfigure}[t]{0.28\columnwidth}
				\centering
				\includegraphics[width=\textwidth]{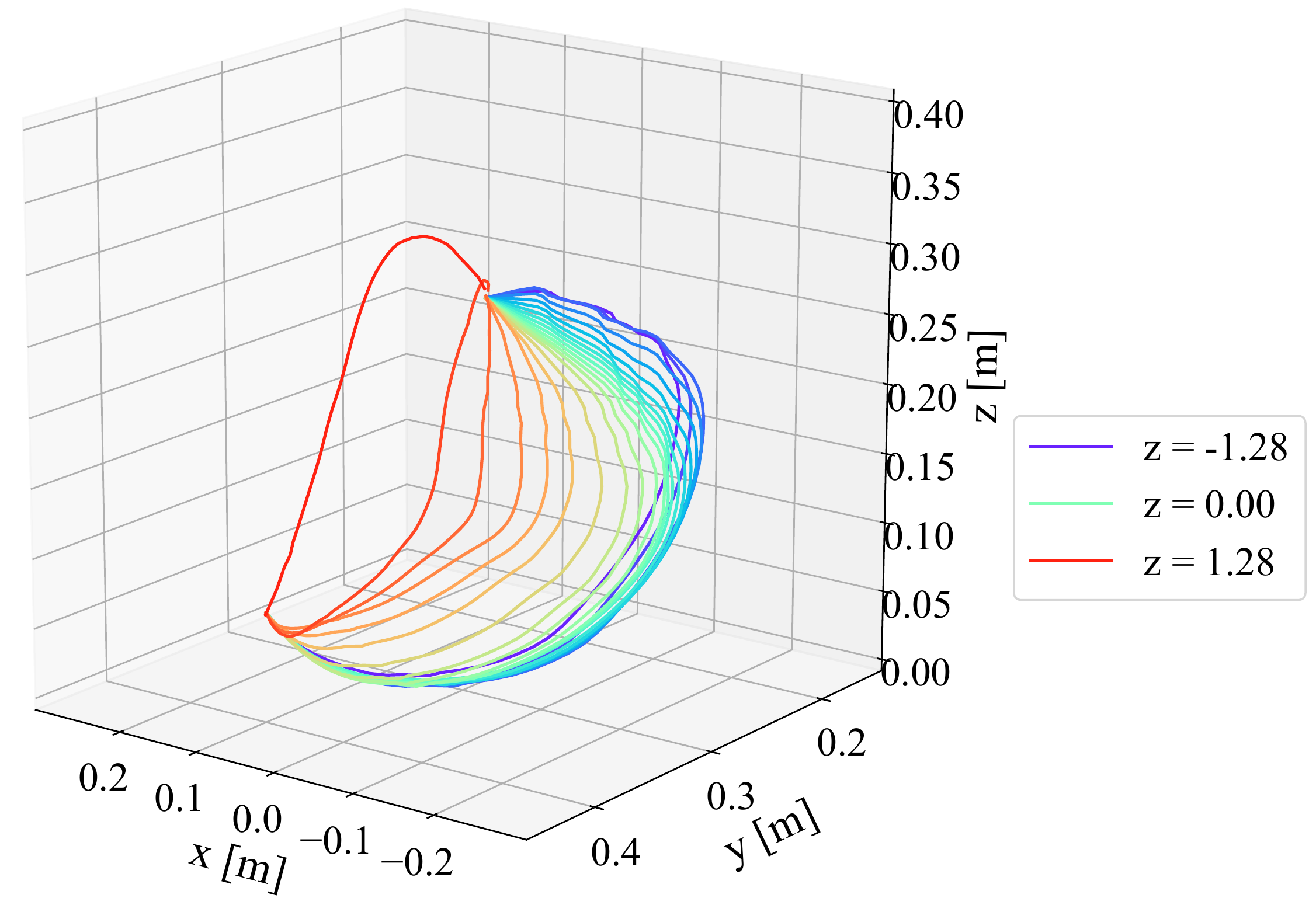}
				\caption{Trajectories in task space. Results with the one-dimensional latent variable. }
			\end{subfigure}
			\begin{subfigure}[t]{0.85\columnwidth}
				\centering
				\includegraphics[width=\textwidth]{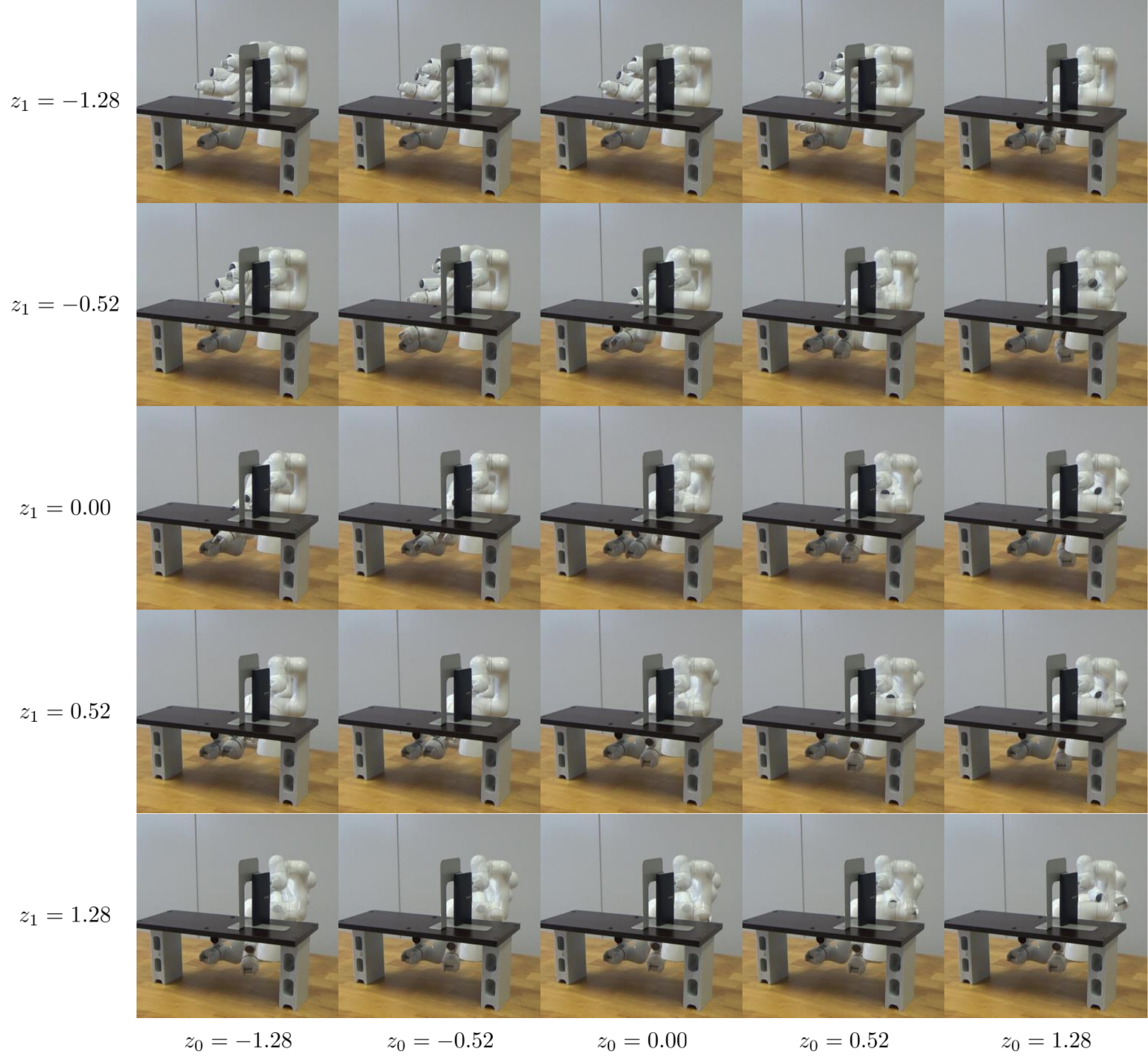}
				\caption{Result with two-dimensional latent variable. }
			\end{subfigure}
			\caption{Solutions generated from $p_{\vect{\theta}}(\vect{\xi}|\vect{z})$ with different values of latent variable $\vect{z}$ on Task~6.  }
			\label{fig:cobotta4}
		\end{figure}
	\end{minipage}
	\begin{minipage}{0.5\textwidth}
		\begin{figure}[H]
			\centering
			\begin{subfigure}[t]{0.4\columnwidth}
				\centering
				\includegraphics[width=\textwidth]{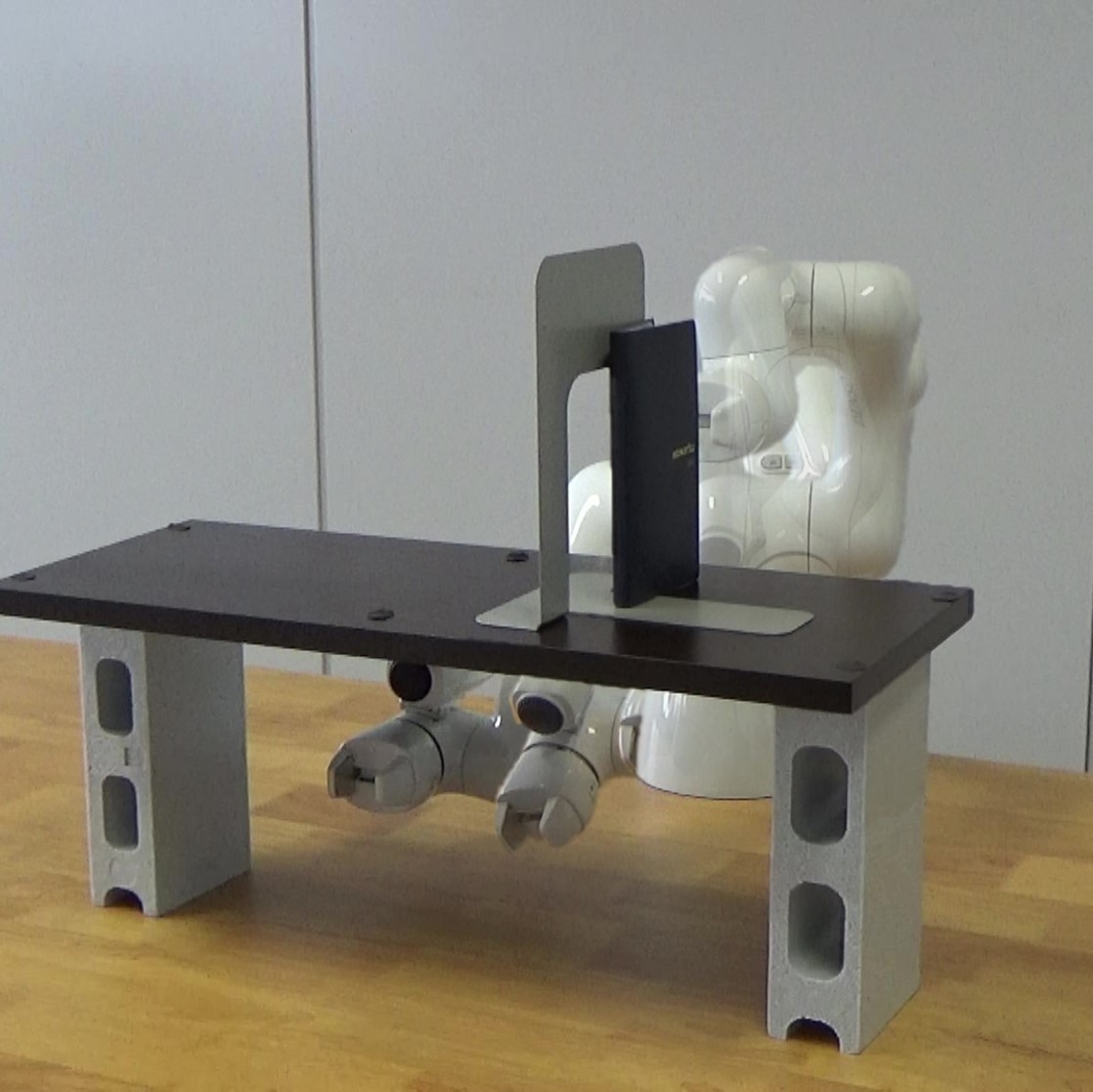}
			\end{subfigure}
			\begin{subfigure}[t]{0.4\columnwidth}
				\centering
				\includegraphics[width=\textwidth]{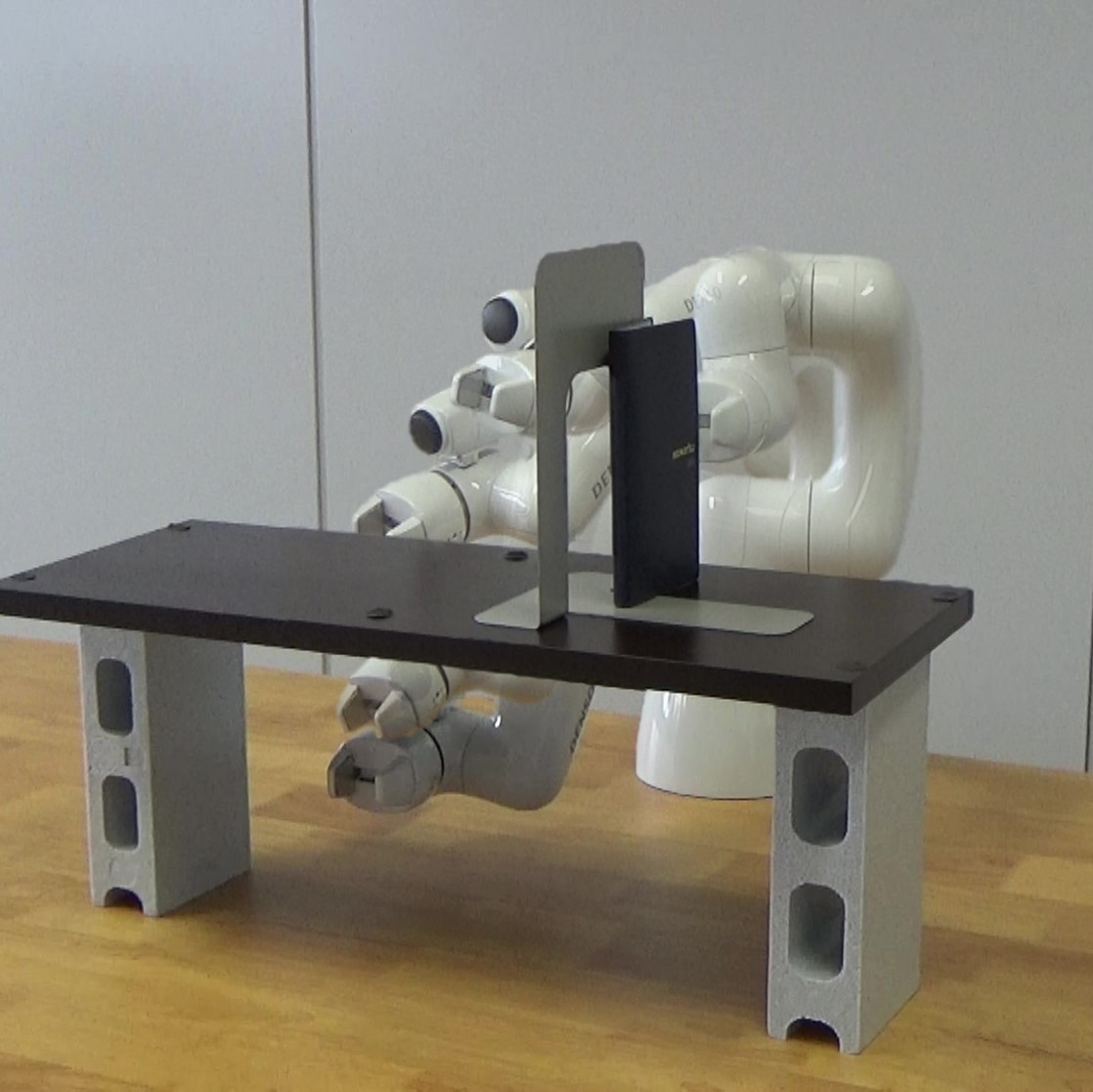}
			\end{subfigure}
			\caption{Two solutions generated by SMTO for Task~6.  }
			\label{fig:cobotta4_smto}
			\vspace{0.cm}
		\end{figure}
	\end{minipage}
	\begin{minipage}{0.5\textwidth}
		\begin{figure}[H]
			\centering
			\begin{subfigure}[t]{0.48\columnwidth}
				\centering
				\includegraphics[width=\textwidth]{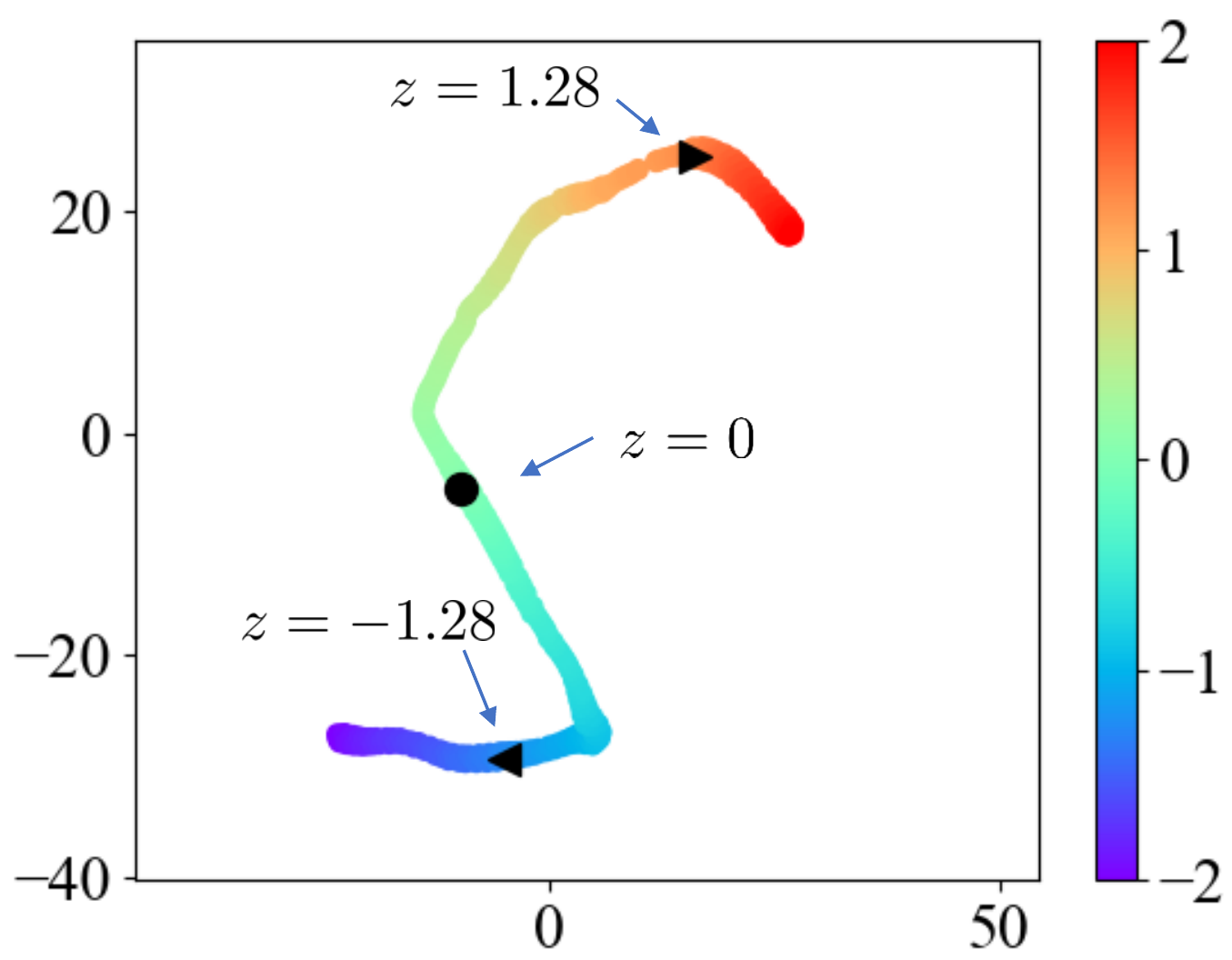}
				\caption{Result with one-dimensional latent variable. }
			\end{subfigure}
			\begin{subfigure}[t]{0.48\columnwidth}
				\centering
				\includegraphics[width=\textwidth]{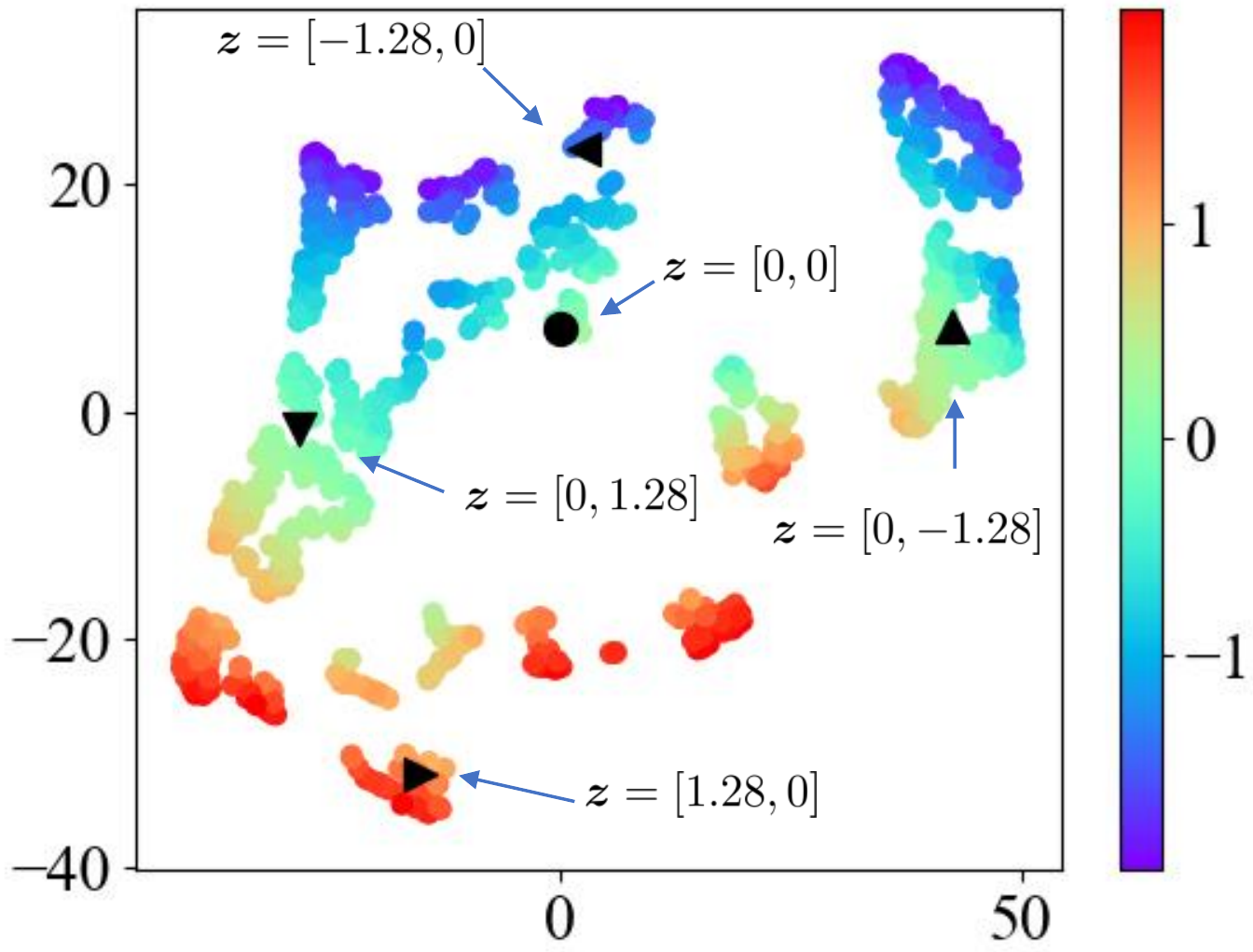}
				\caption{Result with two-dimensional latent variable.}
			\end{subfigure}
			\caption{Distribution of solutions on Task~6. Dimensionality is reduced using same transformation in (a) and (b).  Color bar indicates value of $z$ in (a) and $z_0$ in (b), respectively. }
			\label{fig:cobotta4_t-sne}
		\end{figure}
	\end{minipage}
\end{figure*}

Solutions found for Tasks~4, 5 and 6 are shown in Figure~\ref{fig:cobotta2_app}--\ref{fig:cobotta4_t-sne}.
For Task~5, the height of the end-effector continuously changes as the value of the latent variable changes, as shown in Figure~\ref{fig:cobotta3}.
For Task~6, when the latent variable is one-dimensional, the end-effector avoids the shelf from the right-hand side if $z=-1.28$, whereas the end-effector avoids the shelf from the left-hand side if $z=1.28$, as shown in Figure~\ref{fig:cobotta4}. 
Solutions found by SMTO for Tasks~4, 5, and 6 are shown in Figures~\ref{fig:cobotta2_smto}, \ref{fig:cobotta3_smto}, and \ref{fig:cobotta4_smto}, respectively.
It is evident that MPSM found more diverse solutions than SMTO.

\end{document}